%% file: mainv2.tex
\documentclass[sigconf]{acmart}
\usepackage[utf8]{inputenc}
\usepackage{textcomp}

\renewcommand\footnotetextcopyrightpermission[1]{}
\usepackage{algorithm}
\usepackage{algorithmic}
\usepackage{microtype}
\usepackage{graphicx}
\usepackage{subcaption}
\usepackage{booktabs}
\usepackage{amsmath}
\usepackage{mathtools}
\usepackage{bm}
\usepackage{multirow}
\usepackage{makecell}
\usepackage{array}
\usepackage[dvipsnames,table]{xcolor}
\usepackage{tikz}
\usepackage{listings}
\usepackage{enumitem}
\usepackage{pifont}
\usepackage[capitalize,noabbrev]{cleveref}

\lstdefinestyle{iccadpy}{
  language=Python,
  basicstyle=\ttfamily\footnotesize,
  keywordstyle=\color{blue},
  commentstyle=\color{gray},
  stringstyle=\color{teal},
  breaklines=true,
  columns=fullflexible,
  keepspaces=true,
  showstringspaces=false,
  frame=single,
  tabsize=2
}

\newcolumntype{C}[1]{>{\centering\arraybackslash}m{#1}}

\newcommand{\cmark}{\textcolor[rgb]{0,0.6,0}{\checkmark}}
\newcommand{\xmark}{\textcolor{red}{$\times$}}

\AtBeginDocument{%
  }

\begin{document}

\title{ParasGB: A Graph Benchmark Suite for Parasitic Estimation on AMS Circuits}


\author{
Jiajun Zou$^{1}$,
Jiawei Liu$^2$,
Ao Liu$^1$,
Junnong Tian$^1$,
Yibin Zhang$^3$,
Chengjie Liu$^{4,5}$,
Yuxi Wang$^4$,\\
Shan Shen$^{1*}$,
Wenhua Gu$^{1*}$,
Jun Yang$^{5,6}$,
and Wenjian Yu$^{3*}$
}

\thanks{
This work was supported by the Fundamental Research Funds for the Central Universities (Grant No. 30925010605), Jiangsu Provincial Key Research and Development Program (No. BE2023818), Basic Research Program of Jiangsu (Nos.BK20243046, BK20253062).
$^*$Corresponding authors: S. Shen, W. Gu, and W. Yu.
}

\affiliation{
$^1$\textit{Nanjing University of Science and Technology, Nanjing, China}\\
$^2$\textit{The Chinese University of Hong Kong, Hong Kong SAR, China}\\
$^3$\textit{Dept. Computer Science \& Tech., BNRist, Tsinghua University, Beijing, China}\\
$^4$\textit{School of Electronic Science and Engineering, Nanjing University, Nanjing, China}\\
$^5$\textit{National Center of Technology Innovation for EDA, Nanjing, China}\\
$^6$\textit{School of Integrated Circuits, Southeast University, Nanjing, China}
\country{}
}

\renewcommand{\shortauthors}{Zou et al.}

\begin{abstract}
As chip manufacturing processes advance to deep submicron nodes, parasitic interconnect effects increasingly dominate the performance of analog and mixed-signal (AMS) circuits and often lead to costly layout iterations. This makes early-stage estimation of parasitic capacitance and resistance important for parasitic-aware design exploration before full physical implementation. However, progress on GNN-based parasitic modeling has been hindered by the lack of public, high-fidelity RC benchmarks that support reproducible evaluation.
To address this gap, we introduce ParasGB, the first open-source benchmark suite for pre-layout parasitic parameter prediction on circuit graphs. ParasGB provides large-scale, heterogeneous RC networks extracted with commercial EDA tools from tape-out-proven designs, together with a unified evaluation protocol covering node-level ground capacitance, edge-level resistance, and edge-level coupling capacitance. Within this framework, we benchmark diverse GNN architectures using a standardized training pipeline and expose challenges such as extreme label imbalance, long-tailed parasitic distributions, and strong structural heterogeneity.
By establishing a physically grounded and standardized benchmark for early-stage parasitic prediction, ParasGB provides an open platform for reproducible research on circuit graph learning and parasitic-aware model development. All datasets, preprocessing scripts, and configurations are publicly available in our code repository.
\footnote{https://github.com/ShenShan123/ParasGB.git}
\end{abstract}

\maketitle

\section{Introduction}

\begin{figure}
    \centering
    \includegraphics[width=0.94\linewidth]{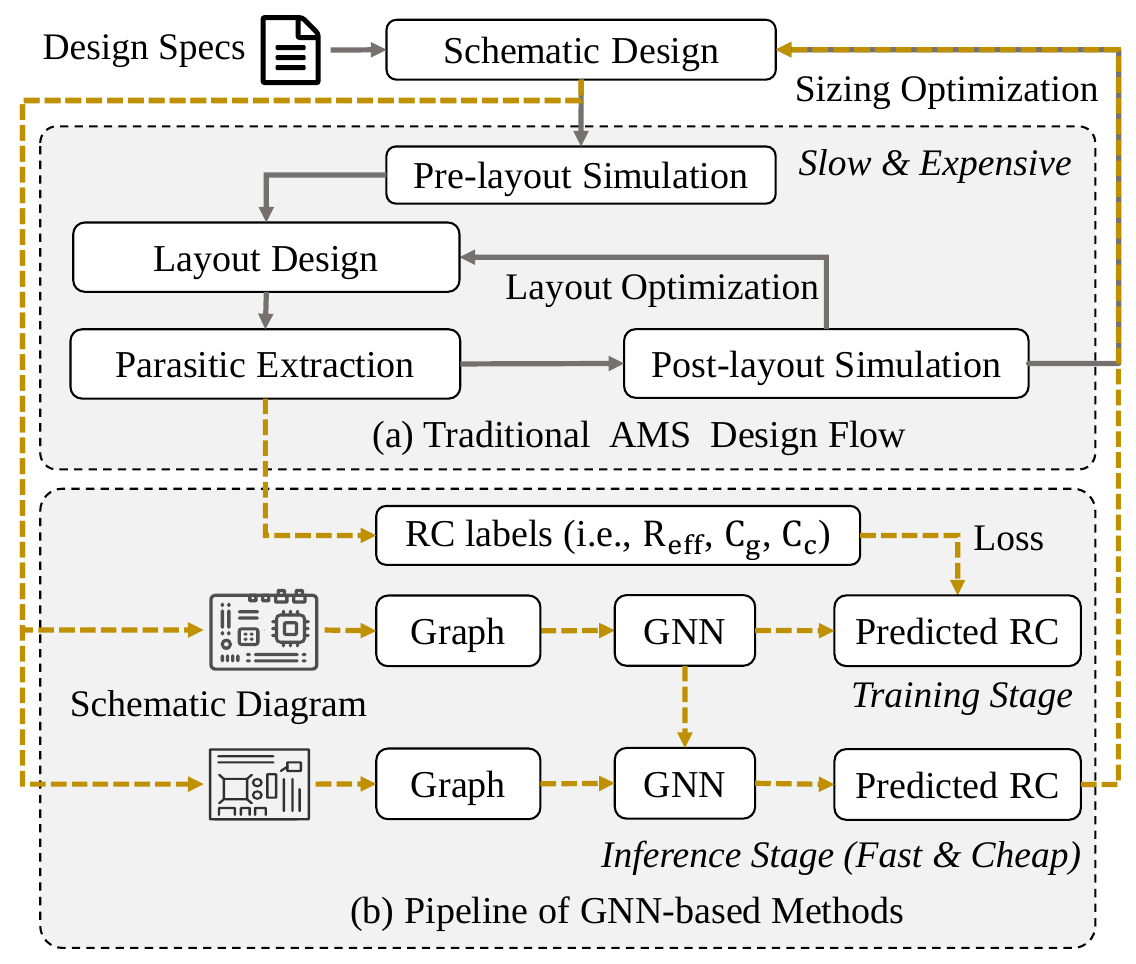}
    \caption{Comparison between traditional AMS design flow and parasitic-aware flow integrating GNNs. Estimating parasitic effects in the early stage can reduce the number of design iterations and help the design converge.} 
    \label{fig:intro_rc_task}
\end{figure}

\input{tables/benchmark}

As chip manufacturing advances toward process nodes of 7nm and below, and as 3D integrated-circuit (IC) technologies continue to emerge~\cite{pentapati2021ml}, parasitic effects, i.e., interconnect resistance ($R$) and capacitance ($C$), have shifted from a secondary consideration to a key bottleneck that constrains design convergence. At these process nodes, the interconnect latency has significantly exceeded the intrinsic latency of the device. For example, in a 5nm node, it accounts for even 70\%–80\% of the total delay of the path \cite{siddiqui2021composite}. 
These effects are particularly severe in analog/mixed-signal (AMS) circuits and memory circuits, where weaker device drive capability at advanced nodes, together with increasingly complex multi-layer metal stacks and abundant vias, further amplifies interconnect RC effects. As a result, parasitics no longer merely perturb circuit behavior, but significantly increase design iterations and make front-end/back-end design convergence substantially more difficult.

As shown in \Cref {fig:intro_rc_task}~(a), the typical design process for analog circuits and memory circuits (e.g., static random access memory, SRAM) starts with schematic design and then moves on to the time-consuming and laborious layout phase. Unlike digital circuit design, the layout design of analog and memory circuits is mostly done manually, which often takes months. Crucially, the impact of parasitic resistance and capacitance (RC) can only be accurately captured through post-layout extraction. This dependency creates a significant design bottleneck: if post-layout simulation fails to meet performance metrics due to unpredictable parasitic effects, designers have to revert to the layout design stage or even restart the schematic design work, resulting in costly and time-consuming iterations. Therefore, the core goal of early parasitic prediction is not simply to accelerate the simulation process or deal with the complexity of large-scale networks, 
but to provide actionable parasitic guidance before layout is finalized. For front-end designers, such guidance can help estimate parasitics on critical paths and support decisions on driver sizing, buffer insertion, and device dimension tuning. For back-end designers, it can help identify critical nets, guide floorplanning (e.g., symmetry-aware placement requirements), and assist routing decisions such as wire width allocation and metal layer selection.
By introducing parasitic parameter-aware design concepts early in the design process, this technology can effectively reduce the number of design iterations and greatly shorten the overall chip development cycle \cite {ren2020paragraph, shen2025cirgps}.

In recent years, graph-based learning, especially graph neural networks (GNNs), has emerged as a promising direction for RC parasitic modeling~\cite{ren2020paragraph,shen2025cirgps,shen2025circuitgcl}, as illustrated in~\Cref{fig:intro_rc_task}(b). Existing studies span schematic-based parasitic prediction, layout-aware circuit representation learning~\cite{zhu2022tag}, and digital circuit RC prediction~\cite{yoon2025paraformer}, indicating growing interest in parasitic-aware learning across different circuit domains and design stages. These efforts also suggest that early parasitic prediction can benefit downstream optimization and analysis tasks, such as parasitic aware design tuning and accelerated post-layout simulation~\cite{liu2021parasiticaware,shen2025openyield}. We defer a more detailed discussion of these methods to Section~\ref{sec:rcgnns}.

Despite this progress, research in this direction remains limited by the lack of a unified benchmark system, and there is still no high-quality public dataset or widely recognized evaluation standard comparable to the Open Graph Benchmark (OGB)~\cite{hu2020ogb}. Due to intellectual-property restrictions and the high cost of industrial-grade IC design, open-source research often relies on outdated or small-scale cases, forming isolated data islands that hinder objective comparison across algorithms and limit the applicability of existing methods to real engineering scenarios.

Beyond its value for EDA research, ParasGB also provides a challenging testbed for graph learning, with large-scale, heterogeneous, and physically grounded circuit graphs derived from silicon-proven designs. It can support future research on more effective and targeted graph foundation models for circuit-specific tasks. Our core contributions are summarized as follows:
\begin{itemize}[leftmargin=*, itemsep=-1pt]
\item This paper publishes a high-quality dataset derived from the AMS circuits designed with commercial EDA tools, providing gold standard parasitics, and laying a reliable foundation for related research.
\item We construct a comprehensive benchmark suite for pre-layout parasitic prediction, covering node-level ground capacitance prediction and edge-level effective resistance and coupling capacitance prediction under both regression and classification settings.
\item This paper develops an easy-to-use research framework that references the OGB design style and integrates data loading and automatic scoring functions, so that researchers familiar with PyTorch Geometric can quickly get started with experiments.
\end{itemize}

\section{Related Work}
In this section, we review prior work on EDA benchmarks and GNN-based parasitic estimation. We first discuss existing machine-learning benchmarks in EDA, and then summarize representative graph-learning methods for parasitic prediction and related tasks.

\input{tables/analog_statistics}
\input{tables/sram_statistics}
\subsection{Machine Learning for EDA Benchmarks}

In recent years, machine learning (ML) has been widely adopted in electronic design automation (EDA), and standardized benchmarks have become increasingly important for ensuring fair comparison and reproducible research. However, most existing benchmarks remain confined to specific design stages and circuit types, as summarized in \Cref{tab:related_works_comparison}. For digital circuits, CircuitNet~\cite{chai2022circuitnet,jiang2024circuitnet2} supports large-scale evaluation on representative tasks such as congestion prediction and timing prediction, while EDA schema~\cite{shrestha2024eda} proposes a unified graph data model to standardize circuit representations across design stages. In contrast, analog circuits follow substantially different design flows and optimization objectives, which motivates dedicated benchmarks in this domain. Early efforts such as CktGNN~\cite{dong2023cktgnn} focus on schematic-level topology generation and device-parameter tuning, while AMSNet~\cite{tao2024amsnet,shi2025amsnet} and AnalogGenie~\cite{gao2025analoggenie} support analog circuit generation and understanding at the netlist level. Nevertheless, these benchmarks predominantly remain at the schematic level and do not provide post-layout parasitic RC supervision for analog and memory circuits. This gap motivates ParasGB as a standardized benchmark for parasitic prediction on physically grounded circuit graphs.

\subsection{Parasitic Estimation via GNNs}\label{sec:rcgnns}
Parasitic parameter estimation is a critical task in EDA, especially at advanced technology nodes where interconnect effects strongly influence circuit performance. Since circuits naturally form graph-structured data, GNNs have become a promising tool for learning the relationship between circuit topology, device features, and parasitic behavior. Compared with rule-based estimation or heuristic methods, GNN-based approaches provide a data-driven alternative for capturing complex parasitic patterns across different circuit structures.

ParaGraph~\cite{ren2020paragraph} pioneered the use of GNNs for parasitic-aware circuit learning by predicting layout parasitics and device parameters directly from circuit graphs. More recent work has extended this direction to different settings and targets. CircuitGCL~\cite{shen2025circuitgcl} improves representation learning for parasitic prediction through contrastive learning, while CircuitGPS~\cite{shen2025cirgps} addresses coupling capacitance prediction in few-shot scenarios. In the digital domain, ParaFormer~\cite{yoon2025paraformer} combines heterogeneous GNNs with transformer-style modeling for pre-routing parasitic RC prediction, highlighting the importance of jointly modeling circuit topology and geometric relationships. Related work, such as TAG~\cite{zhu2022tag} further suggests that layout-aware circuit representations can also benefit parasitic-related downstream tasks. Beyond direct parasitic prediction, predicted parasitic quantities have also been used in downstream tasks such as yield evaluation and parasitic-aware optimization, further demonstrating their practical value~\cite{shen2025openyield,liu2021parasiticaware}. Despite this progress, existing studies remain fragmented in datasets, task definitions, and evaluation protocols, making fair and reproducible comparison difficult. This motivates the development of ParasGB, which provides a unified benchmark for pre-layout parasitic prediction with standardized datasets, tasks, and evaluators.

\section{The ParasGB Benchmark}

In this section, we introduce ParasGB as a benchmark derived from real industrial circuit design flows. We summarize its data characteristics, describe its graph construction principles, and present the evaluation setting for fair and reproducible comparison.

\subsection{Dataset Statistics}
\begin{figure*}[htbp]
    \centering
    \includegraphics[width=\linewidth]{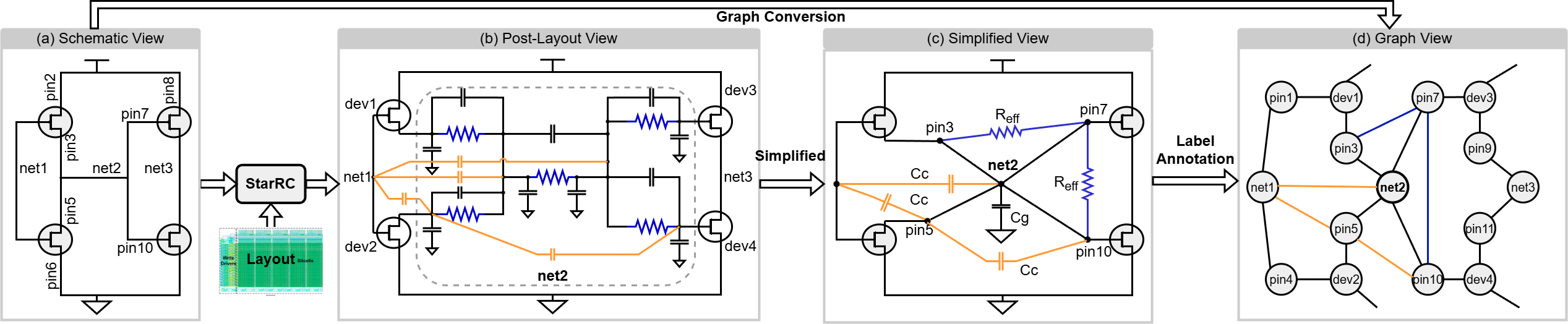}
    \caption{SRAM circuit graph conversion: (a) The schematic. (b) Post-layout netlist with extracted RCs by StarRC. (c) Our simplified netlist with effective RCs. (d) The converted circuit graph. }
    \label{fig:sram_to_graph}
\end{figure*}

Dataset scale and label distribution are fundamental to benchmarking realistic parasitic prediction workloads. ParasGB therefore includes both analog circuits and SRAM circuits, covering a broad range of graph sizes and circuit characteristics. Following the number of net nodes, we group analog designs into three scales (XS, S, and M) and SRAM designs into three scales (L, XL, and XXL), since net count largely determines the size of the circuit matrix and thus the computational cost of simulation~\cite{kapre2009parallelizing}. The overall statistics are summarized in \Cref{tab:analog_statistics} and \Cref{tab:sram_statistics}, while detailed label histograms for all 20 analog circuits and 6 SRAM subsets are provided in the repository.
(details in \Cref{app:dataset})
The analog subset spans from micro-scale blocks to more complex modules. The smallest circuit contains only 231 nodes, while the largest ones reach 6917 nodes. As reported in \Cref{tab:analog_statistics}, these designs also vary substantially in the number of components, pins, nets, and parasitic labels. Although the graph sizes are relatively small compared with SRAM circuits, their parasitic values are extremely fine-grained: most capacitances fall in the range of $10^{-14}\sim10^{-13}\,\text{F}$. Such tiny magnitudes place strict requirements on the numerical accuracy, stability, and robustness of prediction models.

In contrast, the SRAM subset reaches industrial scale and introduces graphs that are orders of magnitude larger. Small SRAM designs such as \textit{digtime} contain about $1.8\times10^{4}$ nodes, while larger designs such as \textit{sandwich} contain up to $1.17\times10^{7}$ nodes and $5.52\times10^{7}$ edges. As shown in \Cref{tab:sram_statistics}, SRAM parasitic graphs are also highly dense, with coupling-capacitor edges dominating the edge set in large designs. This makes SRAM subsets particularly challenging not only in terms of graph scale, but also in terms of memory footprint and computational efficiency.

Taken together, the analog and SRAM subsets form a representative cross-scale benchmark. Analog circuits emphasize fine-grained parasitic variation and prediction precision, while SRAM circuits stress scalability on ultra-large and densely connected graphs. This combination makes ParasGB suitable for evaluating both accuracy and efficiency, providing a more comprehensive benchmark than prior datasets limited to a single scale or circuit domain.

\subsection{Circuit Sourcing and Validation}


All data in ParasGB are derived from real chip design flows and generated with industry-standard commercial EDA toolchains. For the analog subset, the designs are implemented in a 180nm BCD process. In the front-end stage, schematic capture and pre-layout simulation are performed using Cadence Virtuoso together with Spectre. In the back-end stage, physical layouts are created in Virtuoso, followed by DRC/LVS verification and high-accuracy parasitic extraction using Siemens Calibre xRC. The analog subset covers representative AMS modules, including LVDS (Low-Voltage Differential Signaling), BGR (Band Gap Reference), OP (Operational Amplifier), and LDO (Low Dropout Regulator).

For the SRAM subset, the design flow follows a typical memory implementation
and signoff process under the TSMC 28nm process. Starting from verified circuit
designs and completed layouts, circuit-level functional and timing verification
is performed using HSPICE and FineSim at both the pre-layout and post-layout stages. Post-layout parasitics are then extracted using Synopsys StarRC, a silicon-accurate signoff parasitic extraction tool that is widely used for memory, analog/mixed-signal, and custom digital designs. StarRC is designed to support large memory blocks and advanced process effects while preserving signoff-level accuracy, making it well-suited for generating realistic parasitic annotations for SRAM circuits.

Moreover, all circuit designs in ParasGB are based on real tape-out-proven projects rather than purely theoretical or simulation-only examples, which gives the dataset strong physical fidelity and practical relevance. Under advanced process nodes, layouts become increasingly compact and metal spacing is reduced, which intensifies parasitic coupling and increases the structural complexity of parasitic networks. This is a central challenge in parasitic prediction and one of the key scenarios targeted by ParasGB. By extracting data directly from tape-out-proven designs, ParasGB narrows the gap between idealized academic benchmarks and real industrial conditions, and helps ensure that learned representations better reflect underlying physical mechanisms of circuit parasitics rather than artifacts of overly simplified models. More detailed information about the dataset is provided in our repository.

\input{tables/task_overview}
\input{tables/analog_features}
\subsection{Topology-to-Graph Conversion}
The transformation from circuit topology to graph representation follows the same overall framework for both SRAM and analog circuits, while the specific supervision labels depend on the task setup of each circuit family. \Cref{fig:sram_to_graph} illustrates this process using an SRAM example(analog process is in the \Cref{app:analog2graph}). Starting from the schematic and layout, commercial tools generate a post-layout netlist with fully extracted parasitics, which is then simplified and converted into a heterogeneous graph $\mathcal{G} = (\mathcal{V}, \mathcal{E})$. The node set $\mathcal{V}$ contains three types: device nodes representing circuit components (e.g., MOSFETs and resistors), net nodes representing interconnect wires, and pin nodes representing device terminals. Topological edges $\mathcal{E}_{\text{topo}}$ (black lines in the figure) capture schematic connectivity through \textit{device-to-pin} and \textit{pin-to-net} relations and are shared by both SRAM and analog circuits. Among parasitic labels, blue edges denote effective resistance ($R_{eff}$), which is used in both circuit families, whereas yellow edges denote coupling capacitance ($C_c$), which is included only in the SRAM benchmark. For completeness, we provide an analog-specific graph construction pipeline in the repository.

Converting post-layout netlists into lumped models is necessary because raw parasitic extraction results are prohibitively complex. As shown in \Cref{fig:sram_to_graph}~(b), even a single logical net can expand into a large distributed RC network with a huge number of parasitic components, making direct learning intractable. We therefore simplify each distributed network into a lumped circuit model that defines our supervision signals. Specifically, \textit{net nodes} are labeled with lumped ground capacitance ($C_g$), while labeled edges $\mathcal{E}_{\text{label}}$ comprise effective resistance ($R_{eff}$), computed for sampled pin pairs within the same resistive network using a matrix-based method derived from the node admittance matrix and the inverse of its Cholesky factor~\cite{liu2023computing} (The detailed algorithm is in \Cref{app:algorithm}). In addition, for SRAM circuits, coupling edges ($C_c$) aggregate the total coupling capacitance between corresponding net pairs. The ground and coupling capacitances mentioned above both originate from unintended electric-field interactions in layout structures: ground capacitance is formed between interconnects and the substrate, whereas coupling capacitance is formed between adjacent interconnect segments. This reduction preserves the key timing- and noise-related characteristics of the original RC network while making graph-based learning on large industrial designs computationally feasible.

\input{tables/overall_label_distribution}
\subsection{Node Feature Engineering}





The core objective of feature engineering is to turn the schematic design into learnable numerical representations that approximate unobserved layout-dependent parasitics. Statistics extracted from schematic netlists act as high-level descriptors of the underlying physical topology. For instance, the number and type of devices attached to a net (node degree) serve as proxies for interconnect length and via count: higher connectivity typically implies longer wiring, more contacts, and thus larger parasitic resistance and capacitance. Likewise, device geometric parameters such as transistor width ($W$), length ($L$), and multiplier ($M_{\text{mos}}$) directly constrain the size of active regions and poly gates. Increasing $W$ or $M_{\text{mos}}$ raises intrinsic gate and diffusion capacitance, and also forces wider and denser metal routing, which introduces additional parasitic resistance and capacitance in the metal interconnect. These relationships establish a meaningful predictive correlation between schematic-level attributes and post-layout RC values.

Guided by this intuition, we encode analog circuits using a compact hierarchical feature set at the device, net, and pin levels (see~\Cref{tab:analog_node_features}). Device features combine type indicators ($D_{\text{mos}}$, $D_{\text{res}}$, $D_{\text{cap}}$) with key geometric and process parameters ($W$, $L$, $M_{\text{mos}}$, $N_f$, $T$). Net features include functional flags ($F_{\text{pwr}}$, $F_{\text{gnd}}$, $F_{\text{sig}}$, $F_{\text{io}}$) and connectivity statistics ($N_{\text{mos}}$, $N_g$, $N_{sd}$, $N_b$, $L_{\text{tot}}$, $W_{\text{tot}}$) summarizing how many devices and terminals a net drives and the total channel dimensions it aggregates. Pin features ($P_d$, $P_g$, $P_s$, $P_b$, $P_{\text{cap}}$, $P_{\text{res}}$) indicate each terminal’s electrical role and local loading context. SRAM circuits share the same hierarchical feature-design principle and high-level node-feature categories as analog circuits, namely device-, net-, and pin-level features. However, because the underlying circuit structures are different, some feature dimensions have slightly different physical meanings and are adjusted to better reflect memory-netlist characteristics. For example, pin features in analog circuits mainly characterize connection roles and electrical functions, whereas pin features in SRAM circuits primarily encode MOS terminal types such as G/D/S/B (details in Appendix~\ref{app:dataset}).

\subsection{Benchmarking Baselines}

\textbf{Basic GNN Baselines.}
In order to verify the task challenge of the ParasGB dataset, this paper selects the classical model in the graph learning field as the basic benchmark. GCN~\cite{kipf2017gcn}, GAT~\cite{velivckovic2018gat}, GraphSAGE~\cite{hamilton2017graphsage}, and PNA~\cite{corso2020pna} are the mainstream methods in the field of GNNs, which learn the topological features of circuits through the message passing mechanism between neighbor nodes. In addition, this paper also incorporates graph transformer models such as SGFormer~\cite{wu2023sgformer} and PolyNormer~\cite{deng2024polynormer}. In addition to capturing the local features of directly adjacent nodes, they can also model the long-range dependencies through the global attention mechanism. 
\textbf{RC-Specific Baselines.}
In addition to general-purpose GNNs, we also include several circuit-specific models designed for parasitic-related tasks, including CircuitGPS~\cite{shen2025cirgps} and CircuitGCL~\cite{shen2025circuitgcl} (discussed in \Cref{sec:rcgnns}).

\section{Problem Formulation and Data Analysis}
\subsection{Multi-level Predictive Tasks}

Based on the dataset construction described above, ParasGB further defines diverse circuit parasitic prediction tasks with different granularities, target types, and learning objectives, as summarized in \Cref{tab:datasets_overview}.

\textbf{Granularity Variety.} The benchmark includes both \textit{node-level} and \textit{edge-level} tasks, each corresponding to distinct physical phenomena. Node-level tasks predict ground capacitance ($C_g$), a key determinant of circuit timing and power consumption. Edge-level tasks target coupling capacitance ($C_c$) and effective resistance ($R_{\text{eff}}$), where $C_c$ is critical for analyzing signal cross-talk and noise, and $R_{eff}$ directly impacts signal delay (timing) and voltage drop (IR drop) in power distribution networks.

\textbf{Objective Variety.} We formulate both regression and classification tasks to support different use cases. Regression targets fine-grained numerical estimation, which is useful for parasitic-aware analysis and more detailed early-stage evaluation. Classification discretizes parasitic values into magnitude bins, which is useful for rapid screening and early-stage design exploration when identifying the order of magnitude is already informative. In addition, coarse-grained categorization of parasitic values may also serve as a useful signal for downstream simplification or partitioning-oriented decisions, analogous to effective resistance-based criteria used in graph reduction and sparsification\cite{zhao2014spectral, liu2023effective}.

\begin{lstlisting}[
style=iccadpy,
float=!b,
caption={Unified ParasGB API for loading and evaluation.},
label={lst:rcg_unified_usage}
]
from parasgb import RCDataset, Evaluator
# 1) Load dataset
dataset = RCDataset(
    dataset_name='sram',      # or 'analog'
    task_level='node',        # 'node' or 'edge'
    task_type='regression'    # or 'classification'
)
# 2) Build dataloader
loader = dataset.get_dataloader(
    split='train', batch_size=32, shuffle=True
)
# 3) Train model
model = MyModel()
for batch in loader:
    y_pred = model(batch)
    loss = criterion(y_pred, batch.y)
    loss.backward()
    optimizer.step()
    optimizer.zero_grad()
# 4) Evaluate
evaluator = Evaluator(
    dataset_name='sram',
    task='cg_regr'            # e.g., cg_regr, cc_class, r_regr
)
metrics = evaluator.evaluate(y_pred, y_true)
\end{lstlisting}

\input{tables/sram_coupling_classification}

\input{tables/sram_ground_classification}
\subsection{Distribution Analysis and Preprocessing}

Label imbalance is a core challenge in industrial circuit data processing. As shown in~\Cref{fig:overall_label_distribution}, after applying the task-specific normalization described above, the capacitance and resistance labels still exhibit highly skewed, long-tailed, and in some cases multimodal distributions. In several tasks, most samples cluster within a narrow value range, while only a small fraction of nodes or edges lie in the extreme tails. Such distributions make training particularly difficult, since conventional loss functions tend to fit the dominant modes more easily while underemphasizing rare but critical samples, even though these extremes often have a disproportionate impact on timing, noise, and IR-drop behavior. The specific label distribution can be found in \Cref{app:dataset}.



To mitigate these distribution challenges, we adopt a consistent preprocessing framework with task-specific filtering and normalization. Circuit parasitic parameters typically span multiple orders of magnitude, and their numerical scales differ substantially from those of spatial coordinates and device geometries; directly using the raw values can therefore lead to unstable optimization~\cite{FeyLenssen2019PyG}. Before normalization, we remove the sparse physically invalid or numerically unstable labels, including out-of-range capacitance values and zero-valued effective resistances. For SRAM circuits, capacitance labels are transformed using logarithmic scaling, while effective-resistance labels are first mapped to the logarithmic domain and then normalized using the first and ninety-ninth percentiles. For analog circuits, ground-capacitance and effective-resistance labels are normalized through logarithmic maximum-based transformations. All processed labels are clipped to the $[0,1]$ interval. Regression tasks directly use these normalized continuous values as supervision targets, whereas classification tasks discretize them into five magnitude classes using the predefined boundaries $\{0.2,0.4,0.6,0.8\}$ (details in Appendix~\ref{app:normalization}).



\section{Experiments}

Our implementation is built on PyTorch Geometric (PyG) for graph-based processing~\cite{FeyLenssen2019PyG}. All experiments were conducted on two shared computing clusters equipped with Intel Xeon Silver 4314 CPUs (2.4 GHz), 128 GB system memory, four NVIDIA RTX 4090 GPUs (24 GB VRAM), and three NVIDIA A100 GPUs (40 GB VRAM).

In addition, to ensure reproducibility and lower the barrier to adoption, we provide in the repository the hyperparameter settings, training strategies, and hardware and cluster configurations for all benchmarked models. The ParasGB toolkit follows an OGB-style benchmark interface~\cite{hu2020ogb} and is built upon the PyG framework~\cite{FeyLenssen2019PyG}, enabling concise and unified access to datasets and evaluation utilities. As illustrated in Listing~\ref{lst:rcg_unified_usage}, the toolkit supports the full workflow from data download and preprocessing to standardized evaluation. For large-scale SRAM graphs, it also provides memory-efficient subgraph sampling through \texttt{NeighborLoader}. To ensure fair comparison across methods, we recommend using the unified Evaluator module~\cite{hu2020ogb}, which automatically handles task-specific metrics. Additional results are available in the repository.

\input{tables/analog_effective_regression}

\input{tables/analog_ground_regression}
\subsection{Performance on SRAM Tasks}
We further evaluate all methods on SRAM parasitic classification tasks, including edge-level coupling capacitance classification and node-level ground capacitance classification, as shown in Table~\ref{tab:sram_coupling_classification} and Table~\ref{tab:sram_ground_classification}. The regression experiment is in the Appendix~\ref{app:sram_regression}.

The results highlight the effectiveness of circuit-specific models designed for parasitic-related tasks. In particular, for edge-level coupling capacitance classification, CircuitGCL consistently achieves the best results on all four SRAM subsets, obtaining the highest Accuracy and F1-score in every case. This indicates that SRAM coupling-capacitance prediction benefits substantially from circuit-specific inductive bias and from modeling dense local coupling structure.
For node-level ground capacitance classification, the results are more heterogeneous across subsets, but circuit-specific modeling remains highly competitive. PolyNormer performs best on the aggregated subset, while CircuitGCL achieves the best results on \textit{sandwich} and \textit{array\_128\_32\_8t}, and SGFormer performs best on \textit{ultra8t}. This suggests that node-level SRAM parasitic patterns vary considerably across designs, and that different models capture different aspects of the task.

Another notable observation is the persistent gap between Accuracy and F1-score, especially for node-level ground capacitance classification. In several subsets, some methods achieve moderate or even relatively high Accuracy, while their F1-scores remain much lower, indicating that label imbalance and long-tail distributions remain major challenges. Overall, these results show that ParasGB provides a challenging benchmark for SRAM parasitic prediction and clearly reveals the performance gains brought by circuit-specific parasitic-related models.

\subsection{Performance on Analog Tasks}

We further evaluate all methods on analog-circuit parasitic regression tasks, including edge-level effective resistance ($R_{\mathrm{eff}}$) prediction and node-level ground capacitance ($C_g$) prediction, as shown in Table~\ref{tab:analog_resistance_regression} and Table~\ref{tab:analog_ground_regression}. The classification experiment is in the Appendix~\ref{app:analog_classification}.

Overall, $C_g$ node regression is easier and more stable than $R_{\mathrm{eff}}$ edge regression. On the main analog subset (IDs 1--4, 6, 8--12, 15--18), several models achieve strong $C_g$ performance, whereas $R_{\mathrm{eff}}$ prediction remains much more challenging, especially on circuit ID 14, where GCN, GAT, and GraphSAGE still yield negative $R^2$. This indicates that effective-resistance prediction is more sensitive to circuit-specific structural variation.
The results also highlight the value of circuit-specific models designed for parasitic-related tasks. On $R_{\mathrm{eff}}$, PNA achieves the best $R^2$ on the main subset and on ID 5, while CircuitGCL achieves the best $R^2$ on IDs 14 and 20. Notably, on the hardest circuit, ID 14, both CircuitGPS and CircuitGCL improve the negative $R^2$ values observed for several general-purpose GNNs to clearly positive ones, demonstrating clear gains on particularly difficult cases. On $C_g$, the task is more stable overall: PolyNormer achieves the best $R^2$ on the main subset, while CircuitGCL performs best on IDs 5 and 20.

Another notable observation is the large variation across individual analog circuits. Model rankings change substantially on IDs 5, 14, and 20, showing that the challenge lies not only in graph size, but also in strong cross-circuit heterogeneity in topology and parasitic patterns. Overall, these results show that ParasGB provides a challenging benchmark for analog parasitic prediction and clearly reveals the benefits of circuit-specific parasitic-related models, especially on hard and poorly predicted cases.

\section{Conclusions \& Limitations}
This paper presents ParasGB, the first open-source benchmark suite for parasitic parameter prediction on circuit graphs. Built from tape-out-proven designs and RC networks extracted by commercial EDA tools, ParasGB provides a physically grounded and standardized platform for pre-layout parasitic modeling. By covering both node-level and edge-level tasks under regression and classification settings, it fills the long-standing gap in public, high-fidelity parasitic benchmarks for reproducible research.
Using a unified evaluation protocol and standardized training pipeline, we benchmark diverse GNN architectures and reveal several key challenges of parasitic prediction, including severe label imbalance, long-tailed distributions, and strong structural heterogeneity across circuits. These results show that ParasGB supports not only fair comparison but also the study of model robustness and generalization in parasitic-aware circuit learning.

ParasGB also has several limitations. The current benchmark covers only a limited set of process nodes and analog circuit families. Moreover, as a pre-layout benchmark, it does not directly model detailed placement and routing geometry, nor does it aim to replace full post-layout extraction or signoff flows. Future extensions may broaden process-node coverage, include more diverse and larger analog designs, and further connect parasitic prediction with downstream circuit-analysis tasks.
\bibliographystyle{ACM-Reference-Format}
\bibliography{refs}

\newpage
\appendix
\onecolumn

\section*{Appendix}
\section{Dataset Details}
\label{app:dataset}

\subsection{SRAM Features}

\input{tables/sram_features}
For SRAM circuits, we use a statistical aggregated feature set (see~\Cref{tab:sram_node_features} for details), focusing on the statistical characteristics of the global topology and device distribution \cite{shrestha2024eda}. Specific characteristics include: the number of transistors, resistors, capacitors and their multipliers; the number of gates, source drains, and substrates connected to each network cable; and the total width and length of all connectors. These kinds of aggregated features can effectively reduce the feature dimension of ultra-large-scale SRAM arrays, so that the model can robustly capture the global physical characteristics of the circuit at the scale of tens of millions of nodes.

\subsection{Datasets Introduction}

The benchmarks in this study cover two core areas: SRAM and analog circuits, and the specific characteristics of each subset are as follows.

The core characteristics of SRAM datasets are large-scale and dense topologies. With advanced process technology, the wire spacing inside the storage array is extremely small, resulting in a high density of coupling capacitors. The relevant data is extracted from the real circuit netlist (SPF file) after layout. Given that the node size of SRAM circuits generally reaches more than one million, this dataset can effectively verify the scalability and computational efficiency of the model on large-scale graph structures.
\begin{itemize}[leftmargin=*, itemsep=0pt, topsep=0pt, parsep=0pt, partopsep=0pt]
\item \textit{\textbf{ssram:}} is an energy-efficient SRAM-based design composed of digital standard cells and SRAM arrays. It is used as a representative small-to-medium-scale memory design for training and validation \cite{tscache}.
\item \textit{\textbf{digtime:}} corresponds to a digital/SRAM hybrid clock-generation module for SRAMs. It contains digital cells and SRAM columns and is used to evaluate model generalization on internal SRAM clock-generation logic \cite{tscache}.
\item \textit{\textbf{timing\_ctrl:}} is a timing-control module composed of standard digital cells that generate control signals for SRAM operations. It provides a digital-control test case with circuit topology different from the training designs.
\item \textit{\textbf{sandwich:}} is a sandwich-like memory design in which computational digital circuits and SRAM arrays are organized in an alternating structure, forming a representative large-scale mixed digital-memory design \cite{cim1}.
\item \textit{\textbf{ultra8t:}} is a multi-voltage 8T SRAM design containing large analog modules and SRAM arrays. It is used to evaluate the scalability and transferability of graph models on large memory-oriented AMS designs \cite{shen2024ultra8t}.
\item \textit{\textbf{array\_128\_32\_8t:}} is a standalone 128-row 32-column 8T SRAM array. It serves as an array-level memory benchmark for evaluating model performance on regular SRAM-array structures.
\end{itemize}



Unlike SRAM datasets, analog circuits are smaller but have extremely detailed depictions of physical characteristics at the device level. Key parameters such as device channel width $W$, length $L$, and distance from the source-drain area to the edge of the isolation tank (LDE effect) are included in the node feature system. The circuit parasitic parameters are extracted through commercial PEX tools. Since the analog circuit is highly sensitive to noise, even a small parasitic parameter prediction error can cause the circuit simulation results to deviate from the expected \cite{yu2025deep}.
\begin{itemize}[leftmargin=*, itemsep=0pt, topsep=0pt, parsep=0pt, partopsep=0pt]

\item \textit{\textbf{ID 1:}} Low-Voltage Differential Signaling Circuit, It converts a low-frequency reference (5-27 MHz) into a high-frequency clock (100-700 MHz) with minimal phase noise for precision timing applications \cite{leung2003cmos}.
\item \textit{\textbf{ID 2:}} Operational Amplifier Circuit, it utilizes a quasi-linear temperature characteristic to generate a stable 0.4V reference voltage from an ultra-low supply of only 0.56V, consuming just 4.8 $\mu$A of current \cite{4098517}.
\item \textit{\textbf{ID 3:}} Band-Gap Reference Circuit, it generates a stable reference voltage (near the silicon bandgap) by utilizing self-cascode composite transistors and a single resistor, achieving a low temperature coefficient of 25.3 ppm/°C with only $25\text{ }\mu\text{A}$ of current \cite{6180360}. 
\item \textit{\textbf{ID 4:}} Band-Gap Reference Circuit, it utilizes resistor-subdivision and resistor-less methods to generate a highly stable 910.88 mV reference voltage with an ultra-low temperature coefficient of 12.99 ppm/°C \cite{7072988}.
\item \textit{\textbf{ID 5:}} Low-Dropout Regulator Circuit, it generates a stable reference voltage for LDO regulators by balancing the temperature characteristics of N/P-type MOSFETs, achieving a temperature coefficient of 36.9 ppm/°C with a low supply current of 9.7 $\mu$A \cite{1158794}.
\item \textit{\textbf{ID 6:}} Low-Dropout Regulator Circuit, it provides a stable output from a 1.8-4.5V supply with a fast transient response and minimal compensation capacitance ($7\text{ pF}$), supporting up to $100\text{ mA}$ of load current with a low dropout of $0.2\text{ V}$ \cite{5395671}.
\item \textit{\textbf{ID 7:}} Low-Dropout Regulator Circuit, it generates a stable reference voltage for LDO regulators by balancing the temperature characteristics of N/P-type MOSFETs, achieving a temperature coefficient of 36.9 ppm/°C with a low supply current of 9.7 $\mu$A \cite{1158794}.
\item \textit{\textbf{ID 8:}} Low-Dropout Regulator Circuit, it utilizes an Adaptive-Compensation Buffer (ACB) to dynamically switch between pass transistors, enabling stable operation across a wide load range ($0$ to $30\text{ mA}$) without an external capacitor while maintaining a low quiescent current of $6\text{ }\mu\text{A}$ \cite{10870569}.
\item \textit{\textbf{ID 9:}} Low-Dropout Regulator Circuit, it utilizes a tri-loop architecture to achieve an ultra-fast transient response (1.15 ns) and maintains a clean supply with a full-spectrum power supply rejection ($\mathrm{PSR} > -12~\mathrm{dB}$ up to 20 GHz) while consuming only $50\text{ }\mu\text{A}$ of quiescent current \cite{7001720}.
\item \textit{\textbf{ID 10:}} Operational Amplifier Circuit, it provides a flexible, step-by-step methodology to balance noise performance against power consumption, offering greater design control than previous methods as verified by multi-condition SPICE simulations \cite{1495717}.
\item \textit{\textbf{ID 11:}} Low-Dropout Regulator Circuit, it utilizes two parallel active feedback paths to create two pole-zero pairs, providing superior stability and transient response compared to single-path methods while supporting a 100-mA load with only $14\text{ }\mu\text{A}$ of quiescent current \cite{8688481}.
\item \textit{\textbf{ID 12:}} Low-Dropout Regulator Circuit, it utilizes a high-gain, three-stage error amplifier to maintain precise regulation even with an unsaturated pass transistor at ultra-low voltages (0.5 V supply), achieving a state-of-the-art current density of 11.4 A/mm$^2$ and a low-frequency PSR of -62 dB \cite{9964454}.
\item \textit{\textbf{ID 13:}} Operational Amplifier Circuit, it replaces conventional Miller compensation with an active structure that eliminates the right-half-plane (RHP) zero and introduces a left-half-plane (LHP) zero to cancel the first non-dominant pole, resulting in a 9.4x increase in unity-gain frequency and a significantly smaller compensation capacitor \cite{5967452}.
\item \textit{\textbf{ID 14:}} Band-Gap Reference Circuit, it utilizes a combination of four MOSFETs, two lateral PNP transistors, and a well resistor to generate a stable 16 $\mu$A output current with a temperature coefficient of 105 ppm/°C, requiring no external bandgap reference or trimming for process compensation \cite{7763843}.
\item \textit{\textbf{ID 15:}} Low-Dropout Regulator Circuit, it utilizes a damping-zero compensation technique and a slew-rate enhancement circuit to achieve stability and fast transients with only 1.5 pF of on-chip capacitance, supporting a 100 mA load with a 200 mV dropout \cite{5548891}.
\item \textit{\textbf{ID 16:}} Low-Dropout Regulator Circuit, it utilizes a WCF circuit to maintain a fast transient response and stable voltage regulation across a very wide range of load currents (up to 100 mA) and load capacitances (470 pF to 10 nF), while consuming only $14.4\text{ }\mu\text{A}$ \cite{7829748}.
\item \textit{\textbf{ID 17:}} Low-Dropout Regulator Circuit, it utilizes the nested adaptive FVF structure to achieve an ultra-fast transient response (handling load steps from $1\text{ }\mu\text{A}$ to $20\text{ mA}$ in just $10\text{ ps}$) while significantly enhancing PSR ($-58.52\text{ dB}$ at $1\text{ MHz}$) and line regulation \cite{9332114}.
\item \textit{\textbf{ID 18:}} Low-Dropout Regulator Circuit, it provides a stable output with high DC gain (101 dB) and a precision bandgap reference to support high-current loads (up to 450 mA) with a 0.5-V dropout, while maintaining solid power supply rejection (54.5 dB at 100 Hz) \cite{6699480}.
\item \textit{\textbf{ID 19:}} Band-Gap Reference Circuit, it exploits dimension-dependent effects to cancel out process-induced threshold voltage variations, achieving an ultra-low power consumption of 192 pW and highly stable performance (0.53\% process variation) without the need for post-fabrication trimming \cite{8859356}. 
\item \textit{\textbf{ID 20:}} Band-Gap Reference Circuit, it generates a stable reference voltage using an ultra-low power architecture where most of the $5\text{ }\mu\text{A}$ current is dedicated to the output, achieving a temperature coefficient of $<10\text{ ppm/°C}$ from a $1\text{ V}$ supply without requiring a high-area operational amplifier \cite{edward2009low}.

\end{itemize}

For data preparation, full schematic netlists were parsed to construct graph representations and extract node features based on circuit-level statistics, following the preprocessing pipeline described in~\cite{shen2025cirgps}. For SRAM circuits, post-layout netlists in Standard Parasitic Format (SPF) were processed to obtain ground-truth labels for ground capacitance and coupling capacitance. For analog circuits, parasitic extraction (PEX) files were utilized to collect ground-capacitance annotations, while effective resistances were computed directly from the extracted netlists. 

\subsection{Dataset Labels}

\input{tables/analog_effective_each_distribution}

\input{tables/sram_coupling_each_distribution}

\input{tables/sram_resistance_each_distribution}

\input{tables/analog_ground_each_distribution}

\input{tables/sram_ground_each_distribution}

This appendix shows the detailed distribution characteristics of the various labels in the dataset. Clarifying the label distribution characteristics is a key prerequisite for training robustness models, and as mentioned in the text, there is a significant label distribution imbalance in circuit data.

\Cref{fig:analog_ground_each_distribution} shows the distribution characteristics of the 20 analog circuits with ground capacitance ($C_g$) tags. Despite the differences in the functions of these circuits, their label distribution curves are highly similar, and they all show steep long-tail distribution characteristics, that is, the capacitance values of most nodes are concentrated in a narrow range. \Cref{fig:sram_ground_each_distribution} shows the ground capacitance distribution characteristics of the 6 core SRAM circuits, covering typical designs such as array\_128\_32\_8t, Sandwich, etc. The label value span of such large-scale circuits is significantly increased, making the regression task challenging \cite{yu2025deep}.

In addition to capacitance, the label distribution characteristics of effective resistors and coupling capacitors were also analyzed. \Cref{fig:analog_effective_each_distribution} shows the distribution of effective resistors ($R$) in 20 analog circuits, and the results show that the distribution of resistance labels is more sparse than that of capacitors, and the number of outliers is larger. The core reason for this phenomenon is that the calculation of effective resistance relies on a specific current path, and its physical properties are more complex \cite{shrestha2024eda}.

Finally, \Cref{fig:sram_coupling_each_distribution} presents the distribution profile of the coupling capacitance ($C_c$) of the 6 SRAM circuits. The density of wire interconnects is significantly increased under advanced processes, and it can be observed from \Cref{fig:sram_coupling_each_distribution} that the number of coupling capacitors is much higher than that of ground capacitors. Such high-density edge-level labels are the core scenario for verifying the non-local feature capture capability of GNNs \cite{luo2024classic,hu2020ogb}.

\section{Analog Topoology-To-Graph}
\label{app:analog2graph}

\begin{figure*}[h]
    \centering
    \includegraphics[width=\linewidth]{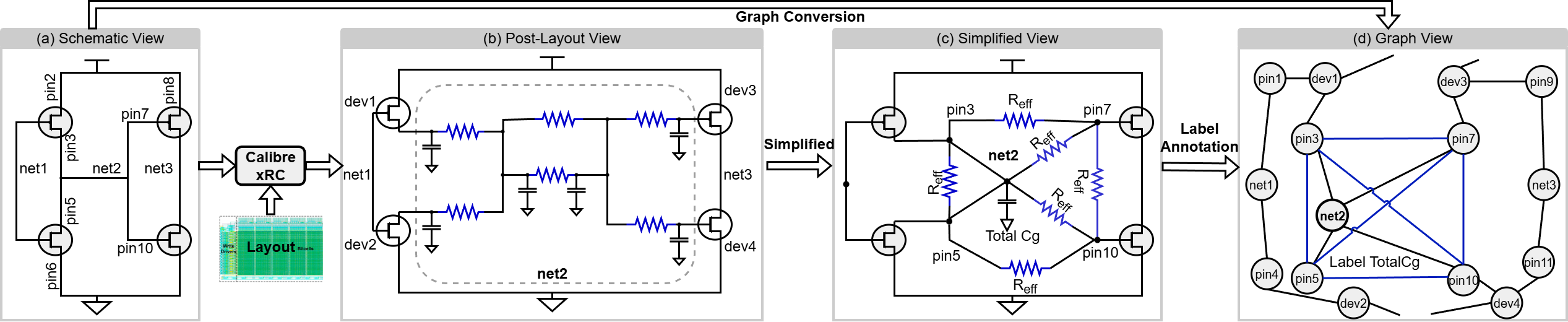}
    \caption{analog circuit graph conversion: (a) The schematic. (b) Post-layout netlist with extracted RCs by StarRC. (c) Our simplified netlist with effective RCs. (d) The converted circuit graph.}
    \label{fig:circuit_to_graph}
\end{figure*}

The conversion from analog circuit schematics to graph representations follows the framework shown in the figure. We model each circuit as a heterogeneous graph $\mathcal{G}=(\mathcal{V},\mathcal{E})$. The node set $\mathcal{V}$ contains three types of nodes: \emph{device nodes} representing circuit components, \emph{net nodes} representing interconnect wires, and \emph{pin nodes} representing device terminals. The topological edges $\mathcal{E}_{\text{topo}}$ (shown as black lines) capture circuit connectivity derived from the schematic, specifically through \emph{device-to-pin} and \emph{pin-to-net} connections; these topological relations constitute the input structure obtained from the schematic-to-graph transformation. 

In contrast, parasitic information is obtained from the extracted parasitic netlist. Blue \emph{pin-to-pin} edges are treated as resistive edges, where the label corresponds to the effective resistance between two pins (details in Appendix~\ref{app:algorithm}). In addition, we assign the total ground capacitance of each net as a node-level label on the corresponding net node. These parasitic labels serve as prediction targets in our benchmark.

\section{Normalization and Label Processing}
\label{app:normalization}

This section provides the detailed normalization and label construction
procedures used in ParasGB. The raw circuit attributes and parasitic
labels have substantially different numerical scales, while the
capacitance and resistance labels may span several orders of magnitude.
We therefore apply task-specific transformations before model training.
All data-dependent normalization statistics are computed once from the
training split and are subsequently fixed for the validation and test
splits. In particular, normalization parameters are not recomputed
separately for individual circuits.

\subsection{Analog Feature Normalization}

For Analog circuits, the continuous attributes of device and net nodes
are normalized separately according to their node types. Let
$x_{v,j}^{(\tau)}$ denote the $j$-th feature of node $v$ with node type
$\tau \in \{\mathrm{device},\mathrm{net}\}$. The maximum value of the
corresponding feature dimension in the training split is defined as

\begin{equation}
m_{\tau,j}
=
\max_{v \in \mathcal{V}_{\mathrm{train}}^{(\tau)}}
x_{v,j}^{(\tau)} .
\end{equation}

The normalized node feature is then calculated as

\begin{equation}
\hat{x}_{v,j}^{(\tau)}
=
\frac{x_{v,j}^{(\tau)}}{m_{\tau,j}+\epsilon},
\end{equation}

where $\epsilon$ is a small positive constant used to avoid division
by zero. The normalization is independently performed for each feature
dimension and each node type.

\subsection{Label Filtering and Normalization}
\label{subsec:label_filtering_normalization}

The raw parasitic labels exhibit wide dynamic ranges and highly
long-tailed distributions. In addition, a very small fraction of labels
have physically abnormal or numerically unstable values. Although these
outliers account for only a negligible portion of the dataset, their
extreme magnitudes can substantially enlarge the label range, distort
the overall distribution, and destabilize model training. We therefore
apply task-specific filtering and normalization before constructing the
regression and classification targets.

\paragraph{SRAM capacitance labels.}

Let $C_g$ denote the ground capacitance of a net node and $C_c$
denote the coupling capacitance of a target edge. A capacitance label is
considered valid only when
\begin{equation}
10^{-21} < C < 10^{-15},
\label{eq:sram_cap_filter}
\end{equation}
where $C$ represents either $C_g$ or $C_c$.

For the edge-level coupling-capacitance task, target edges whose labels
fall outside this range are removed:
\begin{equation}
\mathcal{E}_{c}^{\mathrm{valid}}
=
\left\{
e \in \mathcal{E}_{c}
\mid
10^{-21} < C_c(e) < 10^{-15}
\right\}.
\label{eq:sram_cc_filter}
\end{equation}

For the node-level ground-capacitance task, directly removing an invalid
node would alter the original circuit topology. We therefore preserve
the graph structure and replace its invalid label with a negligible
placeholder value:
\begin{equation}
C_g' =
\begin{cases}
C_g, & 10^{-21} < C_g < 10^{-15},\\
10^{-30}, & \mathrm{otherwise}.
\end{cases}
\label{eq:sram_cg_filter}
\end{equation}

The filtered SRAM capacitance labels are subsequently transformed using
logarithmic normalization. The normalized coupling capacitance is
defined as
\begin{equation}
\hat{C}_{c}^{\mathrm{SRAM}}
=
\operatorname{clip}
\left(
\frac{
\log_{10}\left(C_c \times 10^{21}\right)
}{
6
},
0,
1
\right),
\label{eq:sram_cc_normalization}
\end{equation}
and the normalized ground capacitance is
\begin{equation}
\hat{C}_{g}^{\mathrm{SRAM}}
=
\operatorname{clip}
\left(
\frac{
\log_{10}\left(C_g' \times 10^{20}\right)
}{
6
},
0,
1
\right).
\label{eq:sram_cg_normalization}
\end{equation}
These transformations compress the large capacitance range while
preserving its magnitude ordering.

\paragraph{SRAM effective-resistance labels.}

Let $R_{\mathrm{eff}}$ denote the effective-resistance label of a target
edge. Zero-valued labels are removed before normalization:
\begin{equation}
\mathcal{E}_{r}^{\mathrm{valid}}
=
\left\{
e \in \mathcal{E}_{r}
\mid
R_{\mathrm{eff}}(e) > 0
\right\}.
\label{eq:sram_reff_filter}
\end{equation}
This filtering is required because zero-valued resistance labels are
invalid for the considered target pairs and cannot be transformed into
the logarithmic domain.

For each remaining label, we first compute
\begin{equation}
z
=
\log_{10}\left(R_{\mathrm{eff}}\right).
\label{eq:sram_reff_log}
\end{equation}
Let $p_1$ and $p_{99}$ denote the first and ninety-ninth percentiles of
the logarithmic resistance labels in the training data. The normalized
effective resistance is defined as
\begin{equation}
\hat{R}_{\mathrm{eff}}^{\mathrm{SRAM}}
=
\operatorname{clip}
\left(
\frac{
z-p_1
}{
p_{99}-p_1
},
0,
1
\right).
\label{eq:sram_reff_normalization}
\end{equation}
The percentile-based transformation reduces the influence of the sparse
extreme values while retaining the relative ordering of most resistance
labels.

\paragraph{Analog parasitic labels.}

The analog circuits have different physical label ranges from the SRAM
circuits. Therefore, the fixed capacitance thresholds used for the SRAM
subset are not applied to the analog subset. Instead, analog
effective-resistance and ground-capacitance labels are normalized using
logarithmic maximum-based transformations.

Let
\begin{equation}
M_R
=
\max_{R_{\mathrm{eff}}
\in
\mathcal{Y}_{R,\mathrm{train}}}
R_{\mathrm{eff}}
\label{eq:analog_reff_max}
\end{equation}
denote the maximum effective-resistance label in the analog training
split. This value corresponds to \texttt{edge\_labelmax} in the
preprocessing implementation. The normalized effective resistance is
computed as
\begin{equation}
\hat{R}_{\mathrm{eff}}^{\mathrm{Analog}}
=
\operatorname{clip}
\left(
\frac{
\ln\left(1+R_{\mathrm{eff}}\right)
}{
\ln\left(1+M_R\right)
},
0,
1
\right).
\label{eq:analog_reff_normalization}
\end{equation}

For analog ground capacitance, the raw value is multiplied by $10^{15}$
so that its numerical magnitude is expressed on the femtofarad scale.
We define
\begin{equation}
M_C
=
\max_{C_g
\in
\mathcal{Y}_{C,\mathrm{train}}}
\left(10^{15}C_g\right),
\label{eq:analog_cg_max}
\end{equation}
which corresponds to \texttt{node\_labelmax} in the preprocessing
implementation. The normalized ground capacitance is then given by
\begin{equation}
\hat{C}_{g}^{\mathrm{Analog}}
=
\operatorname{clip}
\left(
\frac{
\ln\left(1+10^{15}C_g\right)
}{
\ln\left(1+M_C\right)
},
0,
1
\right).
\label{eq:analog_cg_normalization}
\end{equation}

The logarithmic transformations compress the long-tailed analog label
distributions while preserving the relative ordering of the original
physical values. The $\operatorname{clip}(\cdot,0,1)$ operation further
restricts all normalized labels to $[0,1]$ and prevents out-of-range
values caused by previously unseen extreme samples.

\subsection{Classification Label Construction}
\label{subsec:classification_label_construction}

The regression tasks directly use the normalized continuous labels
defined above as supervision targets. For classification tasks, each
normalized label $\hat{y}\in[0,1]$ is discretized into five magnitude
classes using the fixed boundary set
\begin{equation}
\mathcal{B}
=
\{0.2,\,0.4,\,0.6,\,0.8\}.
\label{eq:classification_boundaries}
\end{equation}

The corresponding categorical label is defined as
\begin{equation}
y_{\mathrm{class}}
=
\sum_{b\in\mathcal{B}}
\mathbb{I}\left(\hat{y}>b\right),
\label{eq:label_bucketization}
\end{equation}
where $\mathbb{I}(\cdot)$ denotes the indicator function. The five classes, therefore, correspond to
\begin{equation}
\begin{aligned}
\mathcal{C}_0 &: [0,0.2],\\
\mathcal{C}_1 &: (0.2,0.4],\\
\mathcal{C}_2 &: (0.4,0.6],\\
\mathcal{C}_3 &: (0.6,0.8],\\
\mathcal{C}_4 &: (0.8,1].
\end{aligned}
\label{eq:classification_intervals}
\end{equation}

This operation is equivalent to applying \texttt{torch.bucketize} with
the boundaries \texttt{[0.2, 0.4, 0.6, 0.8]} and
\texttt{right=False}. The same fixed boundaries are used for all circuit
designs, ensuring that each class has a consistent numerical
interpretation across the complete benchmark. Regression and
classification tasks consequently share the same filtered and
normalized label space.

\section{Additional Results}
\label{app:results}

This appendix provides additional results that complement the main experimental analysis. Specifically, we report SRAM regression results and analog classification results, covering both node-level and edge-level parasitic prediction tasks.

\subsection{Experimental Settings}
\label{app:experimental_settings}

The additional experiments follow the same data preprocessing, task
definitions, and evaluation protocol as those reported in the main
paper.

\paragraph{Data splits.}
For the SRAM subset, the models are trained and validated on the
combined \textit{\textbf{ssram}}, \textit{\textbf{digtime}}, and
\textit{\textbf{timing\_ctrl}} designs, and evaluated on the unseen
\textit{\textbf{sandwich}}, \textit{\textbf{ultra8t}}, and
\textit{\textbf{array\_128\_32\_8t}} designs. This split evaluates
cross-scale generalization from relatively small training circuits to
substantially larger SRAM graphs.

For the analog subset, circuits \textit{\textbf{1--4}},
\textit{\textbf{6}}, \textit{\textbf{8--12}}, and
\textit{\textbf{15--18}} are used for training and validation, while
circuits \textit{\textbf{5}}, \textit{\textbf{14}}, and
\textit{\textbf{20}} are held out for cross-design evaluation. This
setting evaluates model generalization across unseen analog circuit
topologies and functional categories.

\paragraph{Tasks and baselines.}
The SRAM experiments include ground-capacitance node prediction,
coupling-capacitance edge prediction, and effective-resistance edge
prediction. The analog experiments include a ground-capacitance node
prediction and effective-resistance edge prediction. We evaluate four
message-passing GNNs, including GCN, GAT, GraphSAGE, and PNA; two graph
Transformers, including SGFormer and PolyNormer, and two
circuit-specific methods, including CircuitGPS and CircuitGCL.

\paragraph{Training configuration.}
For the standard SRAM baselines, we use four GNN layers, two prediction
head layers, a hidden dimension of 144, a dropout rate of 0.4, a batch
size of 128, and a learning rate of $10^{-4}$. Four-hop neighborhoods
with at most 64 sampled neighbors per hop are used for mini-batch
training. All models are trained for 200 epochs with random seed 42.

For the standard analog baselines, we use four GNN layers, a hidden
dimension of 128, a dropout rate of 0.3, a batch size of 64, and a
learning rate of $5\times10^{-5}$. Four-hop neighborhood sampling is
used with at most 128 sampled neighbors per hop. For analog edge-level
tasks, 60\% of the target edges are sampled during training. All analog
models are trained for 200 epochs with random seed 42.

CircuitGPS and CircuitGCL use their model-specific configurations. On the
SRAM subset, CircuitGPS uses four GNN layers, two prediction head layers, a
hidden dimension of 144, a dropout rate of 0.3, and a learning rate of
$10^{-4}$. CircuitGCL uses four GNN layers, three prediction head
layers, a hidden dimension of 96, a dropout rate of 0.4, and a learning
rate of $5\times10^{-5}$. Both models use a batch size of 128.

On the analog subset, CircuitGPS uses four GNN layers, two prediction head
layers, a hidden dimension of 84, a dropout rate of 0.4, a learning rate
of $10^{-4}$, and three-hop subgraph sampling. CircuitGCL uses four GNN
layers, two prediction head layers, a hidden dimension of 144, a
dropout rate of 0.4, and a learning rate of $10^{-4}$.

\paragraph{Evaluation metrics.}
Regression performance is evaluated using mean absolute error (MAE) and
the coefficient of determination ($R^2$), where lower MAE and higher
$R^2$ indicate better performance. Classification performance is
evaluated using accuracy and F1-score. 

\subsection{Performance on SRAM Regression Tasks}
\label{app:sram_regression}

\input{tables/sram_ground_regression}
\input{tables/sram_coupling_regression}

\Cref{tab:sram_ground_regression} and \Cref{tab:sram_coupling_regression} report the SRAM ground-capacitance node regression and coupling-capacitance edge regression results. For ground-capacitance regression, CircuitGPS achieves the strongest overall performance, obtaining the best MAE and $R^2$ on most SRAM test circuits. In particular, it generalizes well from smaller training designs to unseen large-scale SRAM circuits, including \textit{\textbf{sandwich}}, \textit{\textbf{ultra8t}}, and \textit{\textbf{array\_128\_32\_8t}}. However, coupling-capacitance edge regression is more challenging, where the best model varies across circuits and metrics. CircuitGCL achieves the best performance on \textit{\textbf{sram+digtime+timing\_ctrl}} and the best $R^2$ on \textit{\textbf{ultra8t}}, while PolyNormer, GraphSAGE, CircuitGPS, PNA, and SGFormer achieve the best scores on different remaining columns. These results indicate that SRAM edge-level coupling prediction is more difficult than node-level ground-capacitance prediction, especially under cross-scale generalization.

\subsection{Performance on Analog Classification Tasks}
\label{app:analog_classification}

\input{tables/analog_ground_classification}
\input{tables/analog_effective_classification}

\Cref{tab:analog_ground_classification} and \Cref{tab:analog_resistance_classification} report the analog ground-capacitance node classification and effective-resistance edge classification results. For ground-capacitance classification, PNA and PolyNormer achieve the best accuracy on the main analog subset, while CircuitGCL obtains the best F1-score, suggesting that imbalance-aware learning improves minority-class recognition. On held-out analog circuits \textit{\textbf{5}}, \textit{\textbf{14}}, and \textit{\textbf{20}}, the best model varies across circuit IDs, reflecting the structural diversity of analog designs. For effective resistance classification, no single model dominates all test circuits. CircuitGCL performs best on the main analog subset and achieves the highest F1-score on circuit \textit{\textbf{14}}, while SGFormer performs best on circuits \textit{\textbf{5}} and \textit{\textbf{20}}. These results further show that analog classification is sensitive to cross-design structural shifts and benefits from both circuit-specific modeling and imbalance-aware training.

\section{Algorithms}
\label{app:algorithm}

To generate high-precision physical labels for the dataset, the effective resistance ($R$) between any two nodes in the circuit graph is calculated. The time cost of direct circuit simulation is high, so this study adopts an efficient calculation method based on matrix operations.

\subsection{Matrix-Based Effective Resistance Calculation}
The core principle of this method is to use the node admittance matrix of the circuit to realize the effective resistance calculation. As shown in \Cref{alg:resistance_calc}, first fill all the resistance information in the circuit netlist to the node admittance matrix $G$, and each row and column of the matrix corresponds to a node in the circuit. 

In the first stage of the algorithm, all the original resistive elements are traversed, the resistance value is converted to a conductance value ($g = 1/r$), and the diagonal and non-diagonal elements of the matrix $G$ are updated according to Kirchhoff's law of current. Then the reference node (usually the ground node) is selected, and the corresponding matrix row/column of the node is eliminated to obtain the invertible admittance matrix. 

The second stage of the algorithm is the core step, i.e., using the inverse of the Cholesky factor of the invertible admittance matrix to obtain the effective resistances.
The inverse of the Cholesky factor effectively captures the inverse of the admittance matrix. And with it, one can easily query the effective resistance between any two nodes in the circuit \cite{liu2023computing}.  
The computational efficiency of this method is significantly better than that of the traditional path search method, and can efficiently support the generation of tens of millions of edge-level labels in the dataset.

\begin{algorithm}[!t]
  \caption{Matrix-Based Effective Resistance Calculation}
  \label{alg:resistance_calc}
  {\small
  \linespread{1.05}\selectfont
  \setlength{\parskip}{0pt}
  \begin{algorithmic}
    \STATE \textbf{Input:} Resistor List $\mathcal{R}$ of a Net, Port List $\mathcal{P}$
    \STATE \textbf{Output:} List of effective resistances $\mathcal{L}_{out} = \{(src, dst, val)\}$
    
    \STATE \textbf{Initialization}
    \STATE $\mathcal{L}_{out} \leftarrow \emptyset$
    \STATE $\mathcal{V} \leftarrow \text{ExtractUniqueNodes}(\mathcal{R})$
    \STATE $N \leftarrow |\mathcal{V}|$
    \IF{$N < 2$ \textbf{or} $|\mathcal{P}| < 2$} 
      \STATE \textbf{return} $\emptyset$
    \ENDIF
    \STATE $\mathcal{M} \leftarrow \text{MapNodesToIndices}(\mathcal{V})$
    
    \STATE
    \STATE \textbf{Stage 1: Construction of Invertible Admittance Matrix}
    \STATE $G \leftarrow \mathbf{0}_{N \times N}$
    \FOR{\textbf{all} $(n_1, n_2, r) \in \mathcal{R}$}
        \STATE $g \leftarrow 1 / r$ 
        \STATE $u \leftarrow \mathcal{M}[n_1], \quad v \leftarrow \mathcal{M}[n_2]$
        \STATE $G_{u,u} \leftarrow G_{u,u} + g, \quad G_{v,v} \leftarrow G_{v,v} + g$
        \STATE $G_{u,v} \leftarrow G_{u,v} - g, \quad G_{v,u} \leftarrow G_{v,u} - g$
    \ENDFOR
    \STATE $ref \leftarrow N - 1$ ~ ~ ~ \textit{// Set last node as reference}  
    \STATE $G_{red} \leftarrow G[0:ref, 0:ref]$ 
    
    \STATE
    \STATE \textbf{Stage 2: Computing the Inverse of Cholesky Factor and Port-to-Port Resistances}
    \STATE \textit{// Compute the Cholesky factor L: $G_{red}=LL^T$}  
    \STATE Compute the Cholesky factor $L$ from $G_{red}$, where $L$ is a lower-triangular    matrix
    \STATE $Z \leftarrow L^{-1}$ ~ ~ ~ ~ ~ ~ \textit{// Compute L's inverse}  
    \FOR{$k = 0$ \textbf{to} $|\mathcal{P}| - 1$}
        \FOR{$l = k + 1$ \textbf{to} $|\mathcal{P}| - 1$}
            \STATE $src\_id \leftarrow \mathcal{P}[k]$
            \STATE $dst\_id \leftarrow \mathcal{P}[l]$
            \STATE $z_{src} \leftarrow$ the $\mathcal{M}[src\_id]$-th column  of $Z$
            \STATE $z_{dst} \leftarrow$ the $\mathcal{M}[dst\_id]$-th column  of $Z$
            \STATE $R_{eq} \leftarrow \| z_{src} - z_{dst}\|^2   $
            \STATE $\mathcal{L}_{out}.\text{add}((src\_id, dst\_id, R_{eq}))$
        \ENDFOR
    \ENDFOR
    \STATE \textbf{return} $\mathcal{L}_{out}$
  \end{algorithmic}
  }
\end{algorithm}










\end{document}

%% file: tables/benchmark.tex
\begin{table*}[t]
\centering
\caption{Comparison of circuit datasets and benchmarks.
Existing benchmarks focus on digital congestion/timing or schematic-level analog tasks.
ParasGB provides RC labels for post-layout parasitic estimation on both analog and SRAM designs.}
\label{tab:related_works_comparison}

\small
\renewcommand{\arraystretch}{1.08}

\begin{tabular*}{0.92\textwidth}
{@{\extracolsep{\fill}}lcccc@{}}
\toprule
\textbf{Dataset}
& \textbf{Domain}
& \textbf{Design Stage}
& \textbf{Key Tasks}
& \textbf{RC Included} \\
\midrule

CircuitNet \cite{chai2022circuitnet}
& Digital
& Schematic+Layout
& Congestion/Timing
& \xmark \\

EDA-schema \cite{shrestha2024eda}
& Digital
& Schematic+Layout
& Representation
& \xmark \\

CktGNN \cite{dong2023cktgnn}
& Analog
& Schematic
& Topology/Sizing
& \xmark \\

AMSNet \cite{tao2024amsnet}
& Analog
& Schematic
& Generation/Understanding
& \xmark \\

AnalogGenie \cite{gao2025analoggenie}
& Analog
& Schematic
& Topology Generation/Discovery
& \xmark \\

\midrule

\textbf{ParasGB (Ours)}
& \textbf{Analog/Memory}
& \textbf{Schematic}
& \textbf{Parasitic Estimation}
& \textbf{\cmark} \\

\bottomrule
\end{tabular*}
\end{table*}

%% file: tables/analog_statistics.tex
\definecolor{col_yellow}{RGB}{255, 255, 255}
\definecolor{col_blue}{RGB}{255, 255, 255}
\definecolor{col_green}{RGB}{255, 255, 255}
\definecolor{col_red}{RGB}{255, 255, 255}

\begin{table*}[t]
\centering
\caption{Analog Circuit Dataset Statistics. Topology edges ($dev2pin$, $pin2net$) form the input graph, while labeled targets are effective resistance ($R_{eff}$) and ground capacitance ($C_g$).}
\label{tab:analog_statistics}

\setlength{\tabcolsep}{6pt} 

\resizebox{\textwidth}{!}{%
\begin{tabular}{ccc 
    >{\columncolor{col_yellow}}c >{\columncolor{col_yellow}}c >{\columncolor{col_yellow}}c >{\columncolor{col_yellow}}c 
    >{\columncolor{col_blue}}c >{\columncolor{col_blue}}c >{\columncolor{col_blue}}c 
    >{\columncolor{col_green}}c >{\columncolor{col_green}}c >{\columncolor{col_green}}c 
    >{\columncolor{col_red}}c >{\columncolor{col_red}}c >{\columncolor{col_red}}c >{\columncolor{col_red}}c 
}
\toprule
\multirow{2}{*}{\textbf{Scale}} & \multirow{2}{*}{\textbf{ID}} & \multirow{2}{*}{\textbf{Type}} & 
\multicolumn{4}{c}{\cellcolor{col_yellow}\textbf{Nodes}} & 
\multicolumn{3}{c}{\cellcolor{col_blue}\textbf{Edges}} & 
\multicolumn{3}{c}{\cellcolor{col_green}\textbf{$C_g$ Statistics}} & 
\multicolumn{4}{c}{\cellcolor{col_red}\textbf{$R_{eff}$ Statistics}} \\

\cmidrule(lr){4-7} \cmidrule(lr){8-10} \cmidrule(lr){11-13} \cmidrule(lr){14-17}

& & & total & device & pin & net & total & $pin2net$ & $dev2pin$ 
& $n_{\text{mean}}$ & $n_{\text{min}}$ & $n_{\text{max}}$ 
& total & $e_{\text{mean}}$ & $e_{\text{min}}$ & $e_{\text{max}}$ \\
\midrule

\multirow{6}{*}{\textbf{XS}} 
 & 19 & BGR \cite{8859356} & 231 & 68 & 156 & 7  & 428  & 156  & 272  & 2.70E-14 & 8.28E-15 & 6.61E-14 & 5166  & 50.21 & 0.23 & 287.26 \\
 & 13 & OP \cite{5967452} & 256 & 66 & 180 & 10 & 444  & 180  & 264  & 2.99E-14 & 5.45E-15 & 7.64E-14 & 3576  & 34.14 & 0.01 & 205.30 \\
 & 4  & BGR \cite{7072988} & 448 & 132& 306 & 10 & 834  & 306  & 528  & 1.78E-13 & 1.41E-14 & 3.41E-13 & 12170 & 74.65 & 0.79 & 590.21 \\
 & 9  & LDO \cite{7001720} & 1541& 478& 1048& 15 & 2960 & 1048 & 1912 & 2.56E-13 & 1.62E-14 & 1.22E-12 & 26796 & 70.16 & 0.01 & 349.64 \\
 & 10 & OP \cite{1495717} & 1298& 408& 873 & 17 & 2501 & 873  & 1628 & 8.31E-14 & 2.81E-15 & 4.24E-13 & 21294 & 41.58 & 0.13 & 615.11 \\
 & 11 & LDO \cite{8688481}& 2899& 902& 1978& 19 & 5586 & 1978 & 3608 & 3.03E-13 & 2.08E-14 & 1.01E-12 & 33078 & 70.24 & 0.00 & 529.63 \\
\midrule

\multirow{9}{*}{\textbf{S}} 
 & 16 & LDO \cite{7829748} & 1364& 413& 924 & 27 & 2550 & 924  & 1626 & 1.65E-13 & 1.04E-15 & 8.08E-13 & 19554 & 133.33 & 0.01 & 526.15 \\
 & 2  & OP \cite{4098517}  & 534 & 152& 350 & 32 & 914  & 350  & 564  & 6.58E-14 & 1.07E-15 & 5.16E-13 & 13774 & 65.72  & 0.01 & 756.66 \\
 & 1  & LVDS \cite{leung2003cmos} & 1100& 251& 814 & 35 & 1776 & 814  & 962  & 5.22E-14 & 1.04E-15 & 3.01E-13 & 21554 & 63.42  & 0.01 & 264.93 \\
 & 12 & LDO \cite{9964454} & 1409& 440& 930 & 39 & 2634 & 930  & 1704 & 5.80E-14 & 1.04E-15 & 5.82E-13 & 15212 & 58.69  & 0.01 & 294.71 \\
 & 6  & LDO \cite{5395671} & 1836& 553& 1242& 41 & 3414 & 1242 & 2172 & 2.18E-13 & 1.04E-15 & 1.85E-12 & 29740 & 133.82 & 0.01 & 512.75 \\
 & 3  & BGR \cite{6180360} & 1225& 382& 800 & 43 & 2260 & 800  & 1460 & 5.60E-14 & 1.04E-15 & 8.97E-13 & 11128 & 20.43  & 0.71 & 235.77 \\
 & 17 & LDO \cite{9332114} & 1451& 438& 961 & 52 & 2637 & 961  & 1676 & 1.09E-13 & 1.04E-15 & 1.70E-12 & 22054 & 405.55 & 0.01 & 3202.96 \\
 & 8  & LDO \cite{10870569} & 1566& 475& 1027& 64 & 2823 & 1027 & 1796 & 6.20E-14 & 1.06E-15 & 9.54E-13 & 20828 & 88.09  & 0.01 & 350.96 \\
 & 14 & BGR \cite{7763843} & 591 & 138& 362 & 91 & 746  & 362  & 384  & 1.13E-14 & 1.03E-15 & 2.94E-13 & 6756  & 134.92 & 0.06 & 386.95 \\
\midrule

\multirow{5}{*}{\textbf{M}} 
 & 18 & LDO \cite{6699480} & 2496& 758 & 1627& 111 & 4467 & 1627 & 2840 & 6.64E-14 & 1.05E-15 & 1.35E-12 & 24972 & 57.12  & 0.02 & 289.77 \\
 & 15 & LDO \cite{5548891} & 3672& 1020& 2235& 417 & 5535 & 2235 & 3300 & 3.07E-14 & 1.03E-15 & 2.14E-12 & 38322 & 60.79  & 0.01 & 326.16 \\
 & 20 & BGR \cite{edward2009low} & 2388& 621 & 1304& 463 & 2871 & 1304 & 1567 & 7.16E-15 & 1.04E-15 & 5.58E-13 & 14874 & 41.76  & 0.00 & 559.43 \\
 & 5  & LDO \cite{1158794} & 6917& 1738& 3496& 1683& 7092 & 3496 & 3596 & 2.18E-15 & 1.04E-15 & 2.76E-13 & 6512  & 171.95 & 0.90 & 1972.48 \\
 & 7  & LDO \cite{1158794} & 6917& 1738& 3496& 1683& 7092 & 3496 & 3596 & 2.18E-15 & 1.04E-15 & 2.76E-13 & 6512  & 171.95 & 0.90 & 1972.48 \\
\midrule

\multicolumn{3}{c}{\textbf{Total Average}} 
& \textbf{2007.0} & \textbf{558.6} & \textbf{1205.5} & \textbf{243.0} 
& \textbf{2978.3} & \textbf{1205.5} & \textbf{1772.8} 
& \textbf{8.91E-14} & \textbf{4.11E-15} & \textbf{7.82E-13} 
& \textbf{17693.6} & \textbf{97.4} & \textbf{0.2} & \textbf{711.5} \\
\bottomrule
\end{tabular}
}
\end{table*}

%% file: tables/sram_statistics.tex
\definecolor{col_yellow}{RGB}{255,255,255}
\definecolor{col_blue}{RGB}{255,255,255}
\definecolor{col_green}{RGB}{255,255,255}
\definecolor{col_red}{RGB}{255,255,255}
\definecolor{col_purple}{RGB}{255,255,255}

\begin{table*}[t]
\centering
\caption{SRAM Circuit Dataset Statistics. Topology edges ($dev2pin$, $pin2net$) form the input graph; parasitic targets include node-level ground capacitance $C_g$, aggregated coupling-capacitance edges ($C_c$) and resistance edges $R_{eff}$.}
\label{tab:sram_statistics}

\setlength{\tabcolsep}{2.5pt} 

\resizebox{\textwidth}{!}{%
\begin{tabular}{lc 
    >{\columncolor{col_yellow}}c >{\columncolor{col_yellow}}c >{\columncolor{col_yellow}}c >{\columncolor{col_yellow}}c 
    >{\columncolor{col_blue}}c >{\columncolor{col_blue}}c >{\columncolor{col_blue}}c 
    >{\columncolor{col_green}}c >{\columncolor{col_green}}c >{\columncolor{col_green}}c 
    >{\columncolor{col_red}}c >{\columncolor{col_red}}c >{\columncolor{col_red}}c >{\columncolor{col_red}}c 
    >{\columncolor{col_purple}}c >{\columncolor{col_purple}}c >{\columncolor{col_purple}}c >{\columncolor{col_purple}}c
}
\toprule
\multirow{2}{*}{\textbf{Scale}} & \multirow{2}{*}{\textbf{Dataset}} & 
\multicolumn{4}{c}{\cellcolor{col_yellow}\textbf{Nodes}} & 
\multicolumn{3}{c}{\cellcolor{col_blue}\textbf{Edges}} & 
\multicolumn{3}{c}{\cellcolor{col_green}\textbf{$C_g$ Statistics}} & 
\multicolumn{4}{c}{\cellcolor{col_red}\textbf{$C_c$ Statistics}} &
\multicolumn{4}{c}{\cellcolor{col_purple}\textbf{$R_{eff}$ Statistics}} \\

\cmidrule(lr){3-6} \cmidrule(lr){7-9} \cmidrule(lr){10-12} \cmidrule(lr){13-16} \cmidrule(lr){17-20}

& & total & device & pin & net & total & $dev2pin$ & $pin2net$ & $n_{\text{mean}}$ & $n_{\text{min}}$ & $n_{\text{max}}$ & \text{total} & $e_{\text{mean}}$ & $e_{\text{min}}$ & $e_{\text{max}}$ & \text{total} & $e_{\text{mean}}$ & $e_{\text{min}}$ & $e_{\text{max}}$ \\
\midrule

\multirow{2}{*}{\textbf{L}} 
 & digtime \cite{tscache} & 17.6K & 4.1K & 12.4K & 1.0K & 24.8K & 12.4K & 12.4K & 3.17E-17 & 1.14E-19 & 8.63E-15 & 44.4K & 8.03E-18 & 3.10E-25 & 4.62E-14 & 99.8K & 427.40 & 2.58 & 3009.74 \\
 & timing\_ctrl & 18.2K & 4.4K & 13.1K & 668 & 26.2K & 13.1K & 13.1K & 1.42E-16 & 3.31E-20 & 4.13E-15 & 43.1K & 2.19E-17 & 4.44E-27 & 3.86E-14 & 99.6K & 212.88 & 4.97 & 1168.11 \\
\midrule

\multirow{2}{*}{\textbf{XL}} 
 & array\_128 & 143K & 32.8K & 98.3K & 12.6K & 197K & 98.3K & 98.3K & 1.19E-16 & 6.66E-19 & 5.98E-14 & 414K & 7.32E-18 & 5.65E-27 & 6.96E-14 & 100K & 232.61 & 9.34 & 1198.62 \\
 & sram & 249K & 57.4K & 172K & 19.9K & 344K & 172K & 172K & 5.48E-17 & 1.55E-25 & 1.43E-14 & 760K & 8.30E-18 & 1.37E-26 & 3.14E-13 & 41.9K & 266.70 & 22.04 & 413.77 \\
\midrule

\multirow{2}{*}{\textbf{XXL}} 
 & ultra8t \cite{shen2024ultra8t} & 10.1M & 2.3M & 6.9M & 861K & 13.8M & 6.9M & 6.9M & 1.60E-16 & 1.41E-23 & 9.46E-12 & 33.6M & 1.03E-17 & 9.02E-38 & 1.84E-11 & 437K & 87.65 & 8.49 & 183.84 \\
 & sandwich \cite{cim1} & 11.7M & 2.6M & 7.9M & 1.1M & 15.8M & 7.9M & 7.9M & 2.46E-16 & 1.11E-25 & 9.10E-12 & 38.4M & 1.13E-17 & 4.78E-34 & 2.08E-11 & 1.0M & 286.03 & 22.10 & 1203.66 \\
\midrule

\multicolumn{2}{c}{\textbf{Total Average}} & 
\textbf{3.70M} & \textbf{833K} & \textbf{2.52M} & \textbf{332K} & \textbf{5.04M} & \textbf{2.52M} & \textbf{2.52M} & 
\textbf{1.26E-16} & \textbf{1.36E-19} & \textbf{3.10E-12} & 
\textbf{12.2M} & \textbf{1.12E-17} & \textbf{5.17E-26} & \textbf{6.53E-12} & \textbf{296K} & \textbf{252.21} & \textbf{11.59} & \textbf{1196.29} \\
 
\bottomrule
\end{tabular}
}

\end{table*}

%% file: tables/task_overview.tex
\newcommand{\graphsize}[1]{\raisebox{-0.15ex}{\textbf{#1}}}
\newcommand{\nmark}{\rule[0.5ex]{0.2cm}{1.5pt}}
\newcommand{\vcmark}{\raisebox{-0.3ex}{\cmark}}
\newcommand{\vxmark}{\raisebox{-0.3ex}{\xmark}} 
\newcommand{\vnmark}{\raisebox{-0.3ex}{\nmark}}

\begin{table*}[!t]
  \centering
  \caption
  {ParasGB Dataset Splits and Task Granularity.}
  \label{tab:datasets_overview}

  \setlength{\tabcolsep}{4pt}

  \resizebox{1\textwidth}{!}{%
  \begin{tabular}{C{1.6cm} C{1.2cm} C{5.0cm} C{1.0cm} C{1.0cm} C{1.0cm} C{1.0cm} C{1.0cm} C{1.0cm} C{1.2cm} C{1.6cm} C{1.6cm}}
    \toprule
    \multirow{3}{*}{\textbf{Category}} &
    \multirow{3}{*}{\textbf{Graph Size}} &
    \multirow{3}{*}{\textbf{Design}} &
    \multicolumn{4}{c}{\textbf{Graph Task}} &
    \multicolumn{3}{c}{\textbf{Parasitic Task}} &
    \multicolumn{2}{c}{\textbf{Metric}} \\
    \cmidrule(lr){4-7}\cmidrule(lr){8-10}\cmidrule(lr){11-12}
    & & &
    \multicolumn{2}{c}{Node-level} &
    \multicolumn{2}{c}{Edge-level} &
    \multirow{2}{*}{$C_g$} &
    \multirow{2}{*}{$C_c$} &
    \multirow{2}{*}{$R_{eff}$} &
    \multirow{2}{*}{Class.} &
    \multirow{2}{*}{Reg.} \\
    \cmidrule(lr){4-5}\cmidrule(lr){6-7}
    & & & Class. & Reg. & Class. & Reg. & & & & & \\
    \midrule

    \multirow[c]{3}{*}{\raisebox{-0.8ex}{\textbf{SRAM}}}
    & \graphsize{L} & digtime, timing\_ctrl &
    \multirow[c]{3}{*}{\vcmark} & \multirow[c]{3}{*}{\vcmark} &
    \multirow[c]{3}{*}{\vcmark} & \multirow[c]{3}{*}{\vcmark} &
    \multirow[c]{3}{*}{\vcmark} & \multirow[c]{3}{*}{\vcmark} & \multirow[c]{3}{*}{\vcmark} &
    \multirow[c]{3}{*}{acc/$F_1$} & \multirow[c]{3}{*}{mae/$R^2$} \\
    \cmidrule(lr){2-3}
    & \graphsize{XL} & ssram, array\_128\_32\_8t & & & & & & & & & \\
    \cmidrule(lr){2-3}
    & \graphsize{XXL} & ultra8t, sandwich & & & & & & & & & \\
    \midrule

    \multirow[c]{3}{*}{\raisebox{-0.8ex}{\textbf{Analog}}}
    & \graphsize{XS} & 19, 13, 4, 9, 10, 11 &
    \multirow[c]{3}{*}{\vcmark} & \multirow[c]{3}{*}{\vcmark} &
    \multirow[c]{3}{*}{\vcmark} & \multirow[c]{3}{*}{\vcmark} &
    \multirow[c]{3}{*}{\vcmark} & \multirow[c]{3}{*}{\vnmark} & \multirow[c]{3}{*}{\vcmark} &
    \multirow[c]{3}{*}{acc/$F_1$} & \multirow[c]{3}{*}{mae/$R^2$} \\
    \cmidrule(lr){2-3}
    & \graphsize{S} & 16, 2, 12, 6, 3, 17, 8, 14 & & & & & & & & & \\
    \cmidrule(lr){2-3}
    & \graphsize{M} & 18, 15, 20, 5, 7 & & & & & & & & & \\
    \bottomrule
  \end{tabular}
  }
\end{table*}


%% file: tables/analog_features.tex
\begin{table}[htbp]
  \centering
  \caption{Definition of Analog Circuit Graph Node Features.}
  \label{tab:analog_node_features}
  
  \resizebox{\linewidth}{!}{%
  \begin{tabular}{lllc} 
    \toprule  
    \textbf{Type} & \textbf{Feature} & \textbf{Definition} & \textbf{Index} \\
    \midrule  

    \multirow{8}{*}{Device}
    & $D_{mos}$ & Is MOS device & 0 \\
    & $D_{res}$ & Is resistor & 1 \\
    & $D_{cap}$ & Is capacitor & 2 \\
    & $W$ & Width of the transistor & 3 \\
    & $L$ & Length of the transistor  & 4 \\
    & $M_{mos}$ & Multiplier of transistors  & 5 \\
    & $N_f$ & Number of fingers & 6 \\
    & $T$ & Type of the device & 7 \\
    \midrule 

    \multirow{10}{*}{Net}
    & $F_{pwr}$ & Power Net flag & 0 \\
    & $F_{gnd}$ & Ground Net flag & 1 \\
    & $F_{sig}$ & Signal Net flag & 2 \\
    & $F_{io}$ & Is connected to external I/O port & 3 \\
    & $N_{mos}$ & Number of connected transistors  & 4 \\
    & $N_{g}$ & Number of connected gate terminals  & 5 \\
    & $N_{sd}$ & Number of connected source/drain terminals & 6 \\
    & $N_{b}$ & Number of connected base terminals & 7 \\
    & $L_{tot}$ & Total channel length of connected transistor & 8 \\
    & $W_{tot}$ & Total channel width of connected transistor & 9 \\
    \midrule 

    \multirow{6}{*}{Pin}
    & $P_{d}$ & Is connected to drain terminal & 0 \\
    & $P_{g}$ & Is connected to gate terminal & 1 \\
    & $P_{s}$ & Is connected to source terminal & 2 \\
    & $P_{b}$ & Is connected to bulk terminal & 3 \\
    & $P_{cap}$ & Is connected to capacitor terminal & 4 \\
    & $P_{res}$ & Is connected to resistor terminal & 5 \\
    \bottomrule 
  \end{tabular}
  }   
\end{table}

%% file: tables/overall_label_distribution.tex
\begin{figure*}[!htbp]
\centering
\begin{minipage}[t]{0.19\textwidth}
\centering
\includegraphics[width=\linewidth]{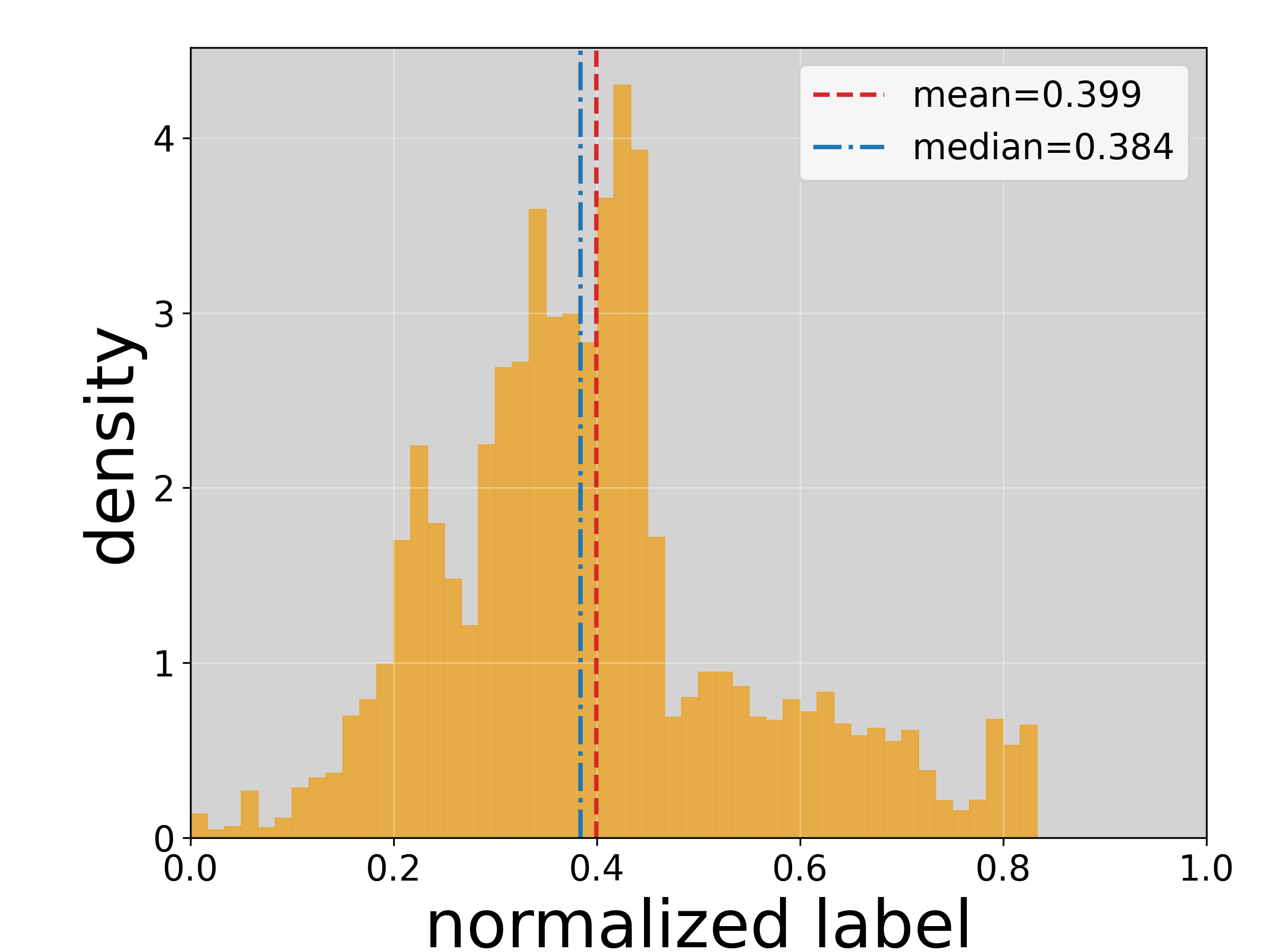}
\subcaption{SRAM $C_g$ Labels}
\end{minipage}
\hfill
\begin{minipage}[t]{0.19\textwidth}
\centering
\includegraphics[width=\linewidth]{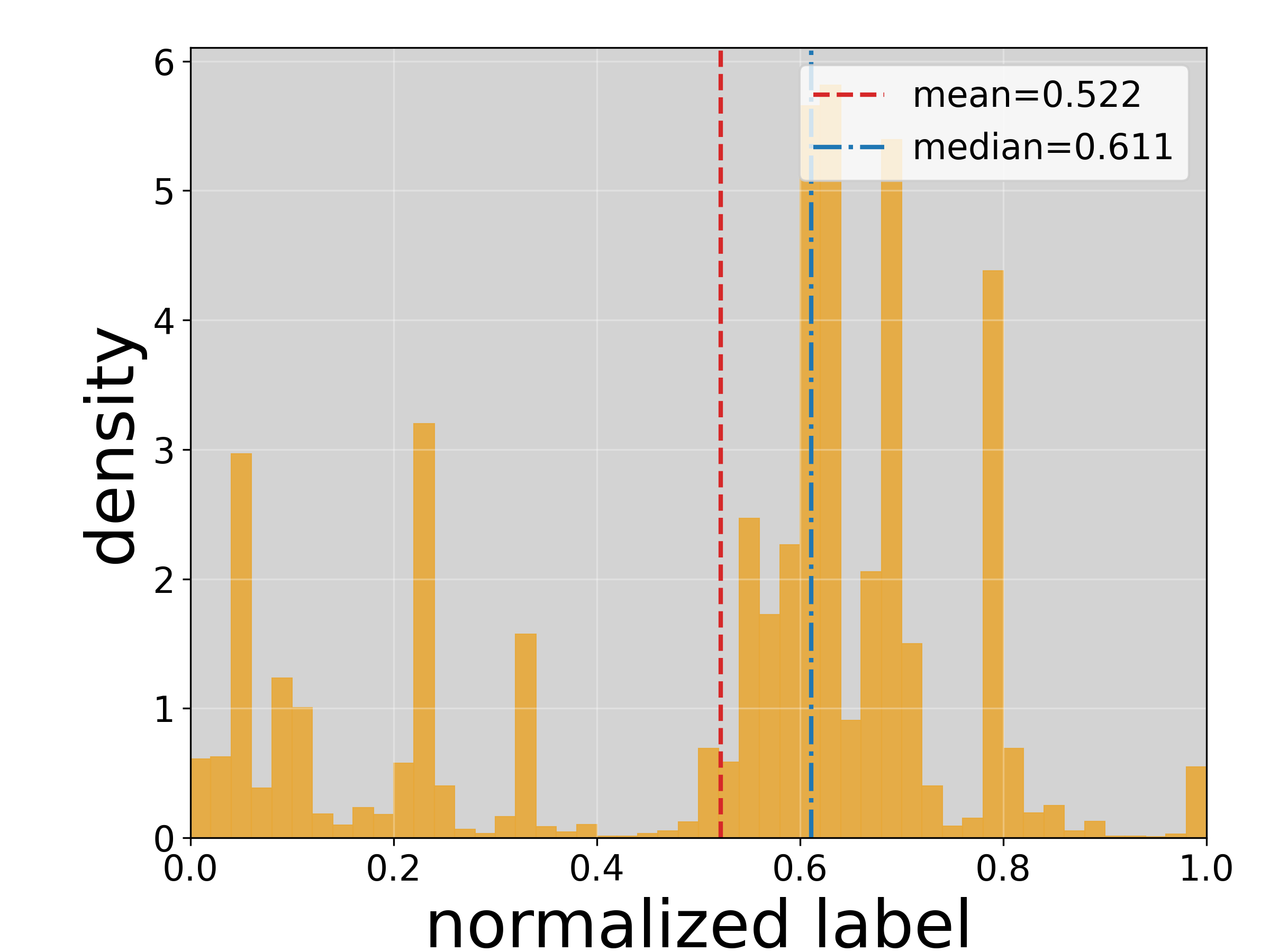}
\subcaption{SRAM $R_{eff}$ Labels}
\end{minipage}
\hfill
\begin{minipage}[t]{0.19\textwidth}
\centering
\includegraphics[width=\linewidth]{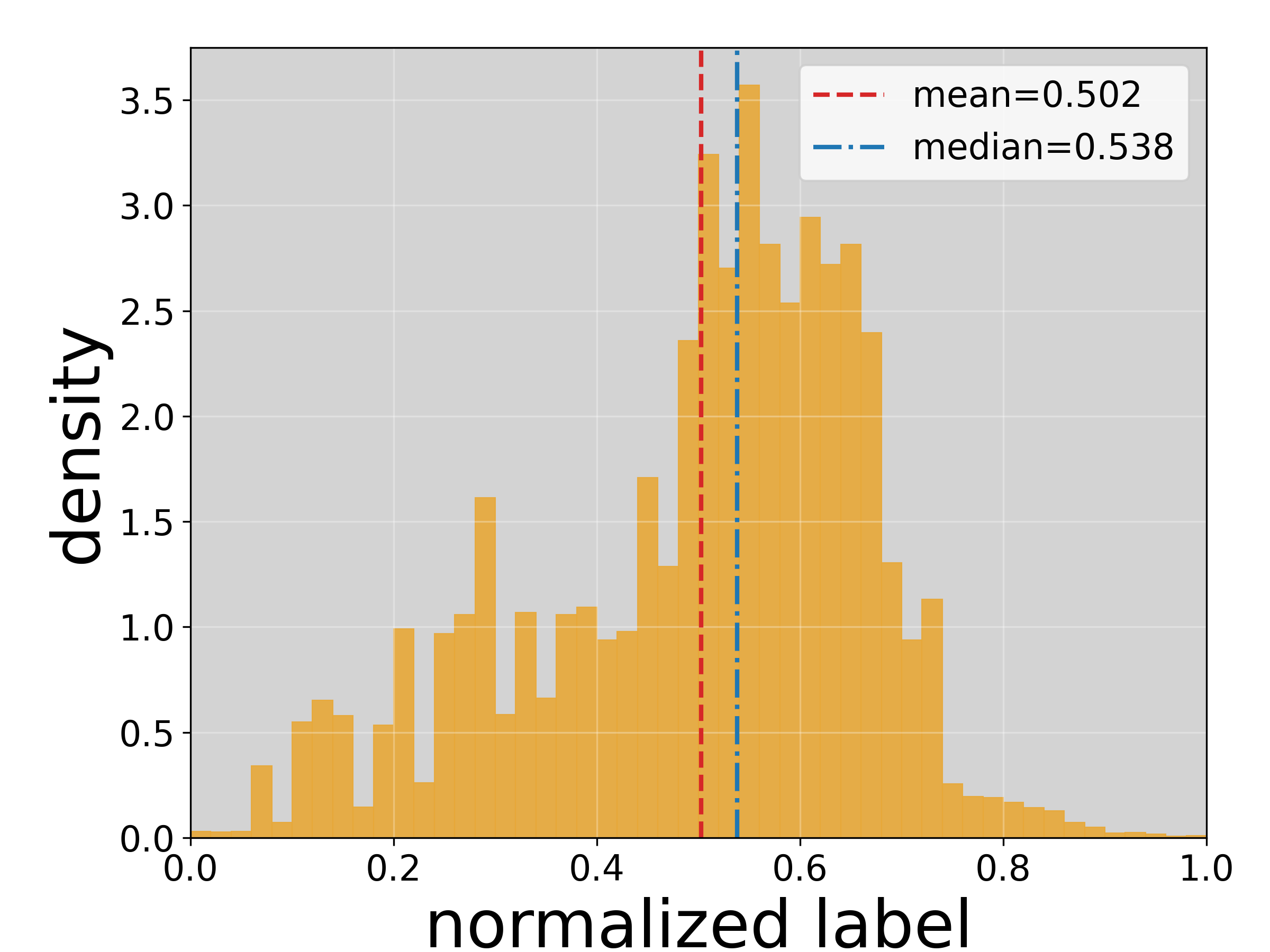}
\subcaption{SRAM $C_c$ Labels}
\end{minipage}
\hfill
\begin{minipage}[t]{0.19\textwidth}
\centering
\includegraphics[width=\linewidth]{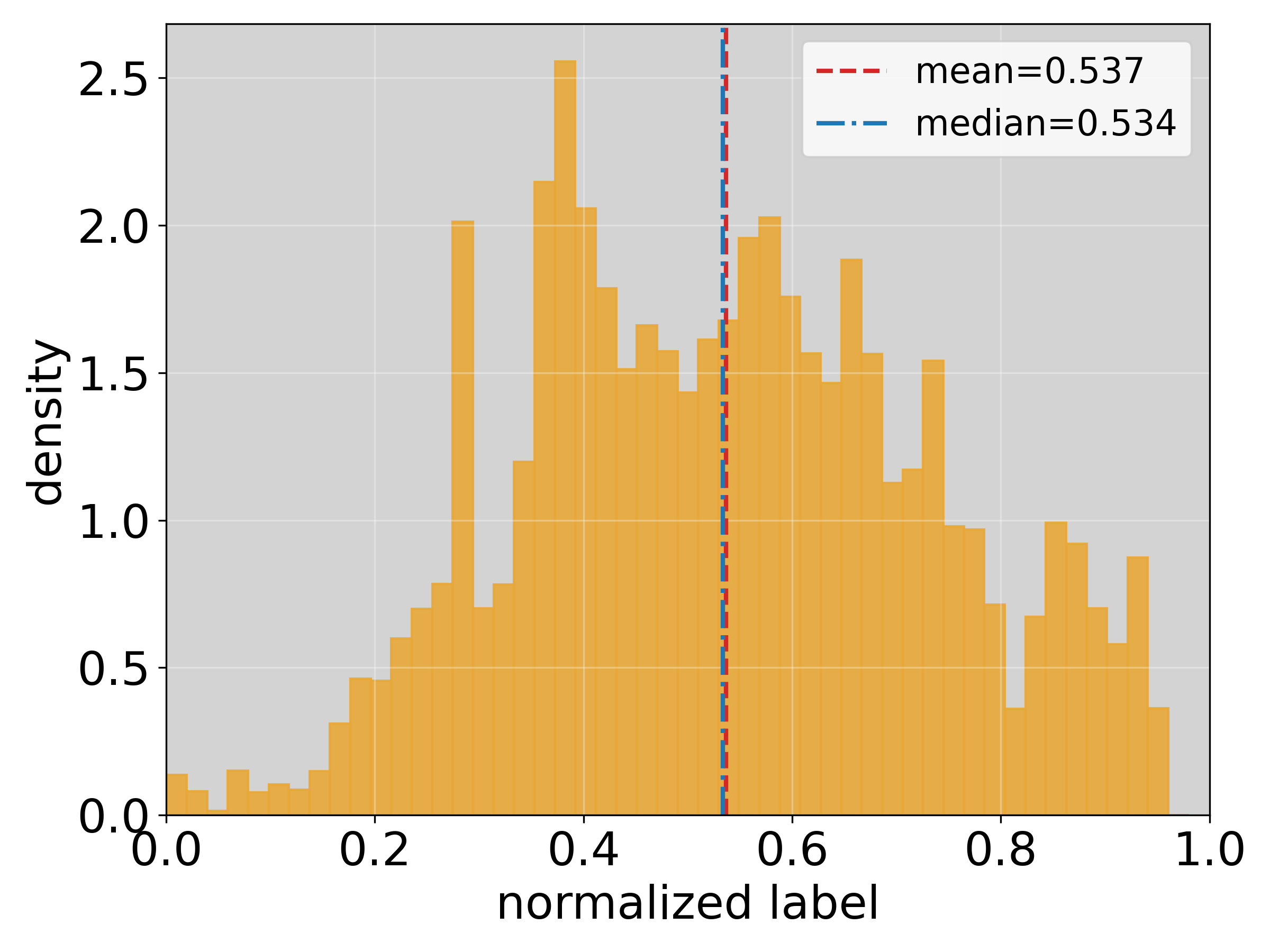}
\subcaption{Analog $R_{eff}$ Labels}
\end{minipage}
\hfill
\begin{minipage}[t]{0.19\textwidth}
\centering
\includegraphics[width=\linewidth]{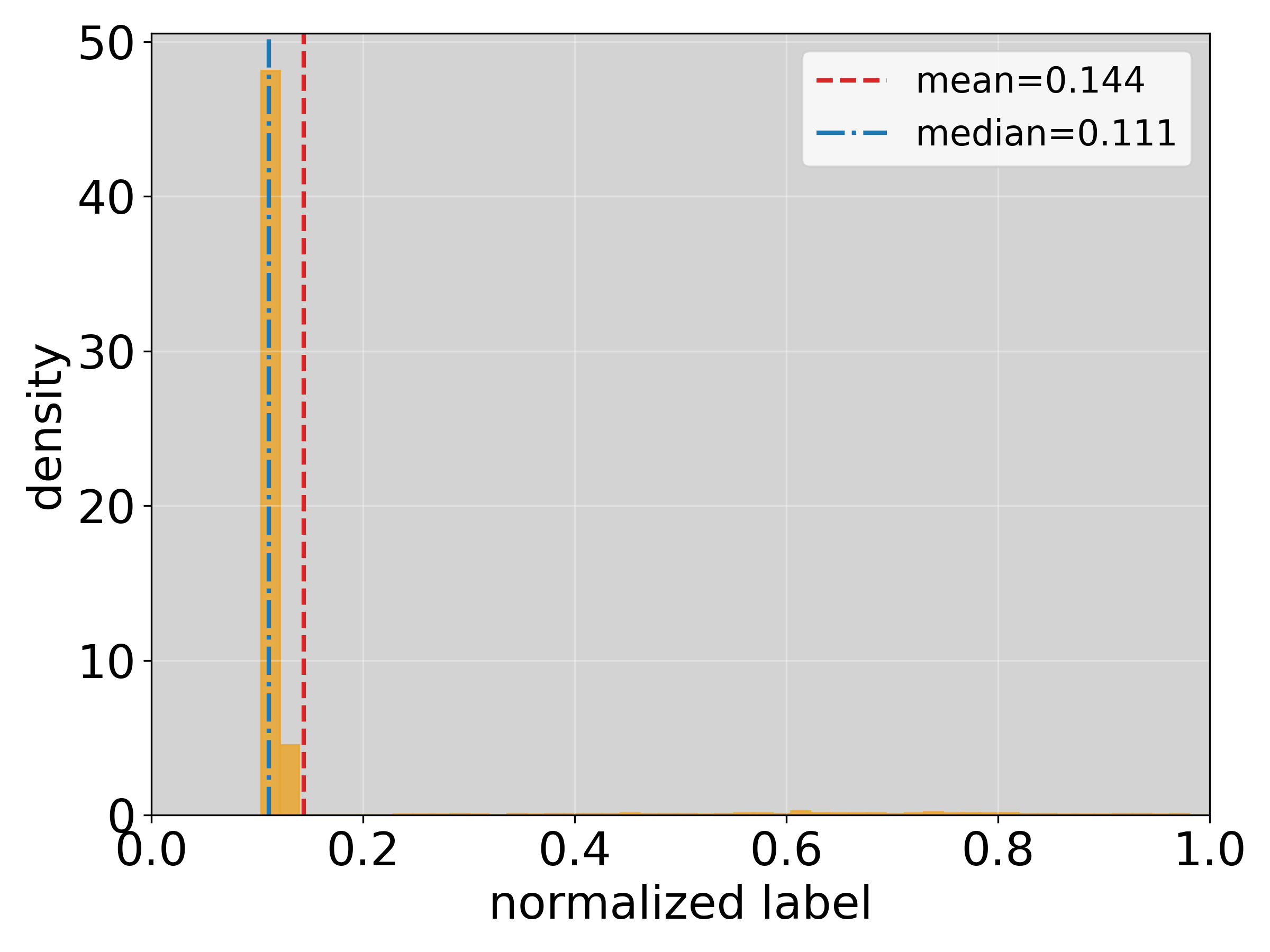}
\subcaption{Analog $C_g$ Labels}
\end{minipage}
\caption{Label Distributions for ParasGB-SRAM and ParasGB-Analog. The pronounced data imbalance poses significant challenges for GNN training, often leading models to overfit dominant samples while neglecting critical outliers.}
\label{fig:overall_label_distribution}
\end{figure*}

%% file: tables/sram_coupling_classification.tex
\begin{table*}[!t]
  \centering
  \setlength{\tabcolsep}{5pt}
  \caption{Performance of Different Models on SRAM Circuits Coupling Capacitance Edge Classification Task}
  \label{tab:sram_coupling_classification}
  \begin{tabular*}{\textwidth}{@{\extracolsep{\fill}}lcccccccc}
    \toprule
    \multirow{2}{*}{\textbf{Metric}}
      & \multicolumn{2}{c}{\textbf{sram+digtime+timing\_ctrl}}
      & \multicolumn{2}{c}{\textbf{sandwich}}
      & \multicolumn{2}{c}{\textbf{ultra8t}}
      & \multicolumn{2}{c}{\textbf{array\_128\_32\_8t}} \\
    \cmidrule(lr){2-3}\cmidrule(lr){4-5}\cmidrule(lr){6-7}\cmidrule(lr){8-9}
      & Accuracy $\uparrow$ & F1-Score $\uparrow$
      & Accuracy $\uparrow$ & F1-Score $\uparrow$
      & Accuracy $\uparrow$ & F1-Score $\uparrow$
      & Accuracy $\uparrow$ & F1-Score $\uparrow$ \\
    \midrule
    GCN & 0.7094 & 0.5395 & 0.5667 & 0.4071 & 0.6154 & 0.4011 & 0.6406 & 0.5361 \\
    GAT & 0.7010 & 0.5142 & 0.5671 & 0.3847 & 0.6328 & 0.4235 & 0.6942 & 0.5590 \\
    GraphSAGE & 0.6735 & 0.4735 & 0.5333 & 0.3685 & 0.6111 & 0.3938 & 0.6182 & 0.3390 \\
    \midrule
    PNA & 0.7263 & 0.5521 & 0.5772 & 0.3970 & 0.6331 & 0.4293 & 0.6924 & 0.5585 \\
    SGFormer & 0.6550 & 0.4971 & 0.5256 & 0.3981 & 0.5828 & 0.4210 & 0.6213 & 0.5264 \\
    PolyNormer & 0.7091 & 0.5383 & 0.5795 & 0.4100 & 0.6446 & 0.4233 & 0.6914 & 0.5632 \\
    \midrule
    CircuitGPS & 0.6541 & 0.5732 & 0.4584 & 0.4023 & 0.6124 & 0.5370 & 0.5534 & 0.4681 \\
    CircuitGCL & \textbf{0.8727} & \textbf{0.7265} & \textbf{0.7673} & \textbf{0.7267} & \textbf{0.7661} & \textbf{0.7121} & \textbf{0.8020} & \textbf{0.7552} \\
    \bottomrule
  \end{tabular*}
\end{table*}

%% file: tables/sram_ground_classification.tex
\begin{table*}[t]
  \centering
  \setlength{\tabcolsep}{5pt}
  \caption{Performance of Different Models on SRAM Circuits Ground Capacitance Node Classification Task}
  \label{tab:sram_ground_classification}
  \begin{tabular*}{\textwidth}{@{\extracolsep{\fill}}lcccccccc}
    \toprule
    \multirow{2}{*}{\textbf{Metric}}
      & \multicolumn{2}{c}{\textbf{sram+digtime+timing\_ctrl}}
      & \multicolumn{2}{c}{\textbf{sandwich}}
      & \multicolumn{2}{c}{\textbf{ultra8t}}
      & \multicolumn{2}{c}{\textbf{array\_128\_32\_8t}} \\
    \cmidrule(lr){2-3}\cmidrule(lr){4-5}\cmidrule(lr){6-7}\cmidrule(lr){8-9}
      & Accuracy $\uparrow$ & F1-Score $\uparrow$
      & Accuracy $\uparrow$ & F1-Score $\uparrow$
      & Accuracy $\uparrow$ & F1-Score $\uparrow$
      & Accuracy $\uparrow$ & F1-Score $\uparrow$ \\
    \midrule
    GCN & 0.5959 & 0.5311 & 0.3661 & 0.3004 & 0.3165 & 0.2930 & 0.2905 & 0.2744 \\
    GAT & 0.6897 & 0.6165 & 0.5034 & 0.4927 & 0.4582 & 0.4575 & 0.4152 & 0.4149 \\
    GraphSAGE & 0.6480 & 0.5799 & 0.3731 & 0.3151 & 0.3634 & 0.3519 & 0.2441 & 0.2149 \\
    \midrule
    PNA & 0.6511 & 0.5923 & 0.4431 & 0.2619 & 0.5980 & 0.3095 & 0.4983 & 0.3399 \\
    SGFormer & 0.6337 & 0.5508 & 0.5669 & 0.5661 & \textbf{0.6718} & \textbf{0.6459} & 0.4307 & 0.4287 \\
    PolyNormer & \textbf{0.7808} & \textbf{0.7557} & 0.5261 & 0.5256 & 0.5891 & 0.5828 & 0.5608 & 0.5094 \\
    \midrule
    CircuitGPS & 0.6031 & 0.5934 & 0.4022 & 0.3315 & 0.3678 & 0.2963 & 0.4012 & 0.3959 \\
    CircuitGCL & 0.7088 & 0.6220 & \textbf{0.5820} & \textbf{0.5819} & 0.5333 & 0.5332 & \textbf{0.6757} & \textbf{0.6879} \\
    \bottomrule
  \end{tabular*}
\end{table*}

%% file: tables/analog_effective_regression.tex
\begin{table*}[t]
  \centering
  \setlength{\tabcolsep}{5pt}
  \caption{Performance of Different Models on Analog Circuits Effective Resistance Edge Regression Task}
  \label{tab:analog_resistance_regression}
  \begin{tabular*}{\textwidth}{@{\extracolsep{\fill}}lcccccccc}
    \toprule
    
    \multirow{2}{*}{\textbf{Metric}}
      & \multicolumn{2}{c}{\makecell[c]{\textbf{1-4, 6, 8-12, 15-18}}}
      & \multicolumn{2}{c}{\textbf{5}}
      & \multicolumn{2}{c}{\textbf{14}}
      & \multicolumn{2}{c}{\textbf{20}} \\
    \cmidrule(lr){2-3}\cmidrule(lr){4-5}\cmidrule(lr){6-7}\cmidrule(lr){8-9}
      & MAE $\downarrow$ & R$^2$ $\uparrow$
      & MAE $\downarrow$ & R$^2$ $\uparrow$
      & MAE $\downarrow$ & R$^2$ $\uparrow$
      & MAE $\downarrow$ & R$^2$ $\uparrow$ \\
    \midrule
    GCN & 0.0587 & 0.8018 & 0.0637 & 0.4772 & 0.1513 & -0.5502 & 0.1687 & 0.6526 \\
    GAT & 0.0550 & 0.8159 & 0.0717 & 0.4891 & 0.1533 & -0.6171 & 0.1871 & 0.5711 \\
    GraphSAGE & 0.0461 & 0.8789 & 0.0511 & 0.7451 & 0.1473 & -0.5093 & 0.1396 & 0.7057 \\
    \midrule
    PNA & 0.0439 & \textbf{0.9051} & 0.0408 & \textbf{0.8565} & 0.1031 & 0.2984 & 0.1463 & \textbf{0.7209} \\
    SGFormer & 0.0489 & 0.8721 & \textbf{0.0397} & 0.8127 & 0.1224 & 0.0443 & 0.1439 & 0.6474 \\
    PolyNormer & \textbf{0.0407} & 0.9035 & 0.0575 & 0.6137 & 0.1976 & 0.3099 & 0.1314 & 0.6991 \\
    \midrule
    CircuitGPS & 0.0769 & 0.7443 & 0.0719 & 0.6890 & 0.2027 & 0.5314 & \textbf{0.0992} & 0.3049 \\
    CircuitGCL & 0.0606 & 0.8160 & 0.0913 & 0.5706 & \textbf{0.0804} & \textbf{0.7257} & 0.1793 & 0.6017 \\
    \bottomrule
  \end{tabular*}
\end{table*}

%% file: tables/analog_ground_regression.tex
\begin{table*}[t]
  \centering
  \setlength{\tabcolsep}{5pt}
  \caption{Performance of Different Models on Analog Circuits Ground Capacitance Node Regression Task}
  \label{tab:analog_ground_regression}
  \begin{tabular*}{\textwidth}{@{\extracolsep{\fill}}lcccccccc}
    \toprule
    \multirow{2}{*}{\textbf{Metric}}
     & \multicolumn{2}{c}{\makecell[c]{\textbf{1-4, 6, 8-12, 15-18}}}
      & \multicolumn{2}{c}{\textbf{5}}
      & \multicolumn{2}{c}{\textbf{14}}
      & \multicolumn{2}{c}{\textbf{20}} \\
    \cmidrule(lr){2-3}\cmidrule(lr){4-5}\cmidrule(lr){6-7}\cmidrule(lr){8-9}
      & MAE $\downarrow$ & R$^2$ $\uparrow$
      & MAE $\downarrow$ & R$^2$ $\uparrow$
      & MAE $\downarrow$ & R$^2$ $\uparrow$
      & MAE $\downarrow$ & R$^2$ $\uparrow$ \\
    \midrule
    GCN & 0.0272 & 0.9570 & 0.0171 & 0.4200 & 0.0216 & 0.8460 & 0.0284 & 0.8432 \\
    GAT & 0.0367 & 0.9336 & 0.0078 & 0.4879 & \textbf{0.0107} & \textbf{0.8749} & 0.0230 & 0.8270 \\
    GraphSAGE & 0.0473 & 0.8861 & 0.0267 & 0.3098 & 0.0311 & 0.8106 & 0.0371 & 0.7869 \\
    \midrule
    PNA & 0.0292 & 0.9502 & 0.0089 & 0.4214 & 0.0094 & 0.8507 & 0.0211 & 0.8423 \\
    SGFormer & 0.0682 & 0.7129 & 0.0180 & 0.3159 & 0.0430 & 0.5846 & 0.0267 & 0.6812 \\
    PolyNormer & \textbf{0.0298} & \textbf{0.9667} & 0.0155 & 0.3828 & 0.0148 & 0.8638 & 0.0281 & 0.7959 \\
    \midrule
    CircuitGPS & 0.0901 & 0.7011 & 0.0379 & -0.1327 & 0.0568 & 0.5613 & 0.0415 & 0.5002 \\
    CircuitGCL & 0.0305 & 0.9624 & \textbf{0.0073} & \textbf{0.6164} & 0.0237 & 0.8308 & \textbf{0.0193} & \textbf{0.8605} \\
    \bottomrule
  \end{tabular*}
\end{table*}

%% file: tables/sram_features.tex
\begingroup
\normalsize

\par
\noindent
\begin{minipage}{\textwidth}
\centering

\captionof{table}{Definition of SRAM Circuit Graph Node Features.}
\label{tab:sram_node_features}

\begin{tabular}{lllc}
\toprule
\textbf{Type} & \textbf{Feature} & \textbf{Definition} & \textbf{Index} \\
\midrule

\multirow{11}{*}{Device}
    & $M_{mos}$ & Multiplier of transistors & 0 \\
    & $L$ & Length of the transistor & 1 \\
    & $W$ & Width of the transistor & 2 \\
    & $M_{res}$ & Multiplier of connected resistors & 3 \\
    & $L_{res}$ & Length of resistor & 4 \\
    & $W_{res}$ & Width of resistor & 5 \\
    & $M_{cap}$ & Multiplier of connected capacitor & 6 \\
    & $L_{r}$ & Length of capacitor & 7 \\
    & $N_{r}$ & Number of capacitor fingers & 8 \\
    & $N_{p}$ & Number of ports in the device instance & 9 \\
    & $T$ & Type code of the device instance & 10 \\
\midrule

\multirow{13}{*}{Net}
    & $N_{mos}$ & Number of connected transistors & 0 \\
    & $N_{g}$ & Number of connected gate terminals & 1 \\
    & $N_{sd}$ & Number of connected source/drain terminals & 2 \\
    & $N_{b}$ & Number of connected base terminals & 3 \\
    & $W_{tot}$ & Total width of connected transistor & 4 \\
    & $L_{tot}$ & Total length of connected transistor & 5 \\
    & $N_{cap}$ & Number of connected capacitors & 6 \\
    & $Lr_{tot}$ & Total length of connected capacitors & 7 \\
    & $Nr_{tot}$ & Total Number of connected capacitor fingers & 8 \\
    & $N_{res}$ & Number of connected resistors & 9 \\
    & $W_{tot,res}$ & Total width of connected resistors & 10 \\
    & $L_{tot,res}$ & Total length of connected resistors & 11 \\
    & $N_{port}$ & Number of connected ports & 12 \\
\midrule

Pin & -- & Pin types (G/D/S/B for MOS) & 0 \\

\bottomrule
\end{tabular}

\end{minipage}
\par

\endgroup

%% file: tables/analog_effective_each_distribution.tex
\begin{figure}[H]  
  \centering       
  \setlength{\tabcolsep}{2pt}  
  
  \begin{minipage}[t]{0.19\textwidth}  
    \centering
    \includegraphics[width=\linewidth]{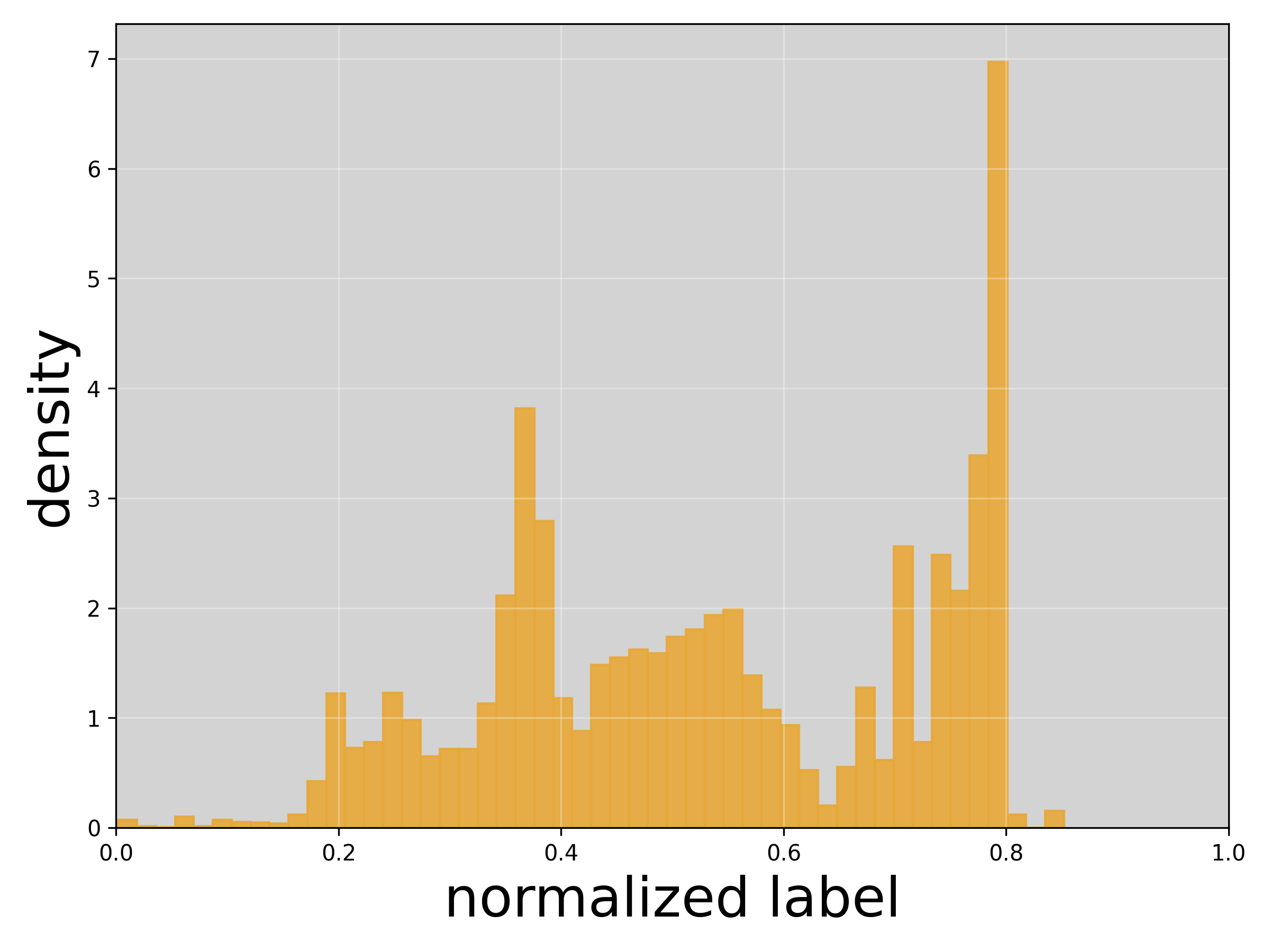}
    \subcaption{Case 1}  
  \end{minipage}
  \hfill  
  \begin{minipage}[t]{0.19\textwidth}
    \centering
    \includegraphics[width=\linewidth]{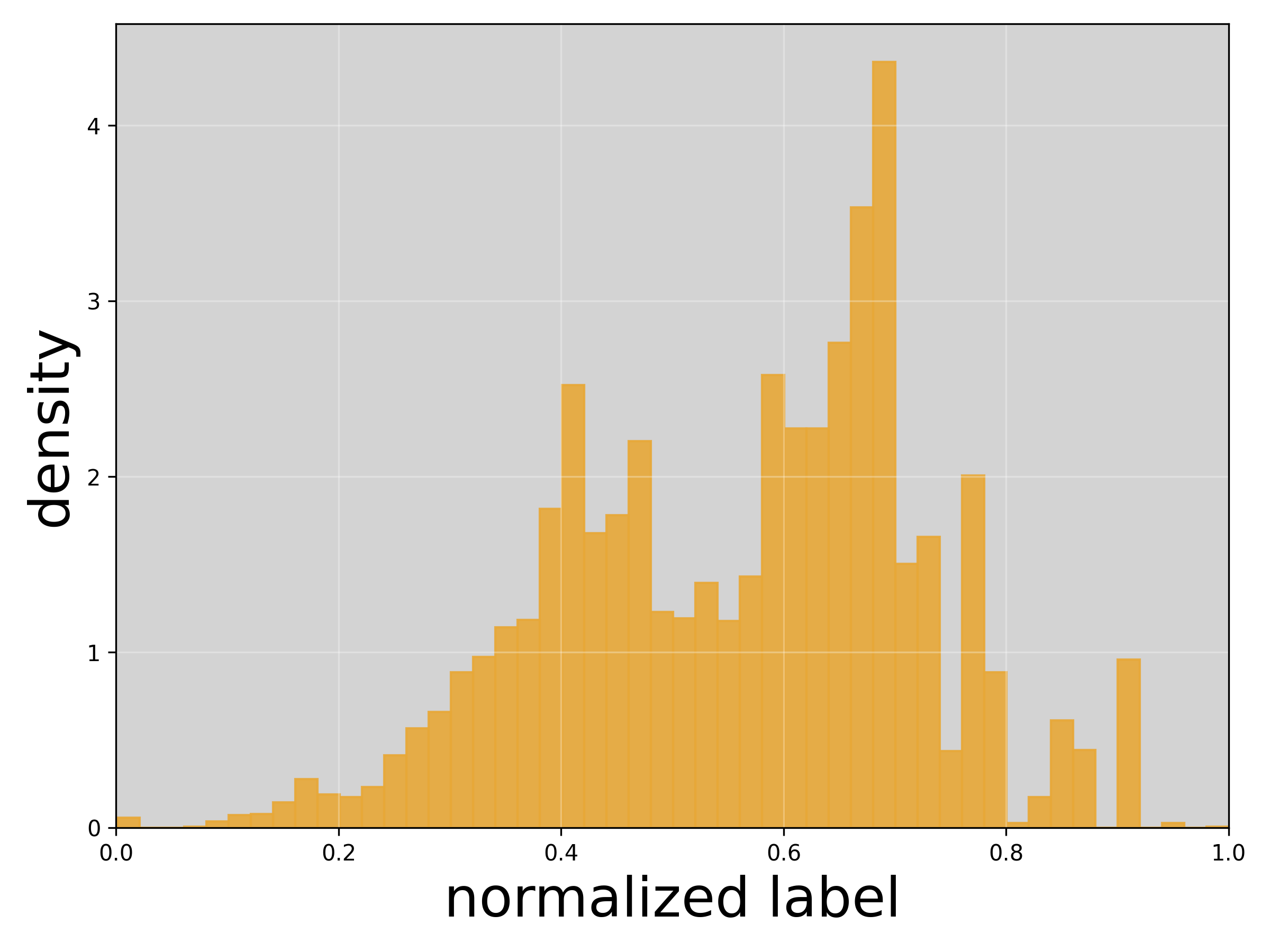}
    \subcaption{Case 2}
  \end{minipage}
  \hfill
  \begin{minipage}[t]{0.19\textwidth}
    \centering
    \includegraphics[width=\linewidth]{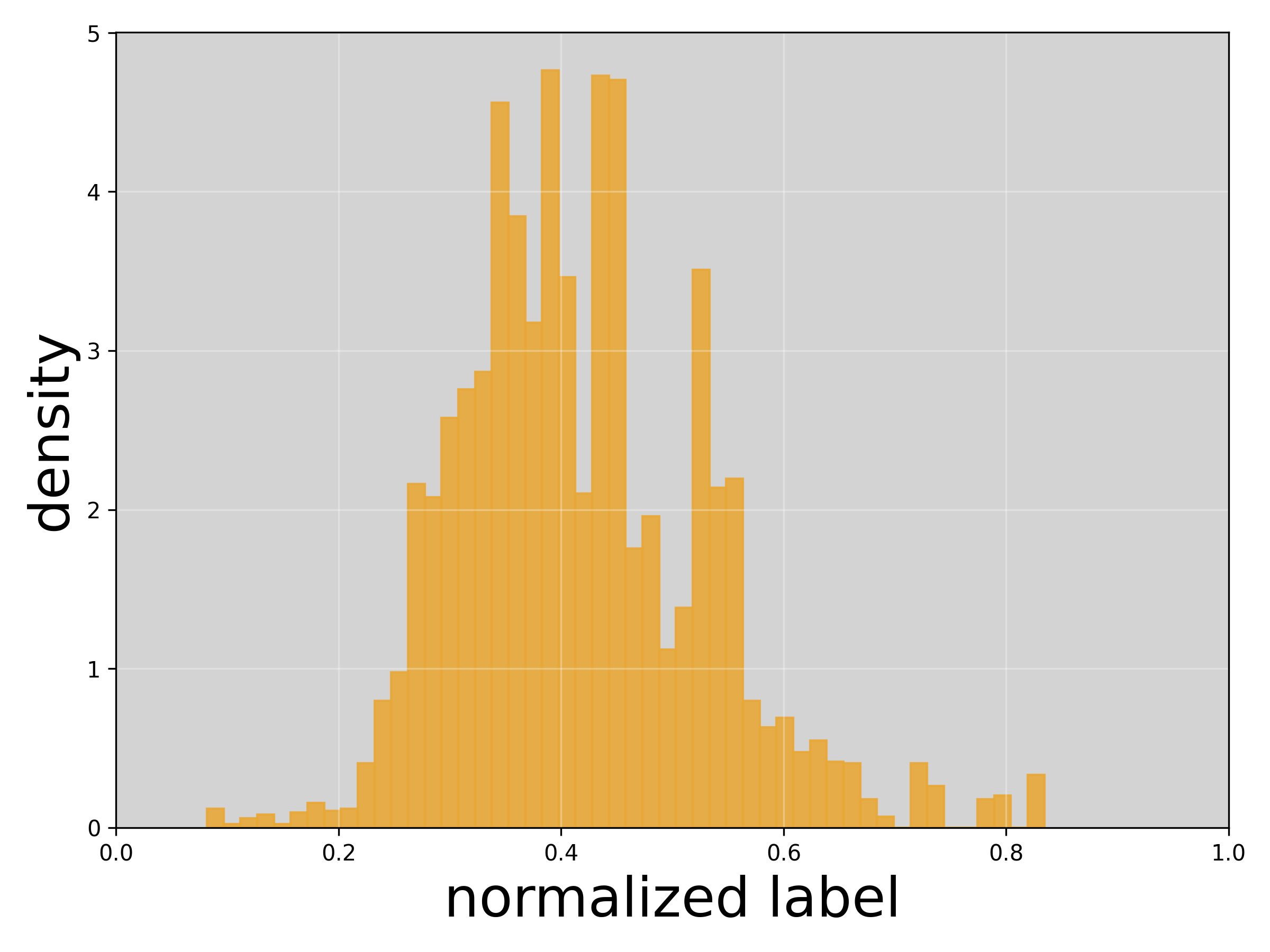}
    \subcaption{Case 3}
  \end{minipage}
  \hfill
  \begin{minipage}[t]{0.19\textwidth}
    \centering
    \includegraphics[width=\linewidth]{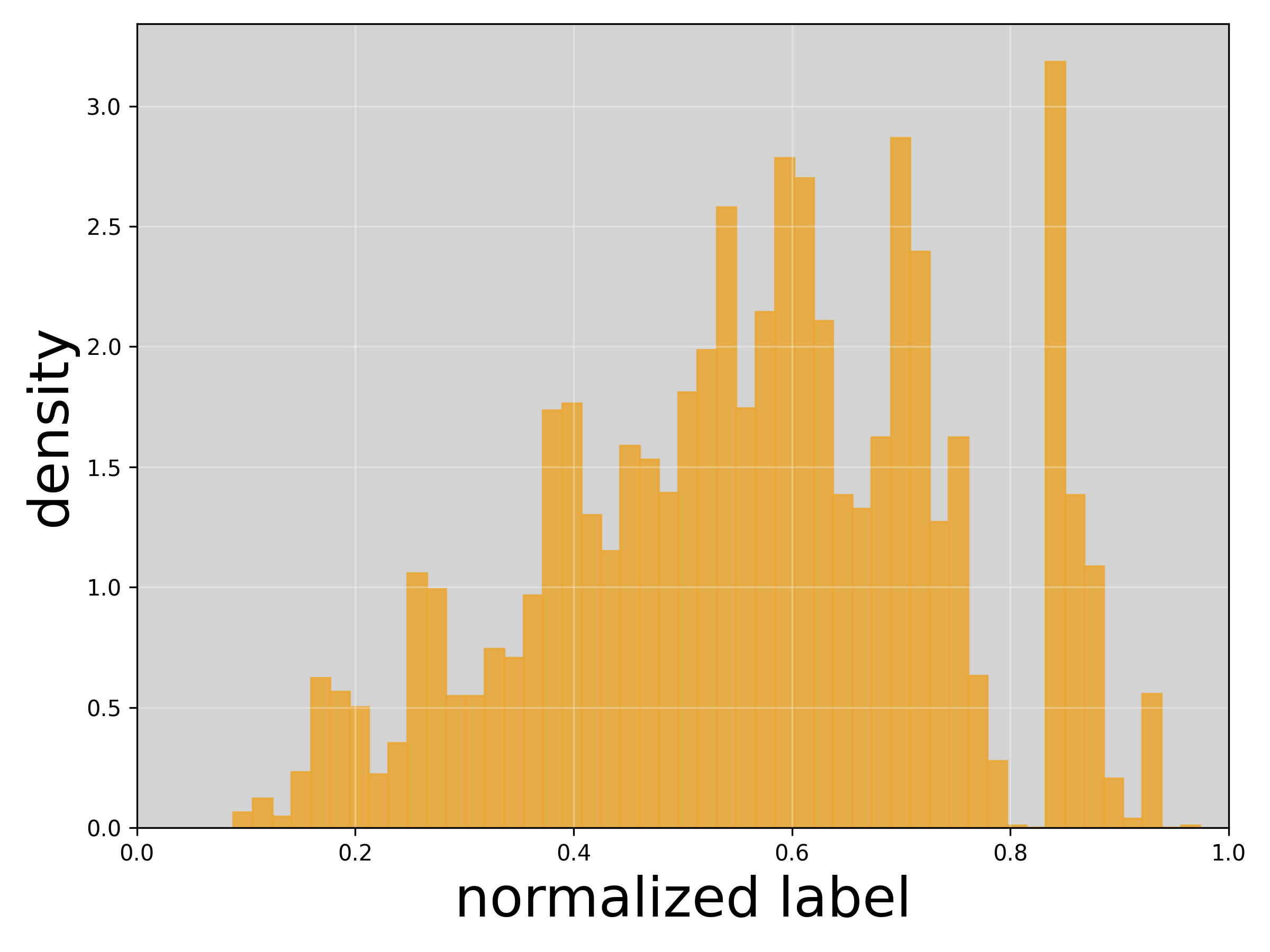}
    \subcaption{Case 4}
  \end{minipage}
  \hfill
  \begin{minipage}[t]{0.19\textwidth}
    \centering
    \includegraphics[width=\linewidth]{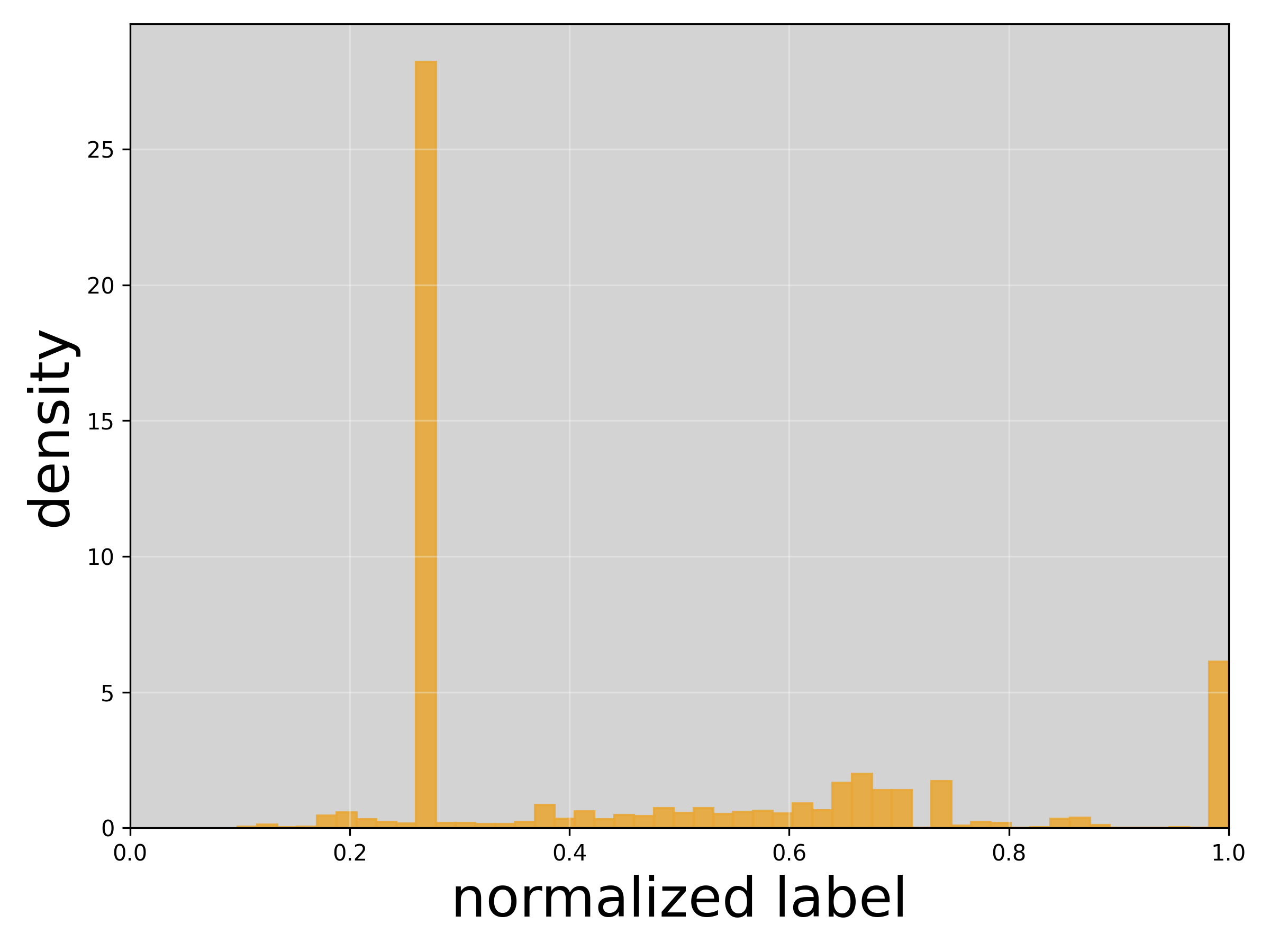}
    \subcaption{Case 5}
  \end{minipage}
  
  \vspace{5pt}  
  
  \begin{minipage}[t]{0.19\textwidth}
    \centering
    \includegraphics[width=\linewidth]{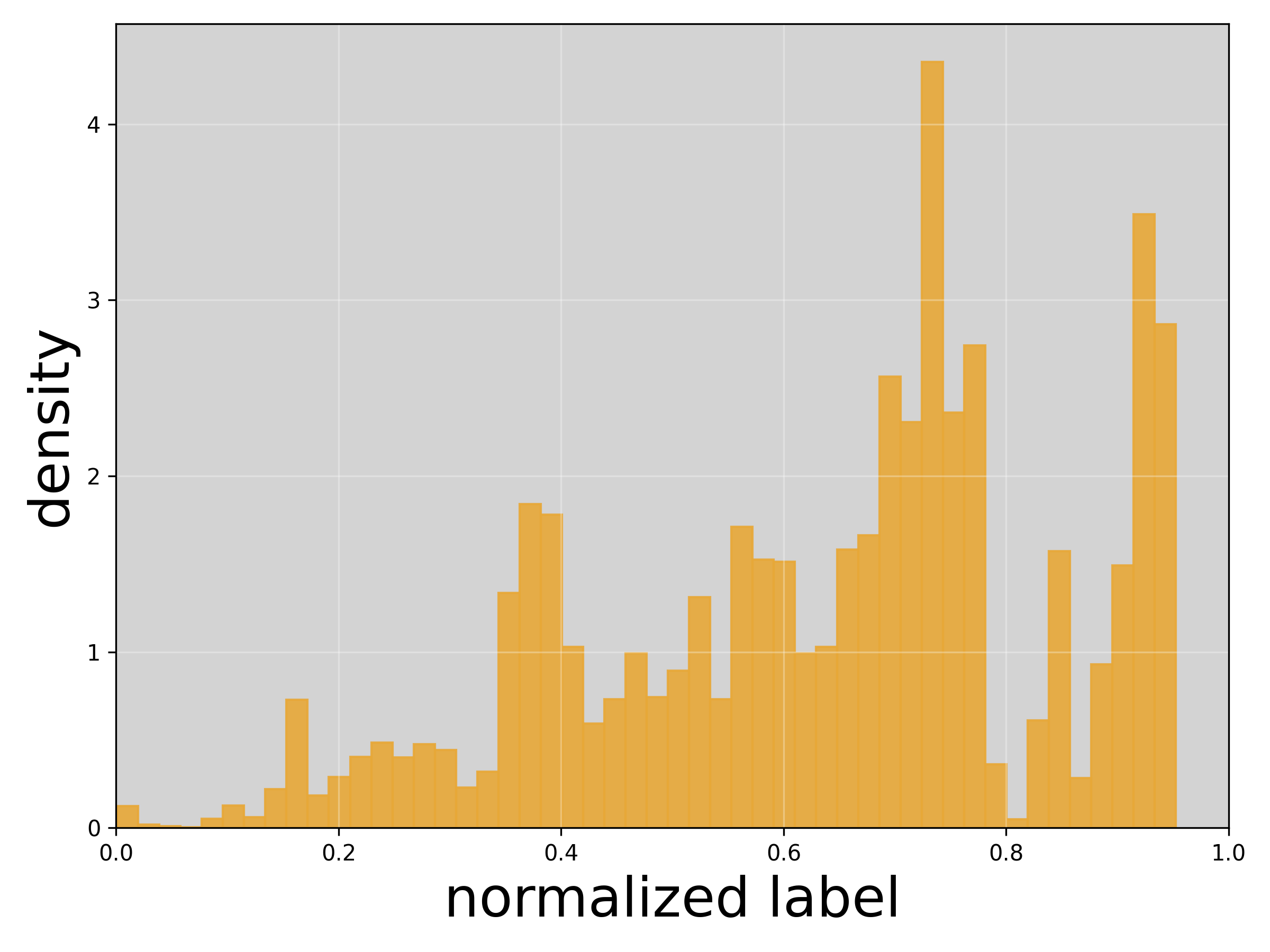}
    \subcaption{Case 6}
  \end{minipage}
  \hfill
  \begin{minipage}[t]{0.19\textwidth}
    \centering
    \includegraphics[width=\linewidth]{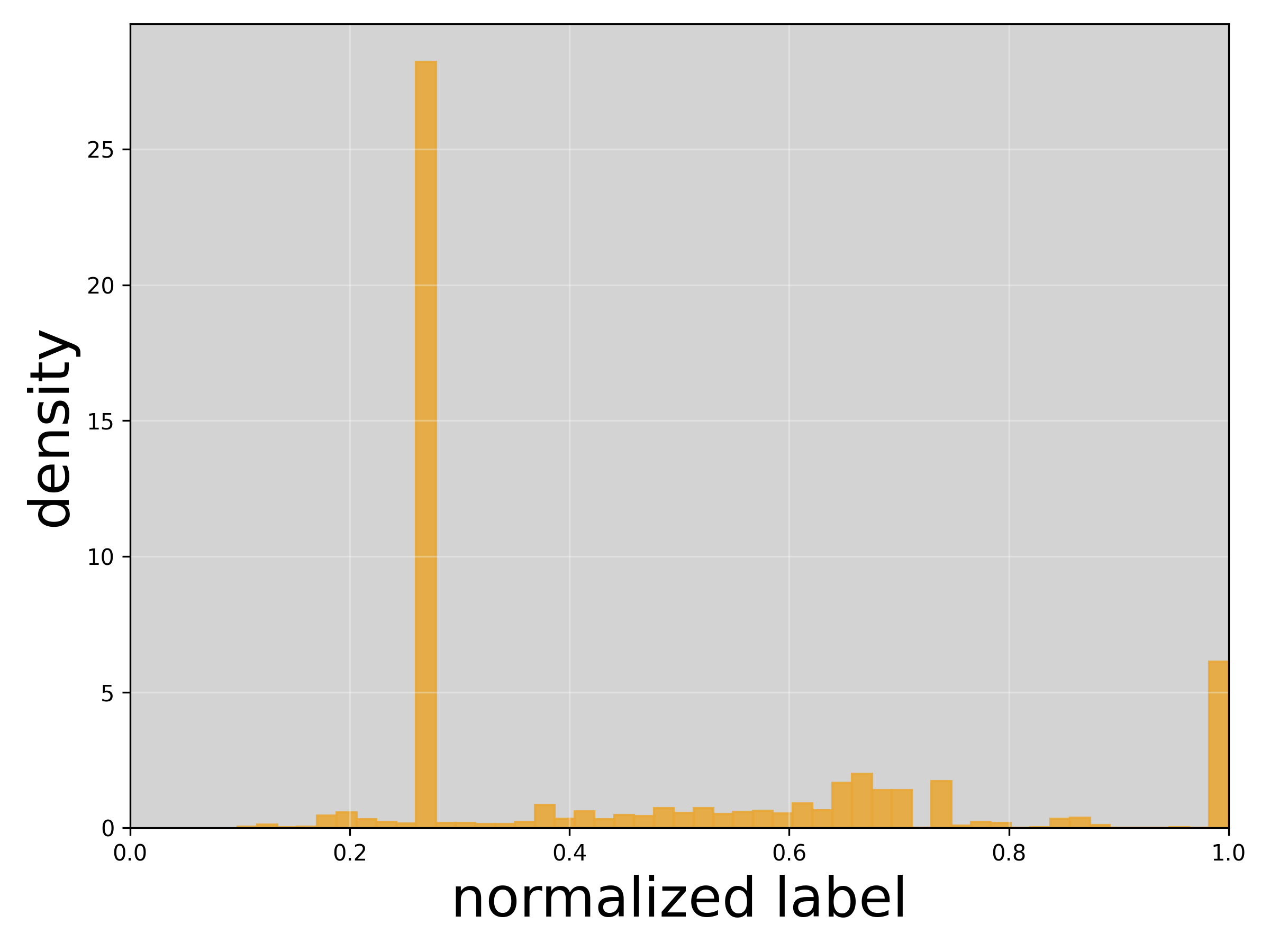}
    \subcaption{Case 7}
  \end{minipage}
  \hfill
  \begin{minipage}[t]{0.19\textwidth}
    \centering
    \includegraphics[width=\linewidth]{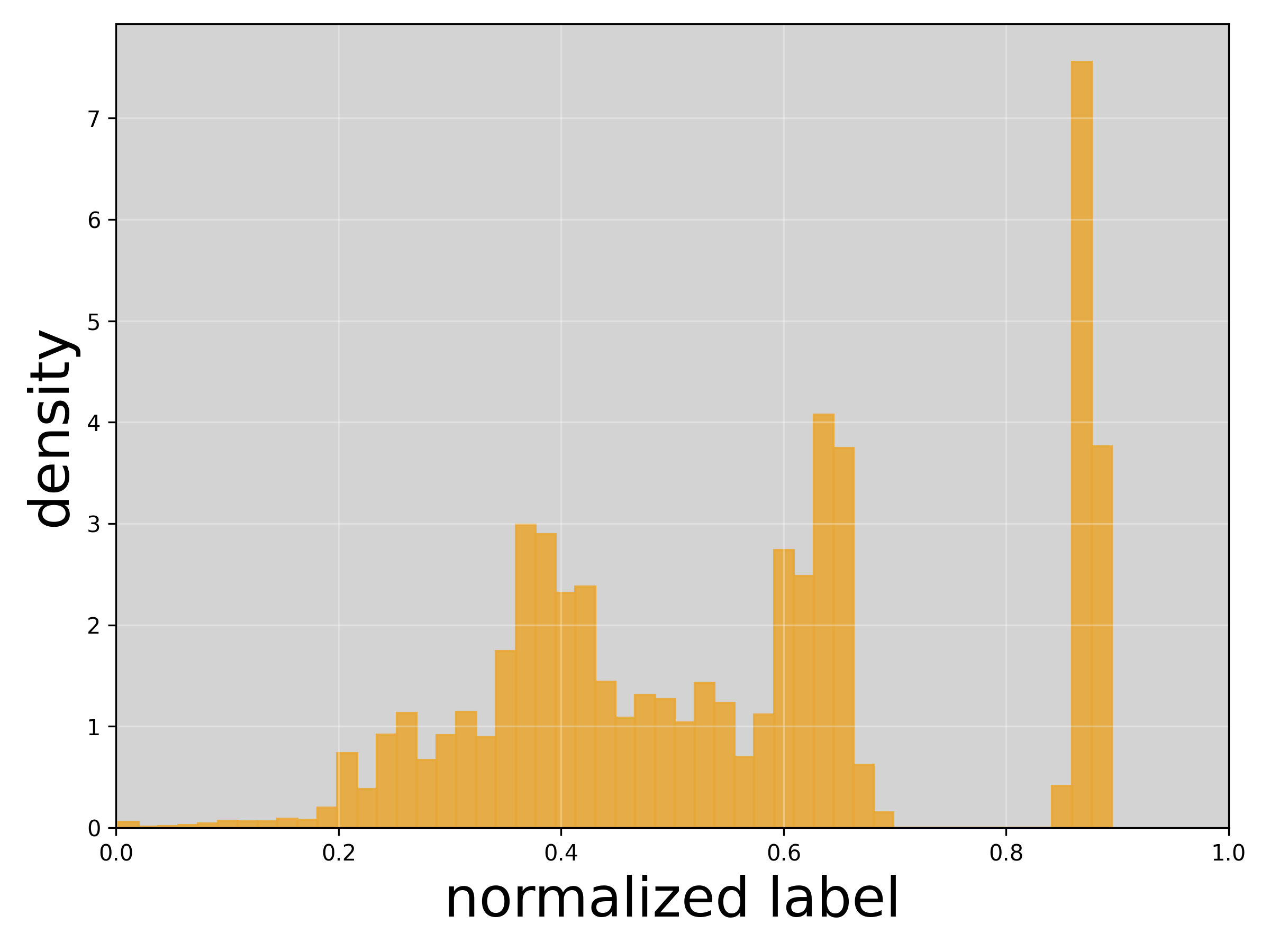}
    \subcaption{Case 8}
  \end{minipage}
  \hfill
  \begin{minipage}[t]{0.19\textwidth}
    \centering
    \includegraphics[width=\linewidth]{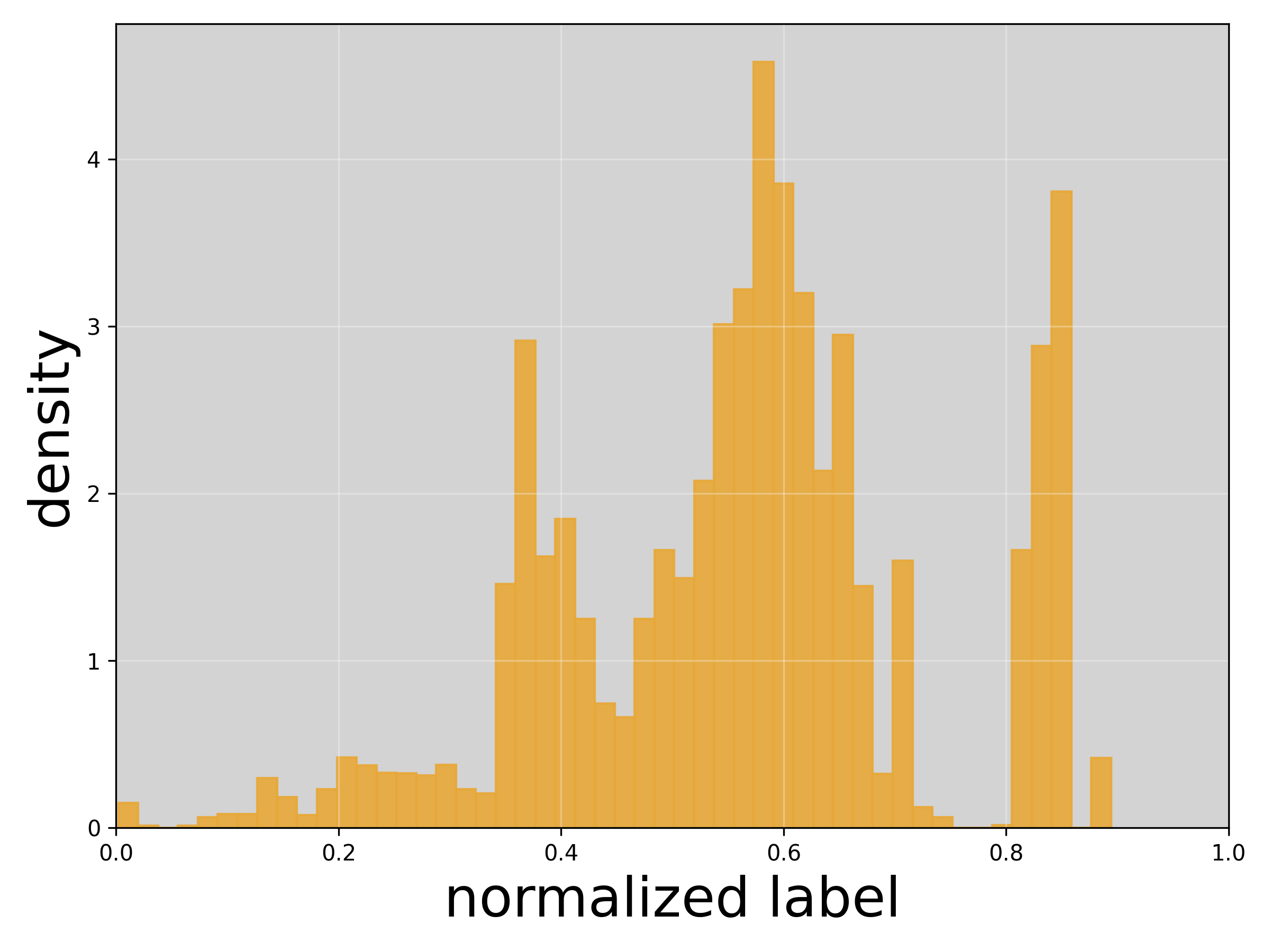}
    \subcaption{Case 9}
  \end{minipage}
  \hfill
  \begin{minipage}[t]{0.19\textwidth}
    \centering
    \includegraphics[width=\linewidth]{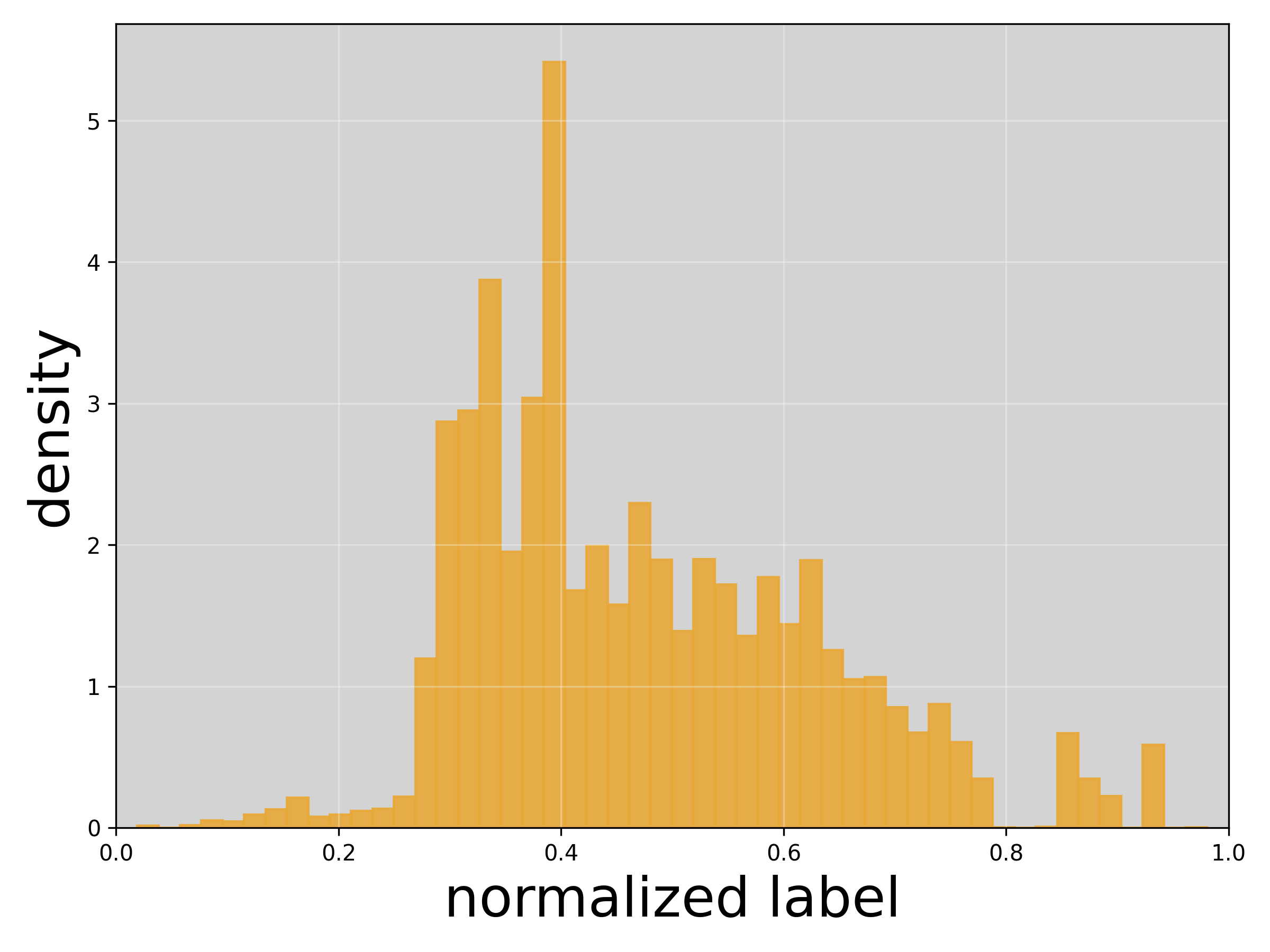}
    \subcaption{Case 10}
  \end{minipage}
  
  \vspace{5pt}
  
  \begin{minipage}[t]{0.19\textwidth}
    \centering
    \includegraphics[width=\linewidth]{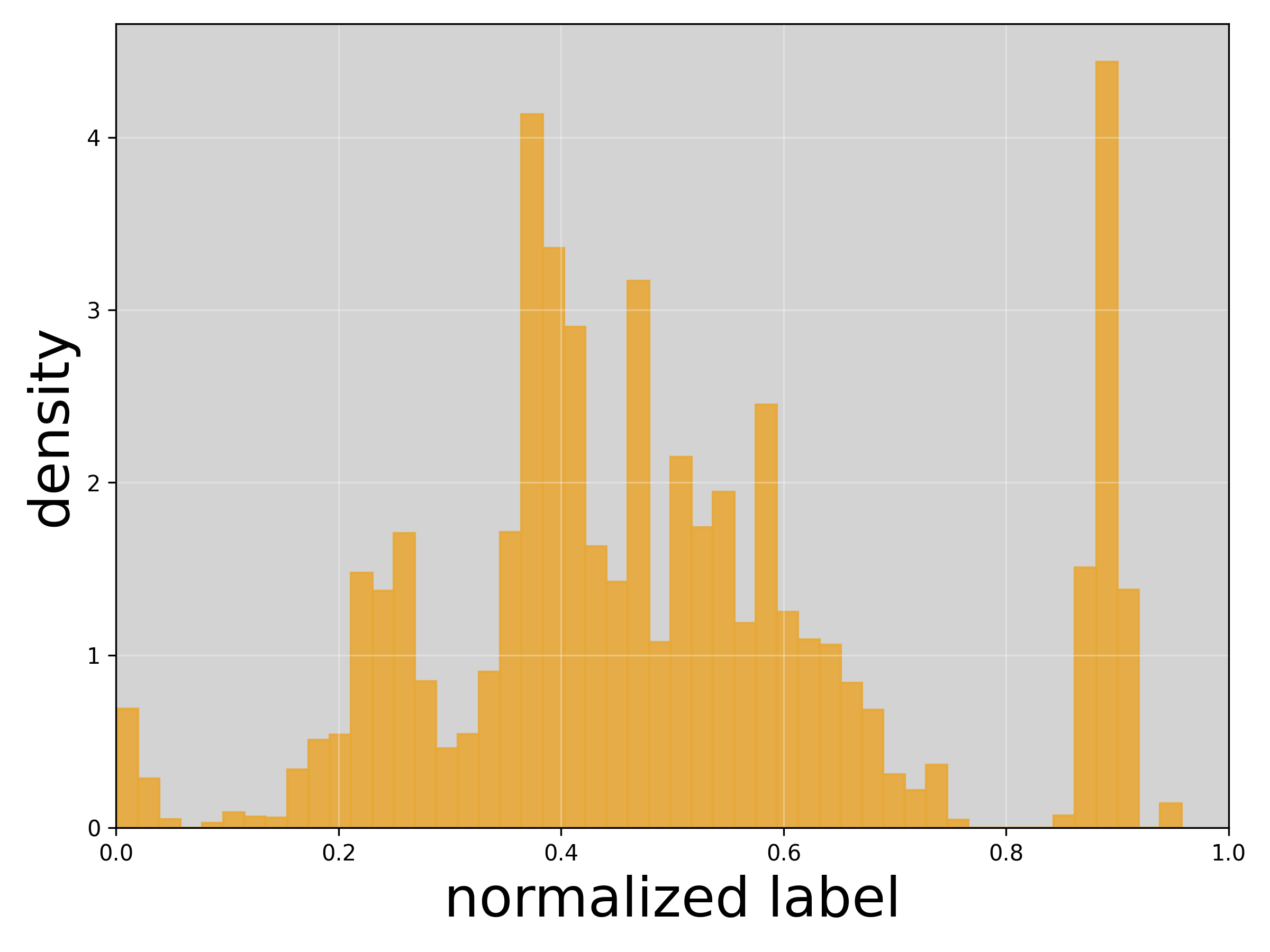}
    \subcaption{Case 11}
  \end{minipage}
  \hfill
  \begin{minipage}[t]{0.19\textwidth}
    \centering
    \includegraphics[width=\linewidth]{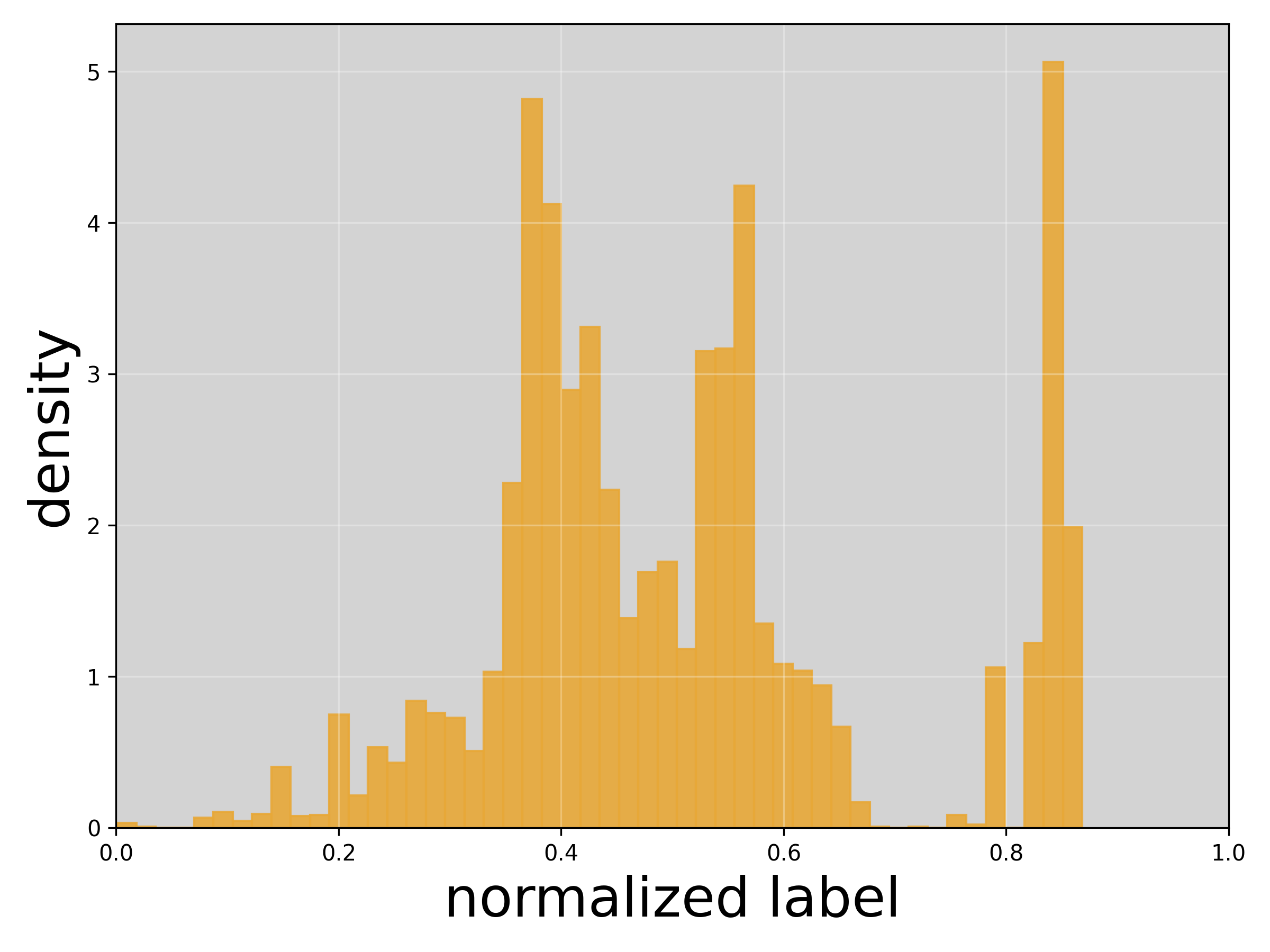}
    \subcaption{Case 12}
  \end{minipage}
  \hfill
  \begin{minipage}[t]{0.19\textwidth}
    \centering
    \includegraphics[width=\linewidth]{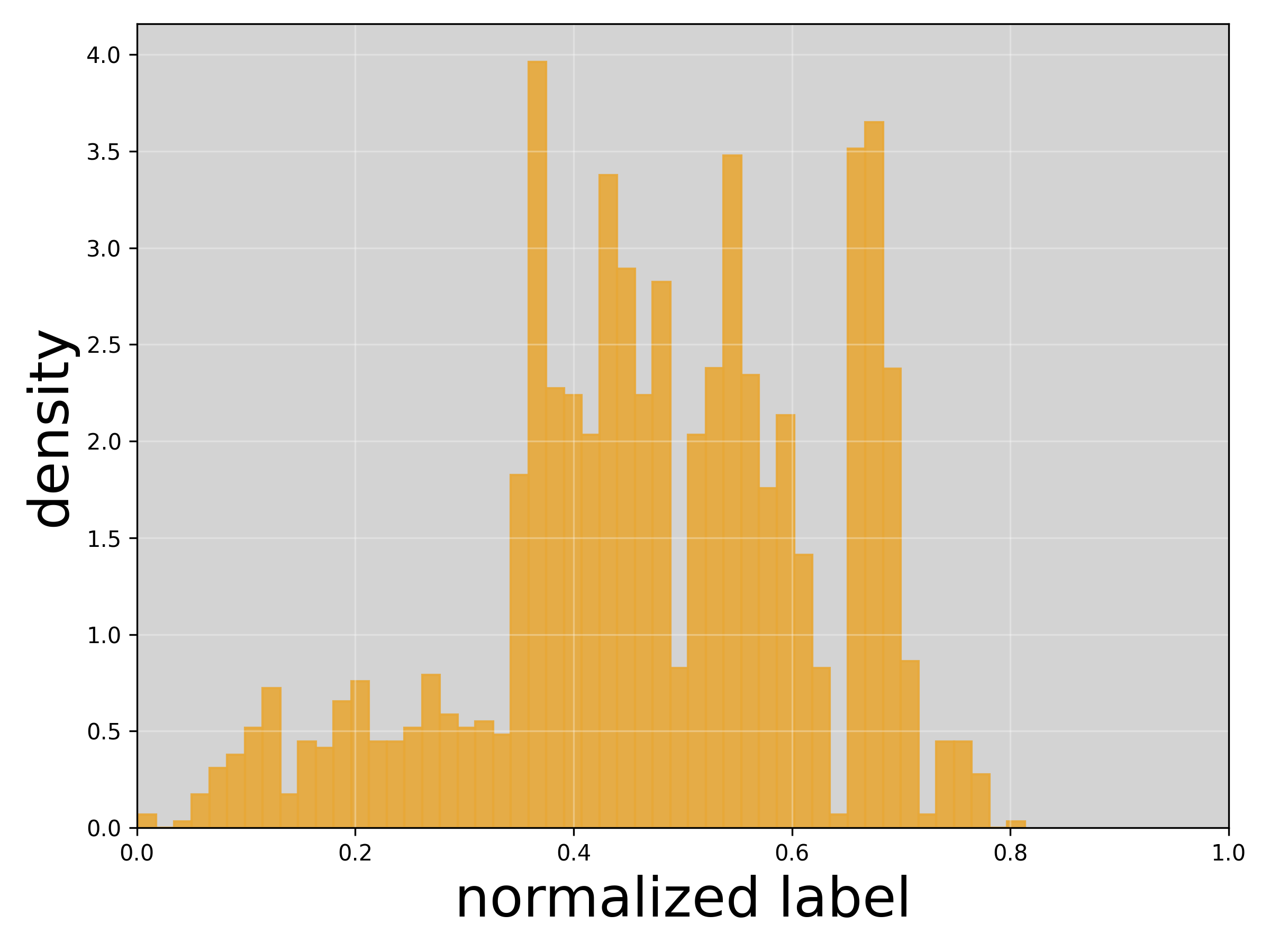}
    \subcaption{Case 13}
  \end{minipage}
  \hfill
  \begin{minipage}[t]{0.19\textwidth}
    \centering
    \includegraphics[width=\linewidth]{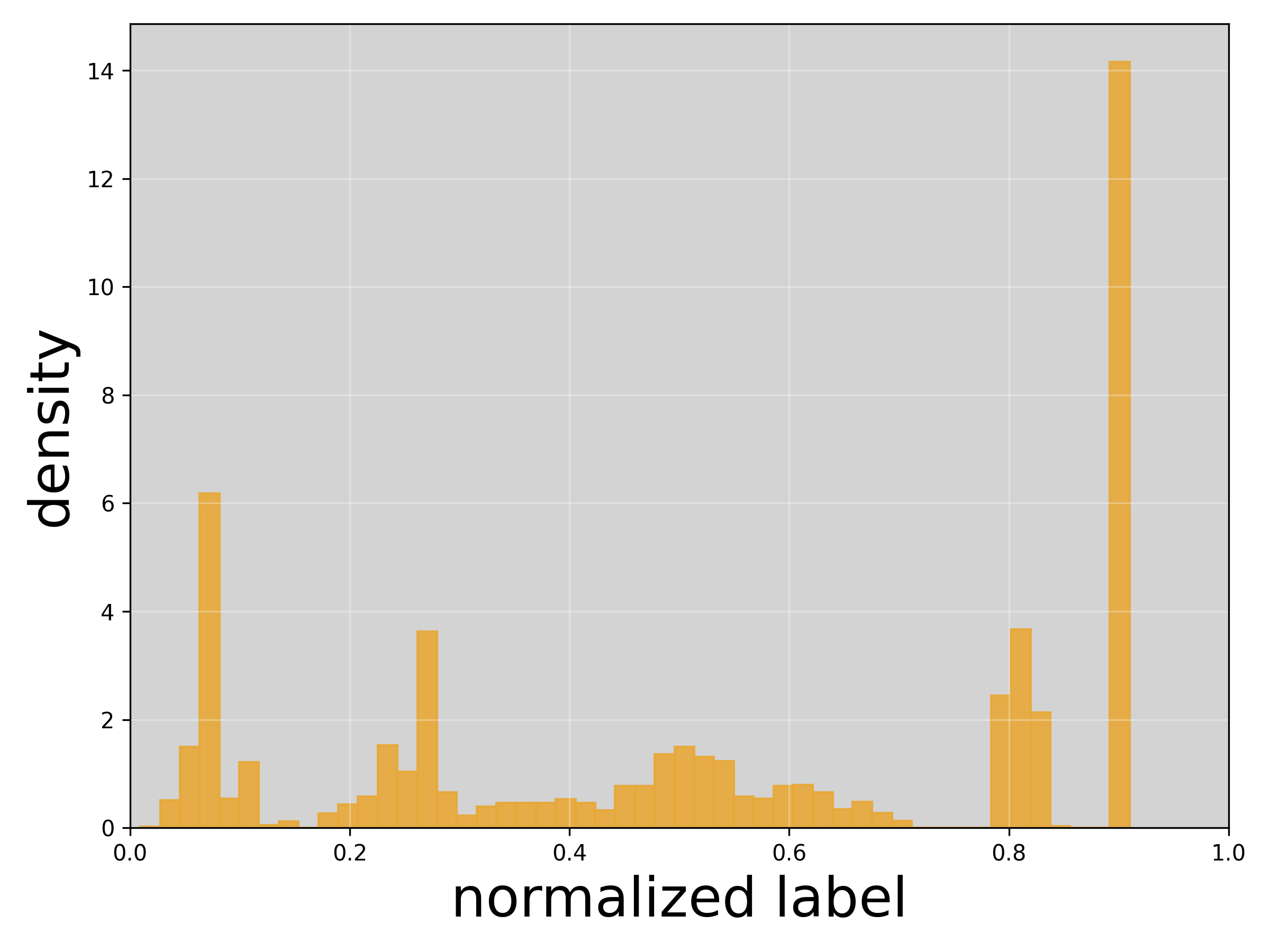}
    \subcaption{Case 14}
  \end{minipage}
  \hfill
  \begin{minipage}[t]{0.19\textwidth}
    \centering
    \includegraphics[width=\linewidth]{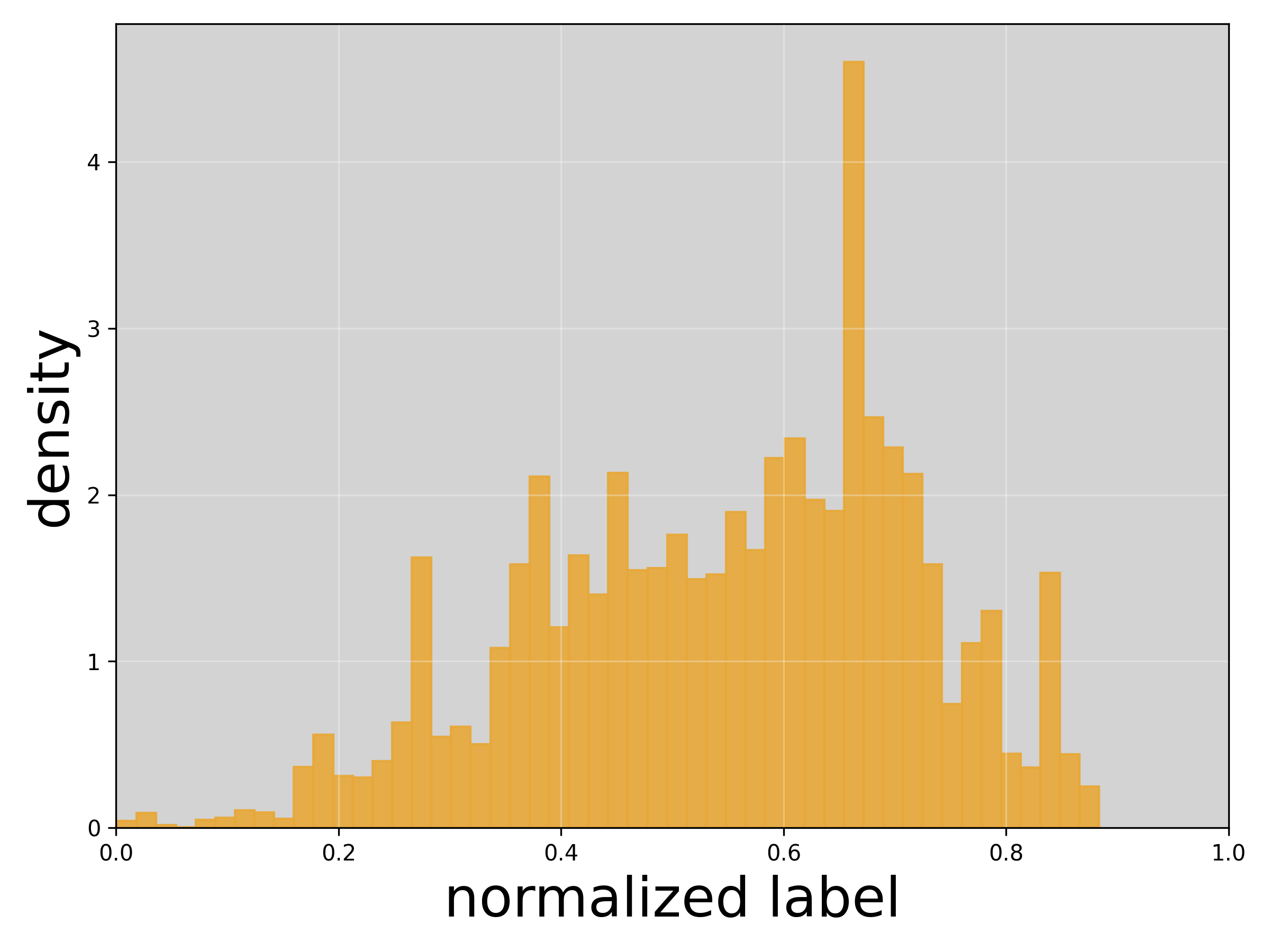}
    \subcaption{Case 15}
  \end{minipage}
  
  \vspace{5pt}
  
  \begin{minipage}[t]{0.19\textwidth}
    \centering
    \includegraphics[width=\linewidth]{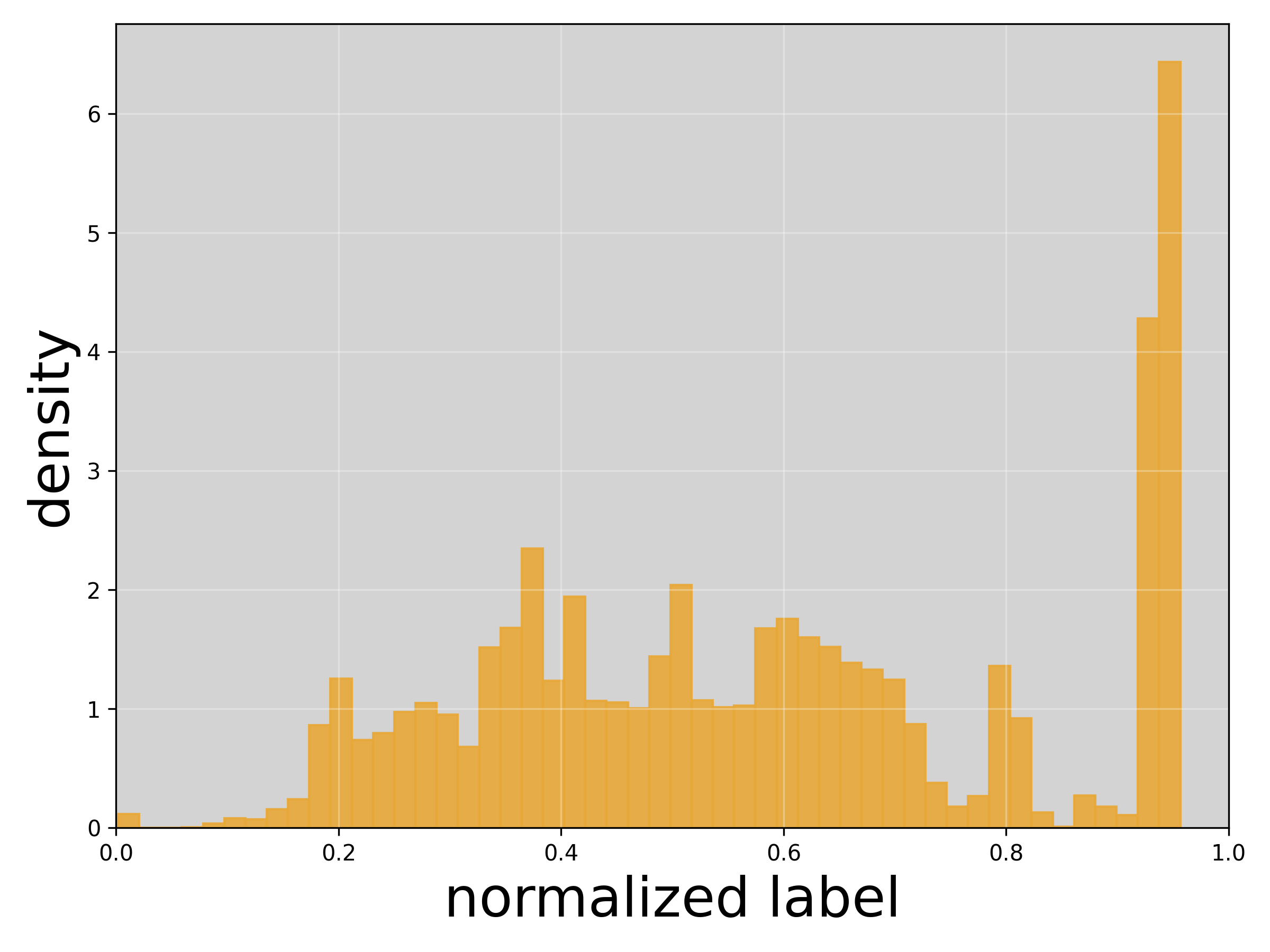}
    \subcaption{Case 16}
  \end{minipage}
  \hfill
  \begin{minipage}[t]{0.19\textwidth}
    \centering
    \includegraphics[width=\linewidth]{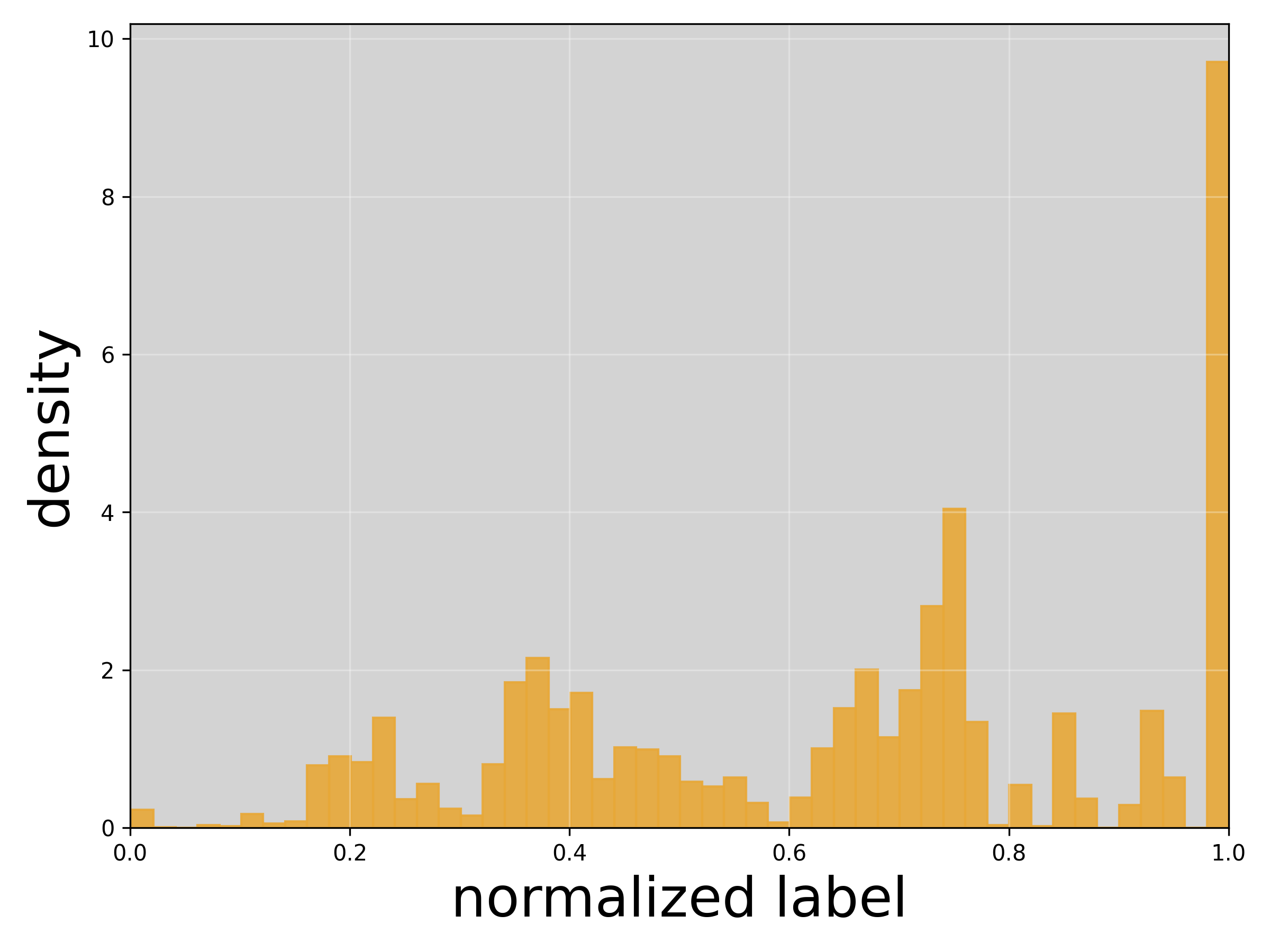}
    \subcaption{Case 17}
  \end{minipage}
  \hfill
  \begin{minipage}[t]{0.19\textwidth}
    \centering
    \includegraphics[width=\linewidth]{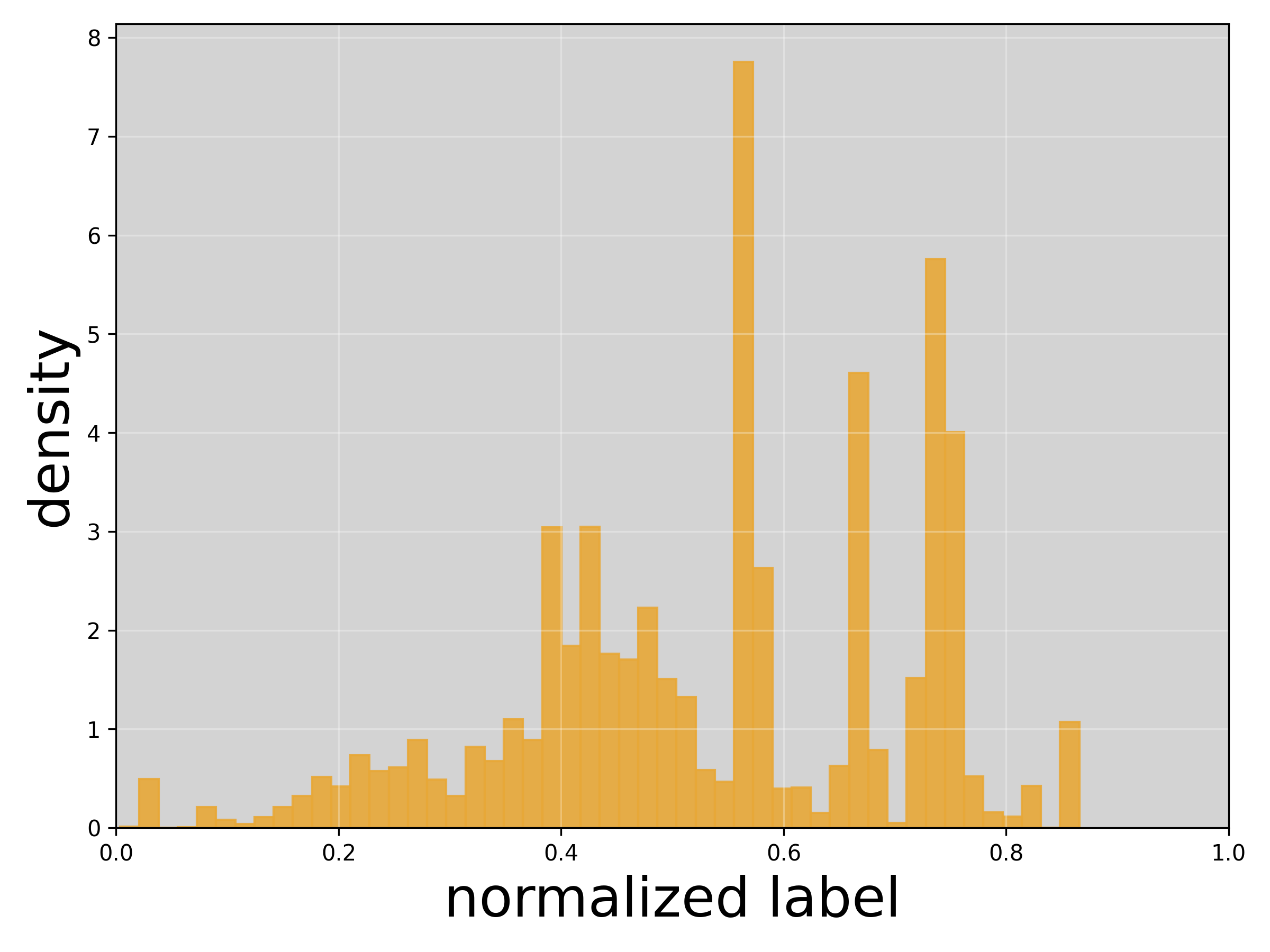}
    \subcaption{Case 18}
  \end{minipage}
  \hfill
  \begin{minipage}[t]{0.19\textwidth}
    \centering
    \includegraphics[width=\linewidth]{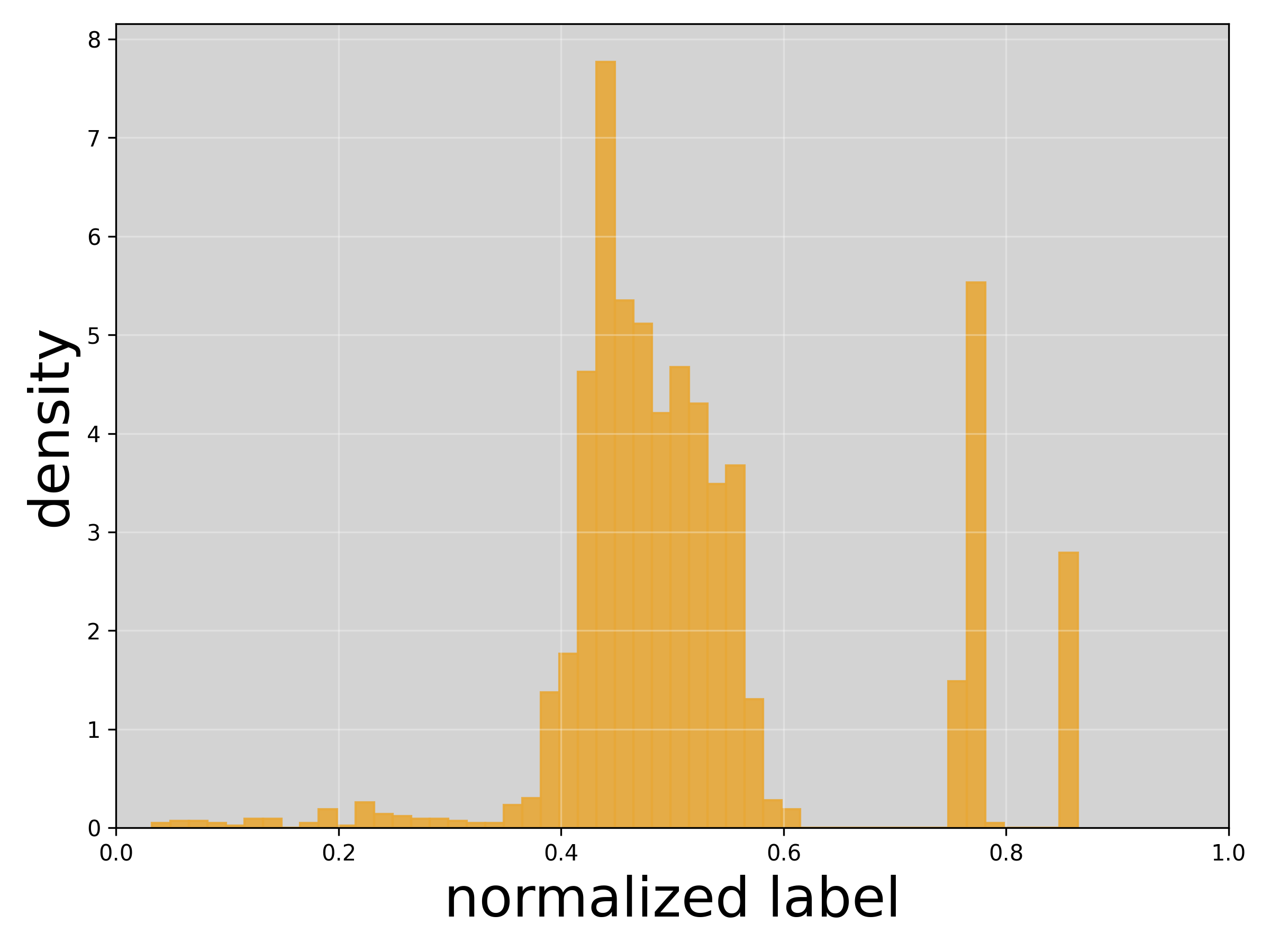}
    \subcaption{Case 19}
  \end{minipage}
  \hfill
  \begin{minipage}[t]{0.19\textwidth}
    \centering
    \includegraphics[width=\linewidth]{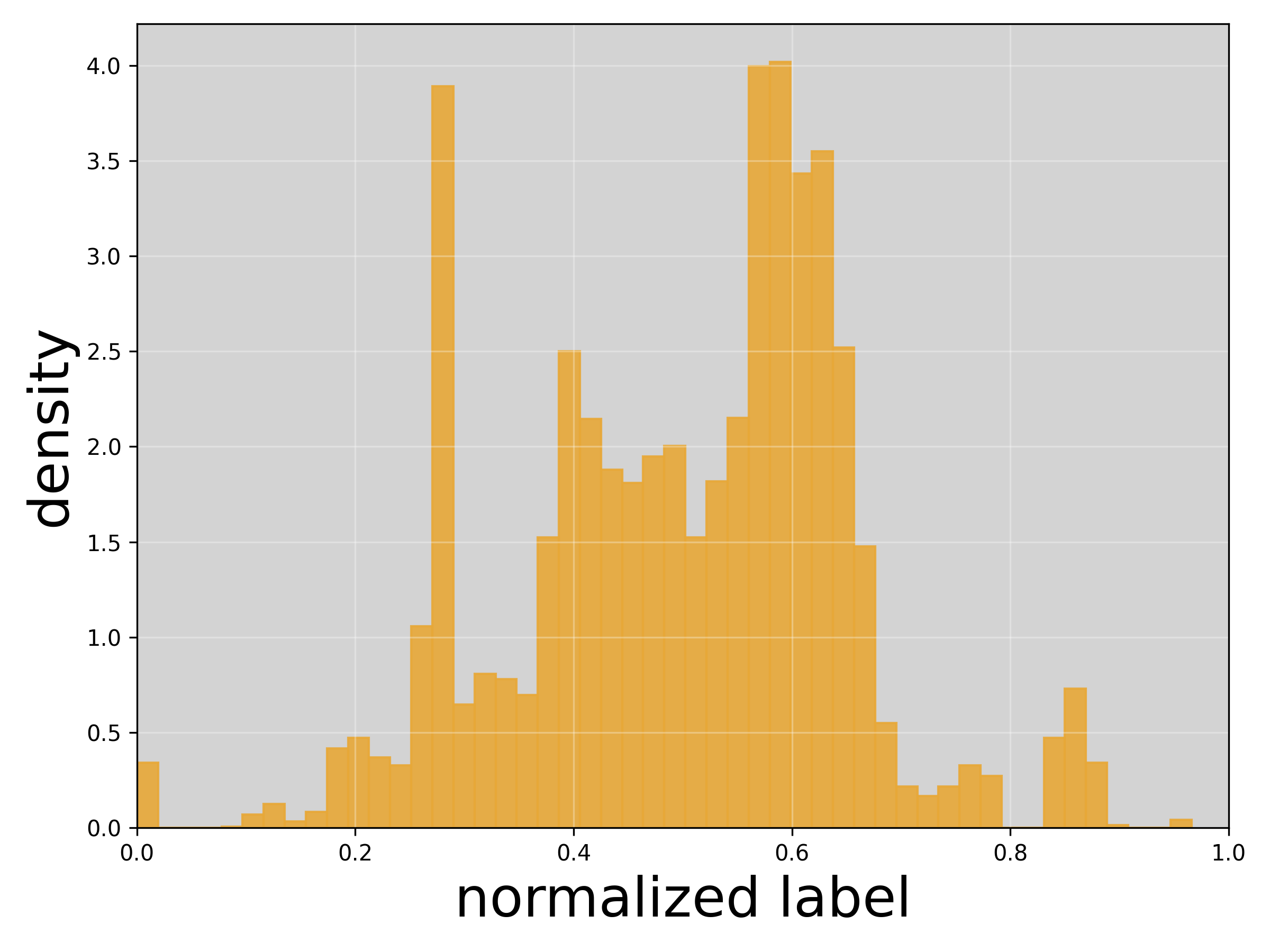}
    \subcaption{Case 20}
  \end{minipage}

  \caption{Overview of 20 Analog Circuit Effective Resistance Label Distribution}
  \label{fig:analog_effective_each_distribution}
\end{figure}

%% file: tables/sram_coupling_each_distribution.tex
\begin{figure*}[htbp]  
  \centering
  \begin{minipage}[t]{0.16\linewidth}  
    \centering
    \includegraphics[width=\linewidth]{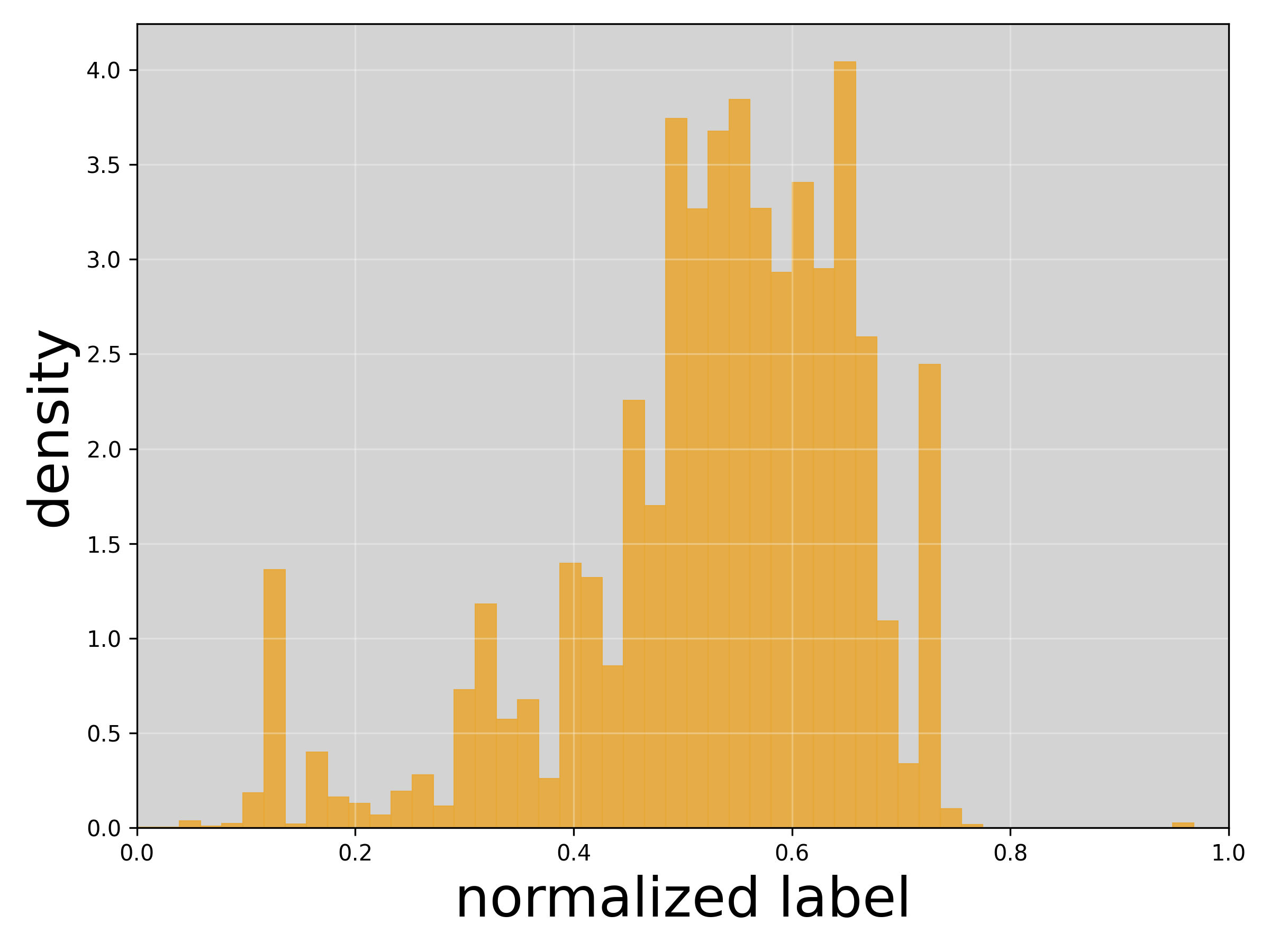}
 \subcaption{Array\_128\_32\_8t}
  \end{minipage}
  \begin{minipage}[t]{0.16\linewidth}  
    \centering
    \includegraphics[width=\linewidth]{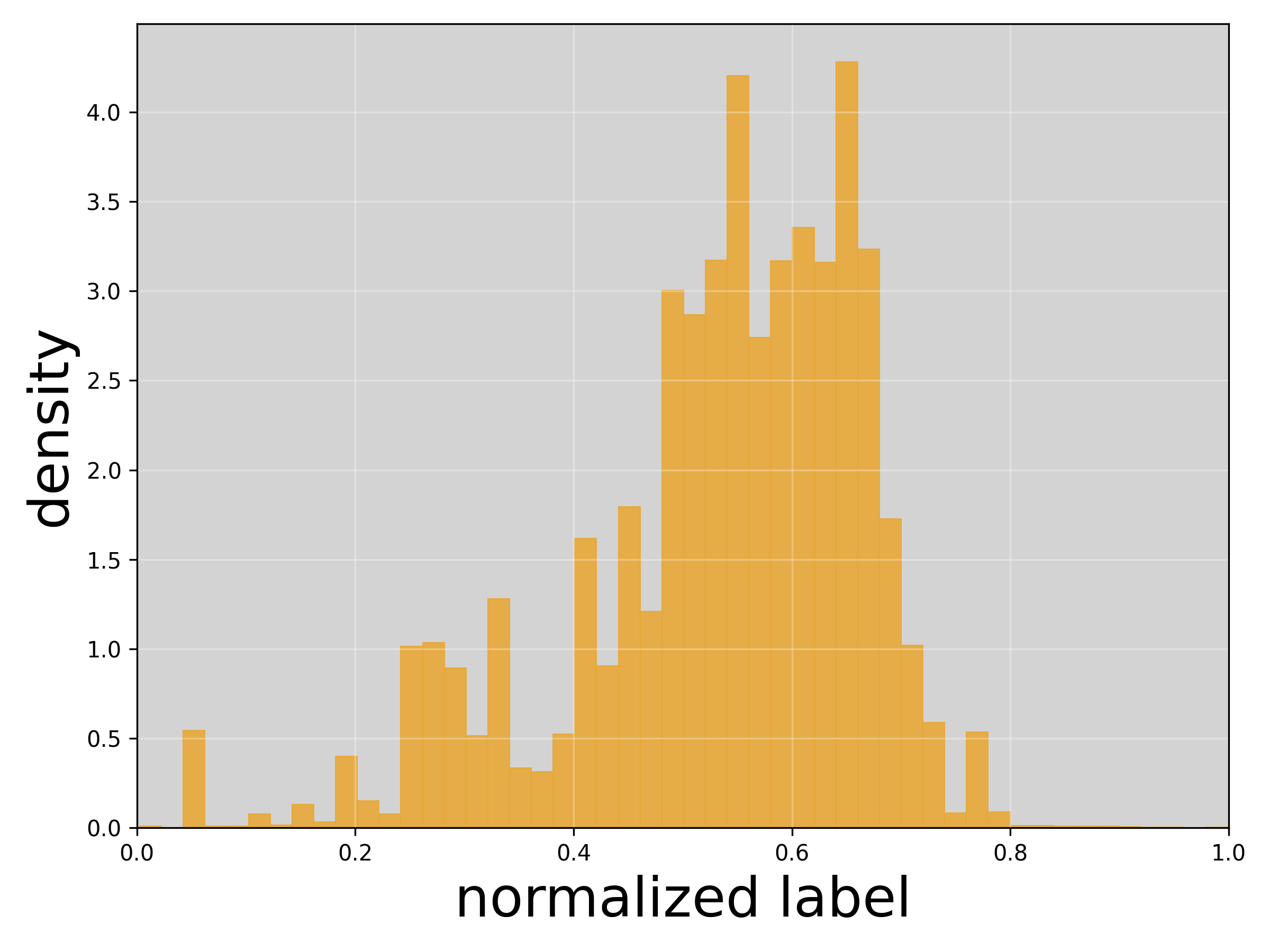}
\subcaption{Digtime}
  \end{minipage}
  \begin{minipage}[t]{0.16\linewidth}  
    \centering
    \includegraphics[width=\linewidth]{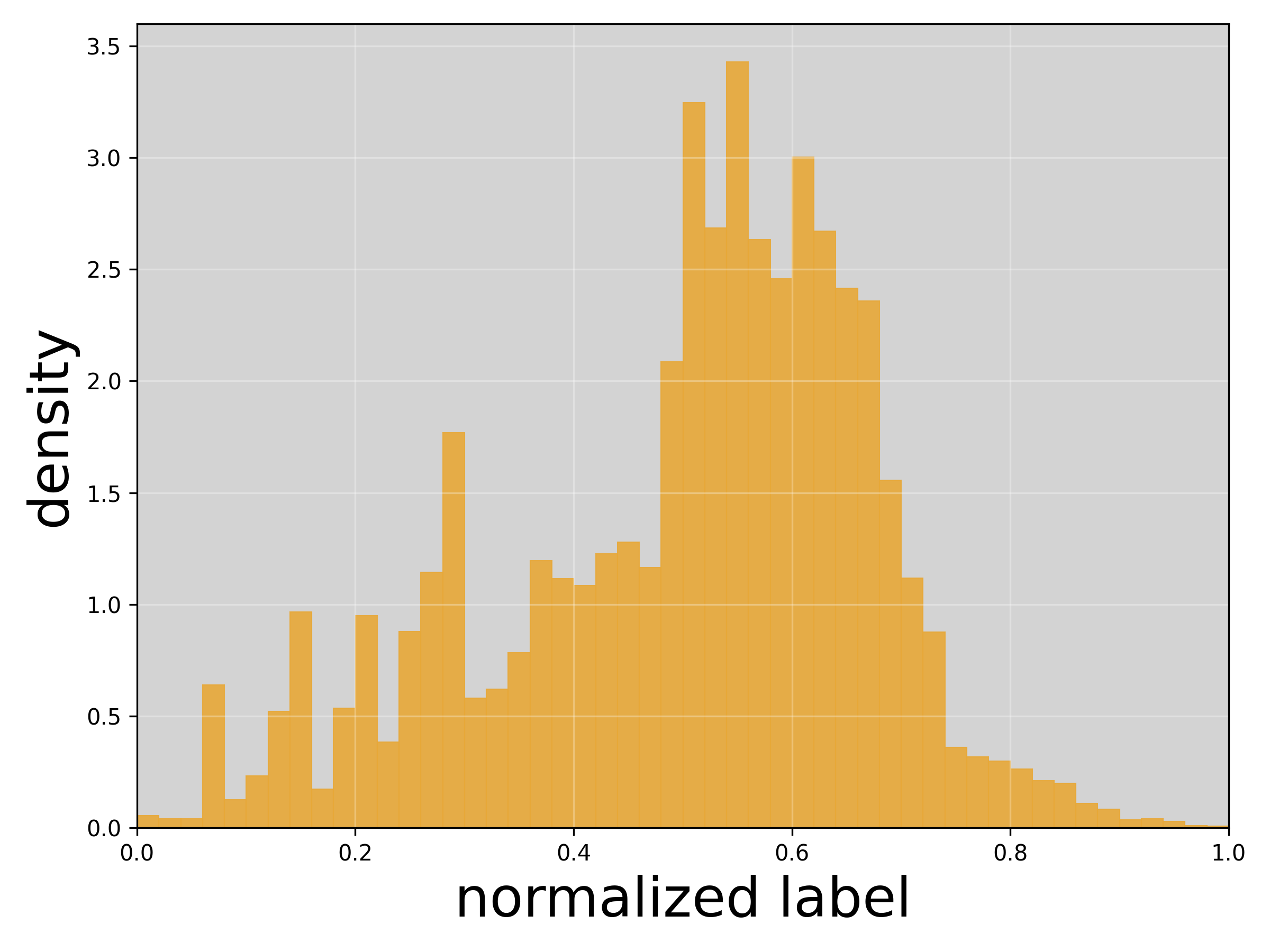}
\subcaption{Sandwich}
  \end{minipage}
  \begin{minipage}[t]{0.16\linewidth}  
    \centering
    \includegraphics[width=\linewidth]{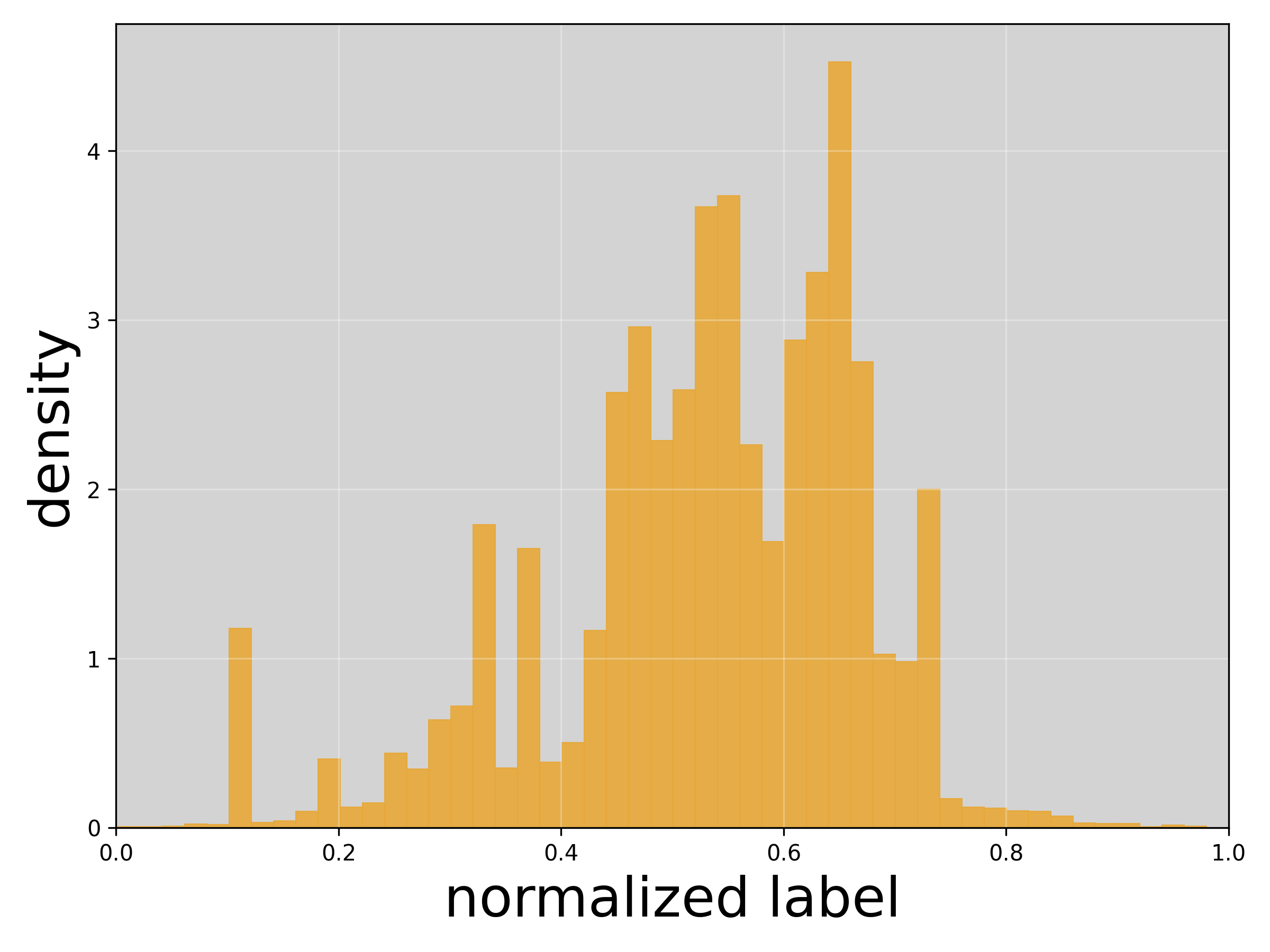}
    \subcaption{SSRAM}
  \end{minipage}
  \begin{minipage}[t]{0.16\linewidth}  
    \centering
    \includegraphics[width=\linewidth]{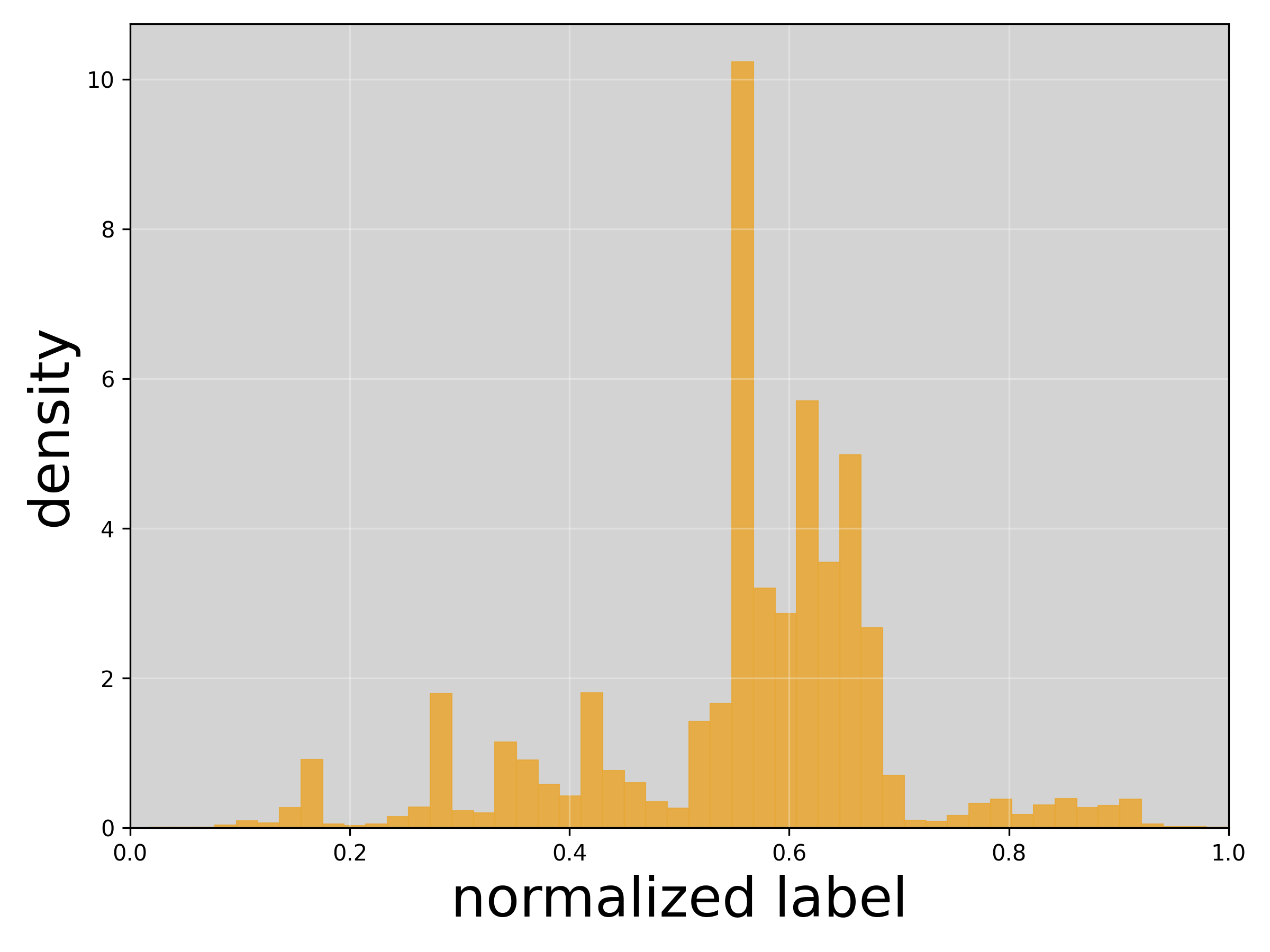}
    \subcaption{Timing\_Ctrl}
  \end{minipage}
  \begin{minipage}[t]{0.16\linewidth}  
    \centering
    \includegraphics[width=\linewidth]{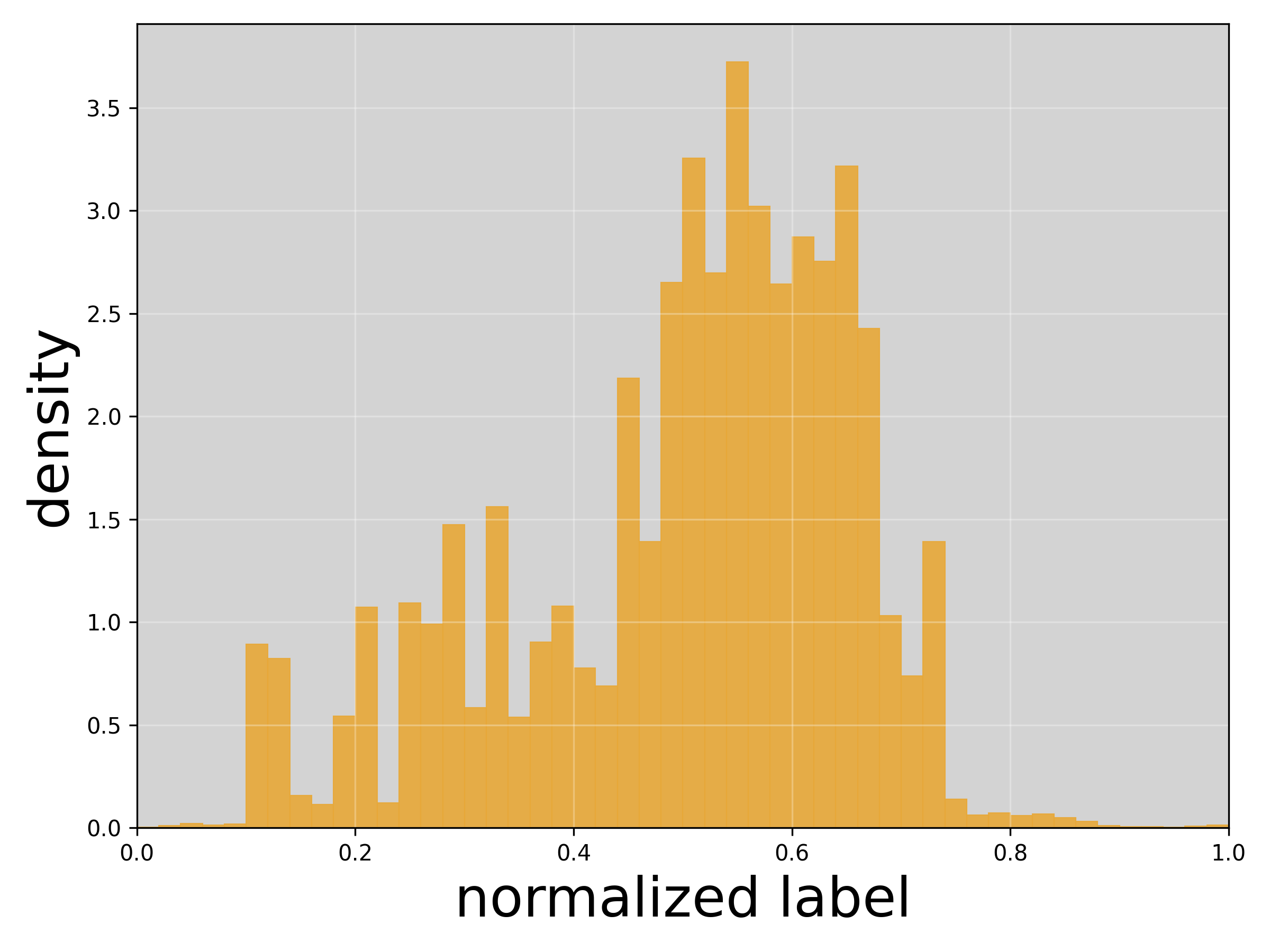}
    \subcaption{Ultra8t}
  \end{minipage}
    \caption{Overview of 6 SRAM Circuit Coupling Capacitance Label Distribution}
  \label{fig:sram_coupling_each_distribution}
  \hfill
\end{figure*}

%% file: tables/sram_resistance_each_distribution.tex
\begin{figure*}[htbp]  
  \centering
  \begin{minipage}[t]{0.16\linewidth}  
    \centering
    \includegraphics[width=\linewidth]{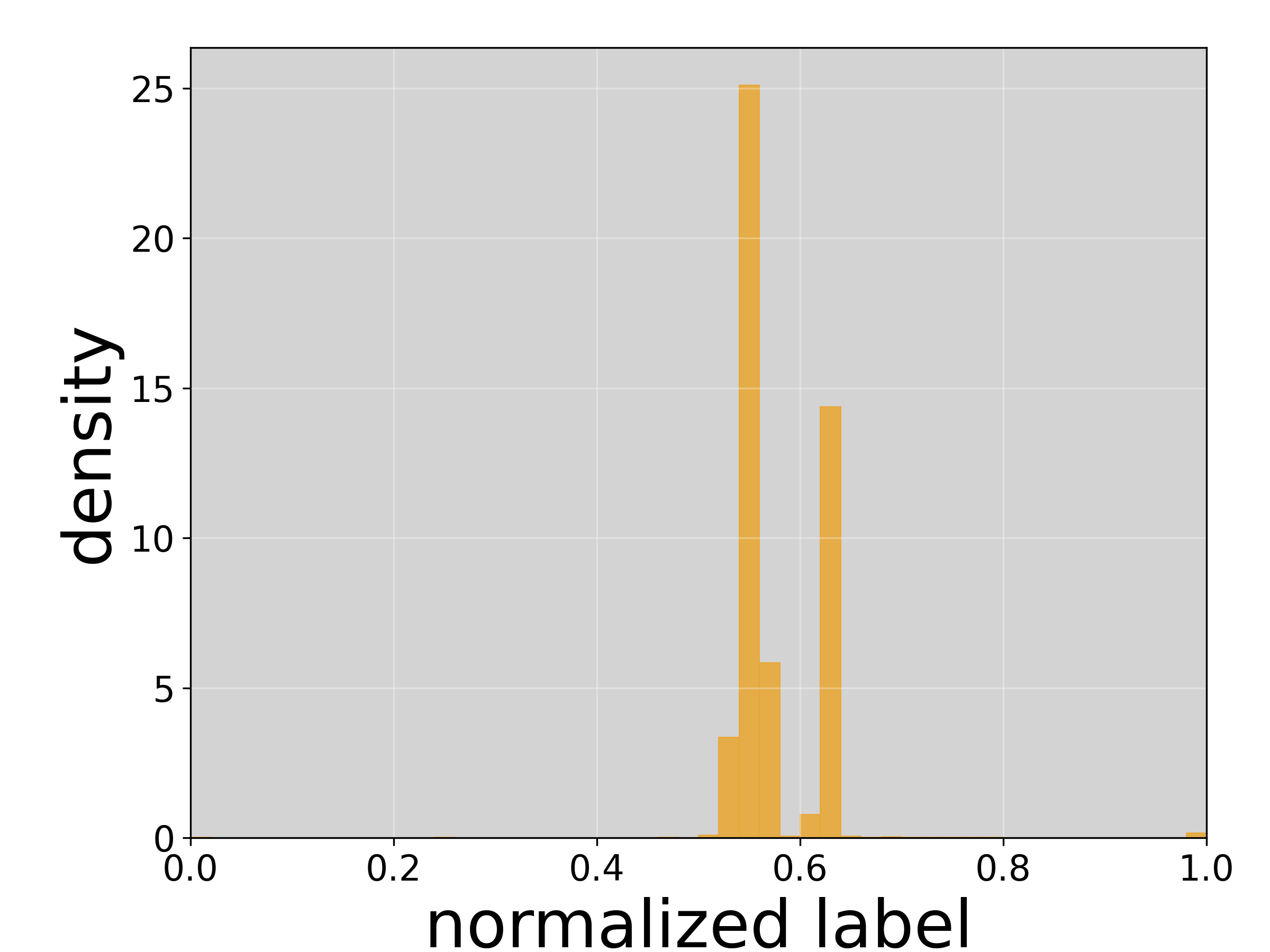}
 \subcaption{Array\_128\_32\_8t}
  \end{minipage}
  \begin{minipage}[t]{0.16\linewidth}  
    \centering
    \includegraphics[width=\linewidth]{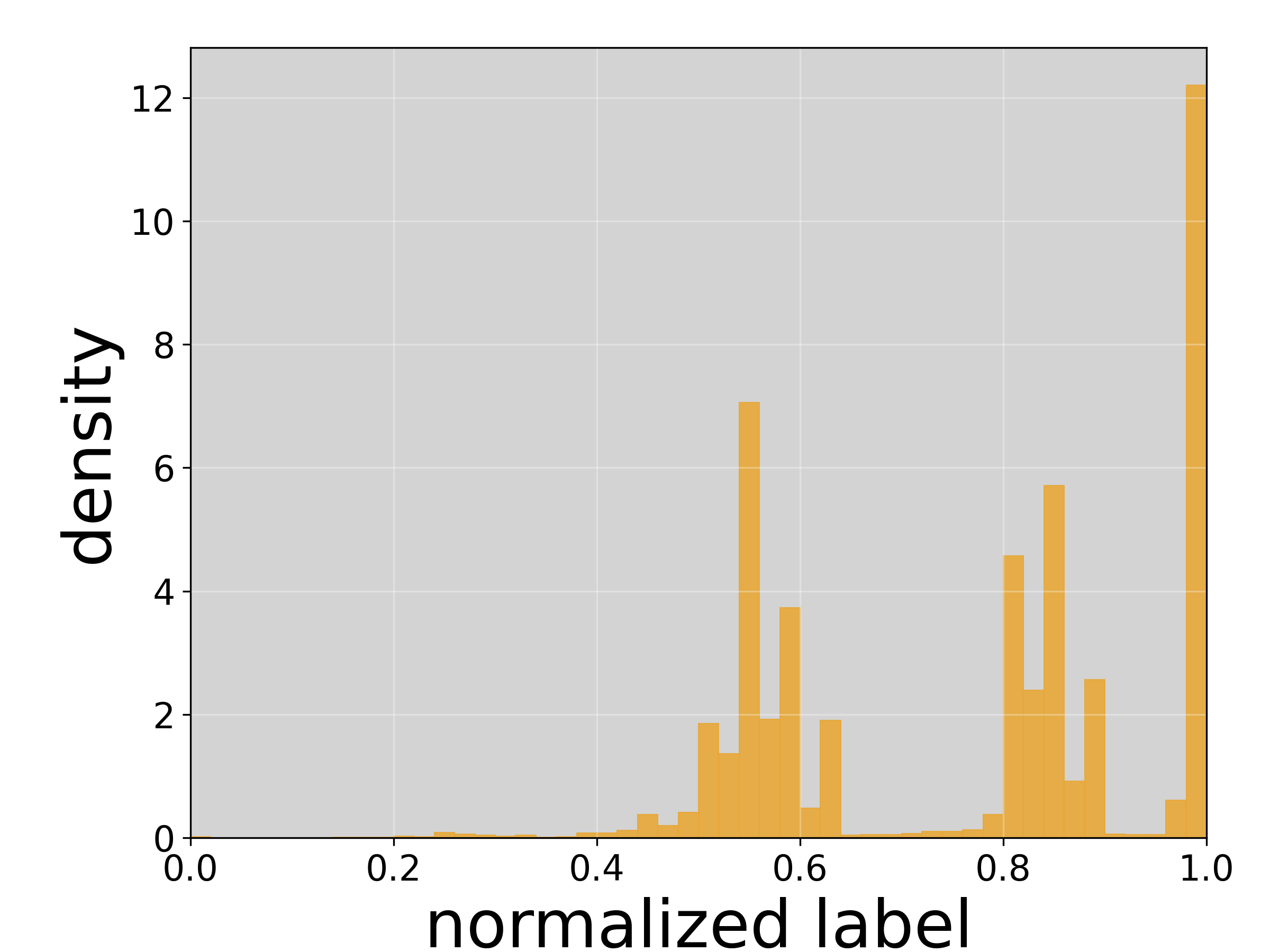}
\subcaption{Digtime}
  \end{minipage}
  \begin{minipage}[t]{0.16\linewidth}  
    \centering
    \includegraphics[width=\linewidth]{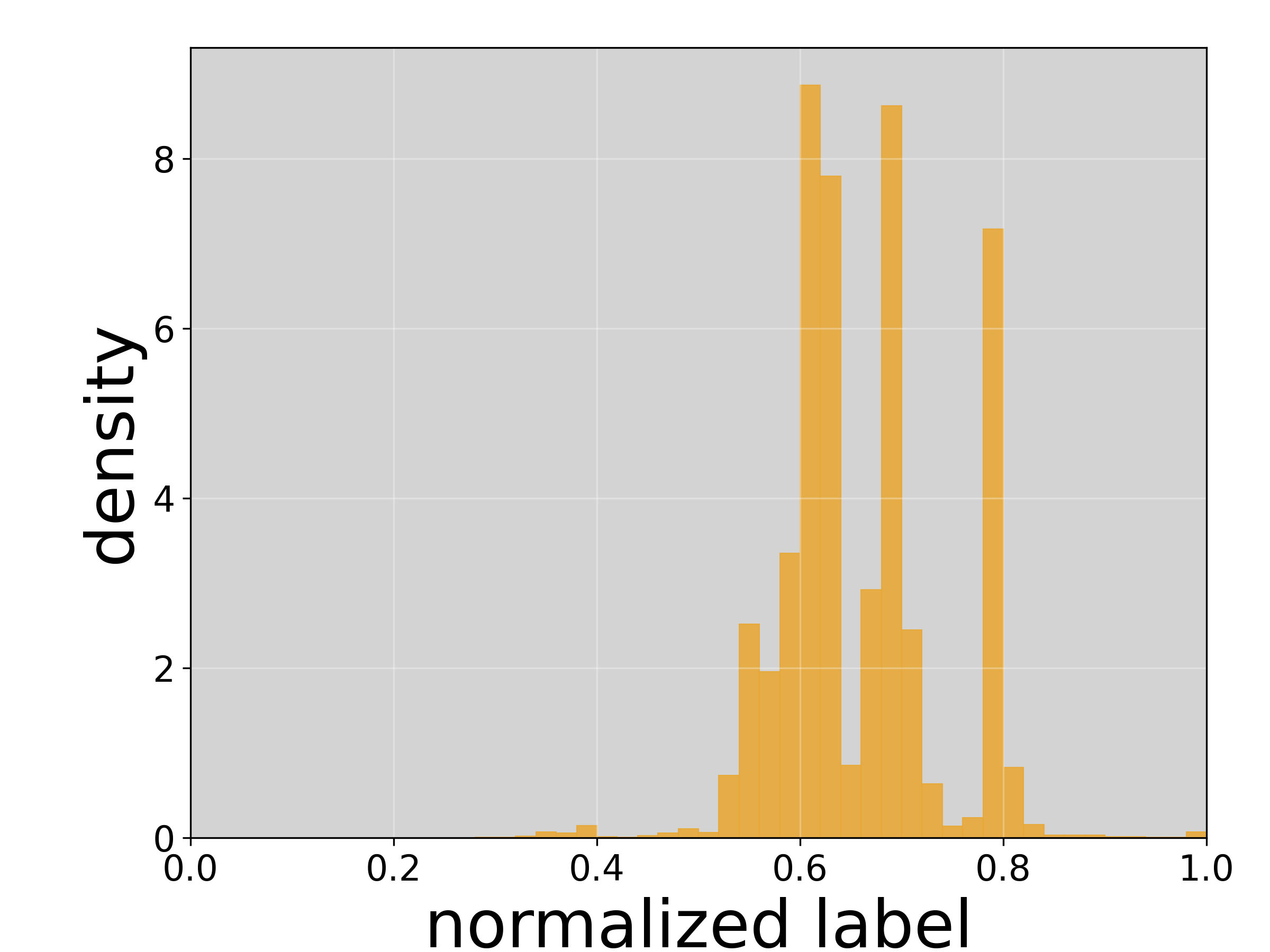}
\subcaption{Sandwich}
  \end{minipage}
  \begin{minipage}[t]{0.16\linewidth}  
    \centering
    \includegraphics[width=\linewidth]{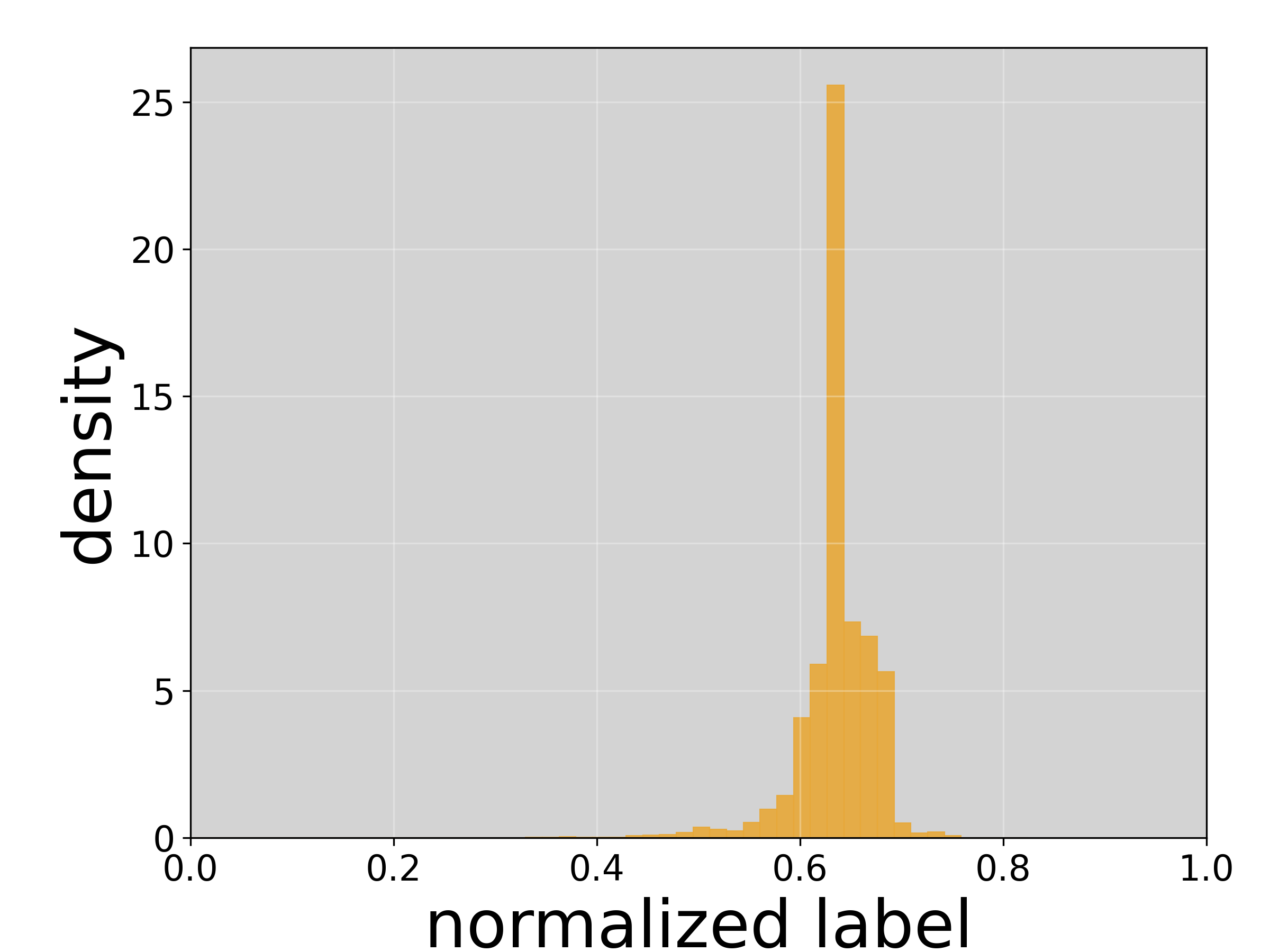}
    \subcaption{SSRAM}
  \end{minipage}
  \begin{minipage}[t]{0.16\linewidth}  
    \centering
    \includegraphics[width=\linewidth]{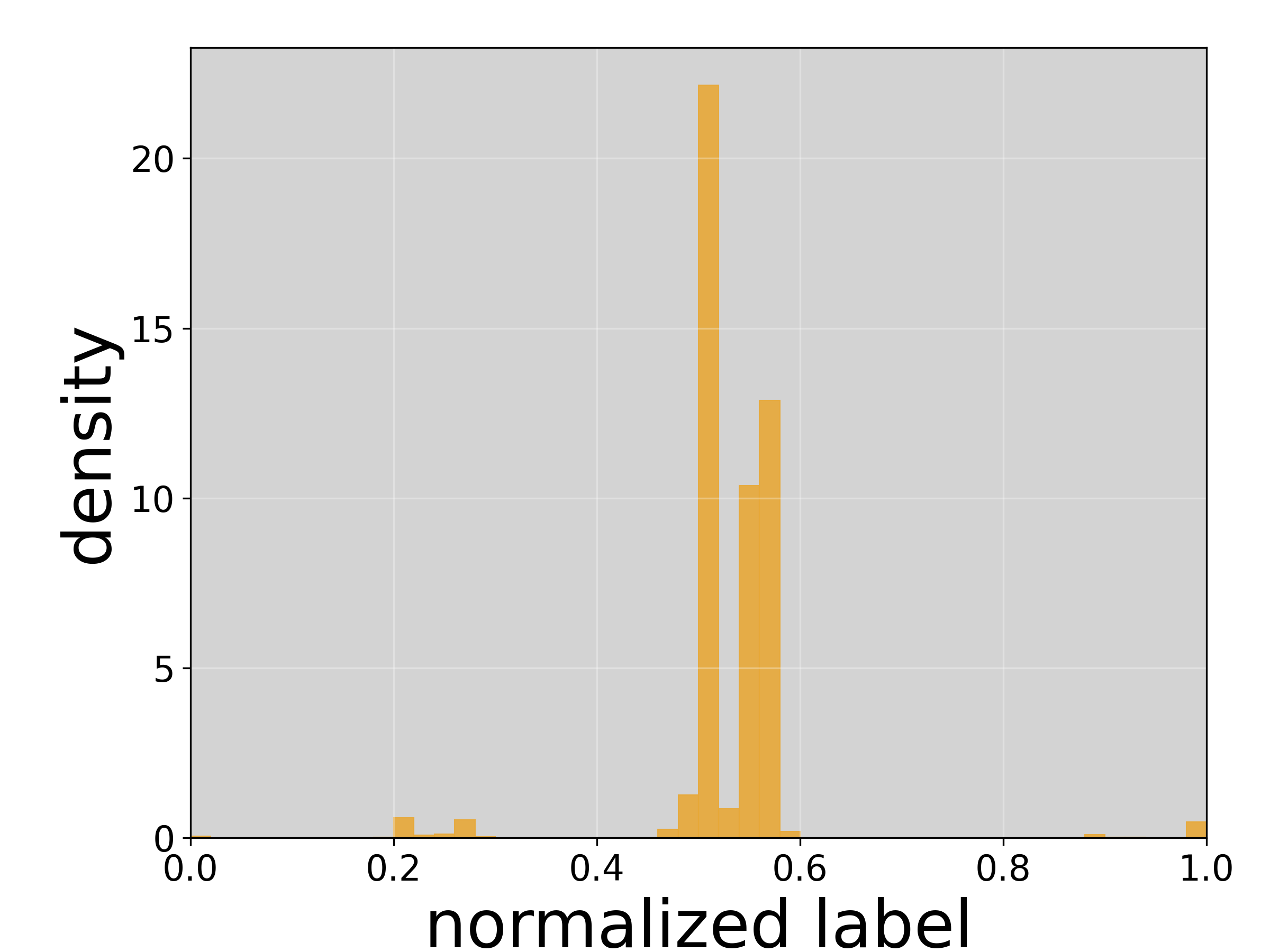}
    \subcaption{Timing\_Ctrl}
  \end{minipage}
  \begin{minipage}[t]{0.16\linewidth}  
    \centering
    \includegraphics[width=\linewidth]{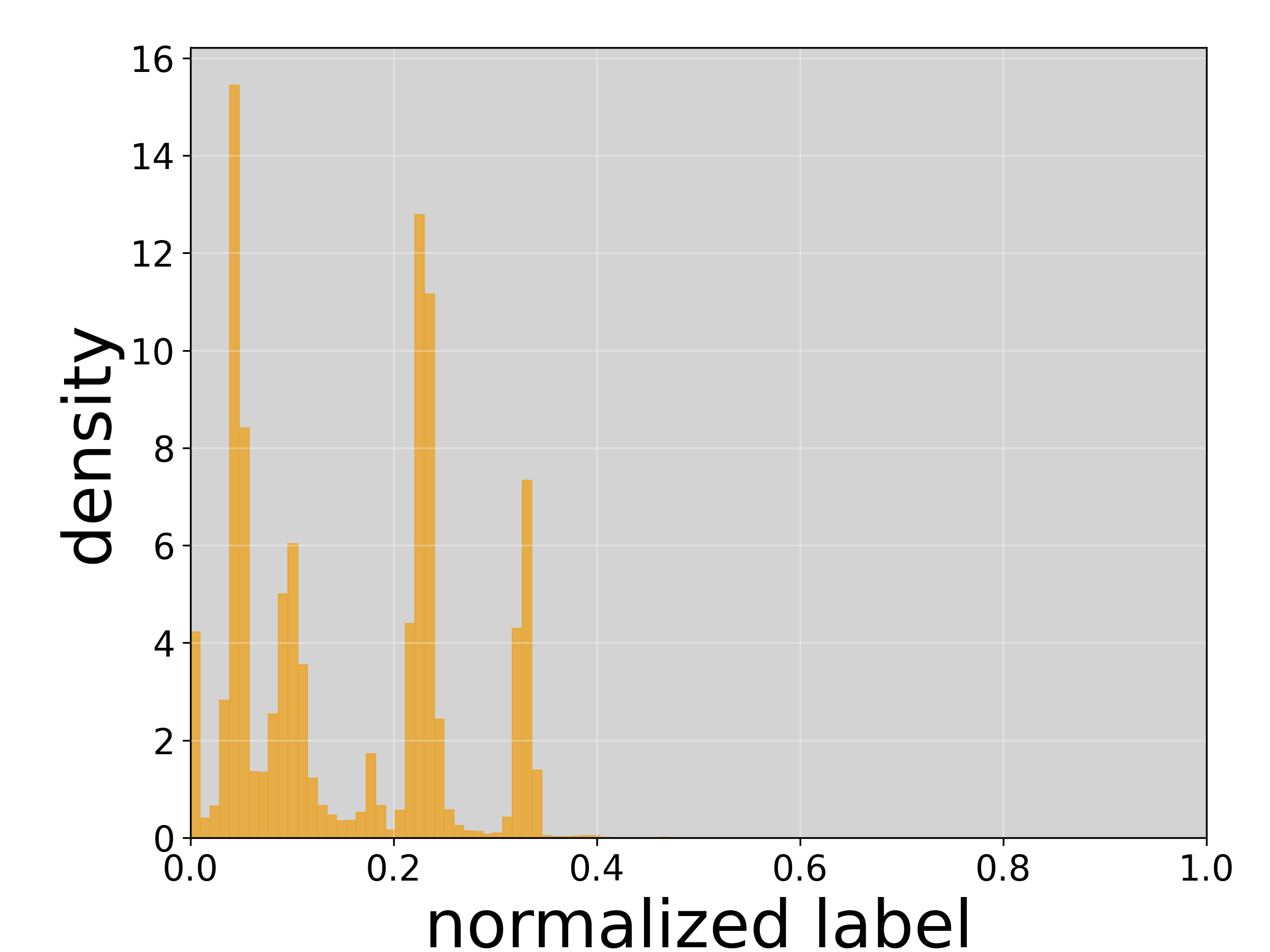}
    \subcaption{Ultra8t}
  \end{minipage}
    \caption{Overview of 6 SRAM Circuit Effective Resistance Label Distribution}
  \label{fig:sram_resistance_each_distribution}
  \hfill
\end{figure*}

%% file: tables/analog_ground_each_distribution.tex
\begin{figure}[H]  
  \centering       
  \setlength{\tabcolsep}{2pt}  
  
  \begin{minipage}[t]{0.19\textwidth}  
    \centering
    \includegraphics[width=\linewidth]{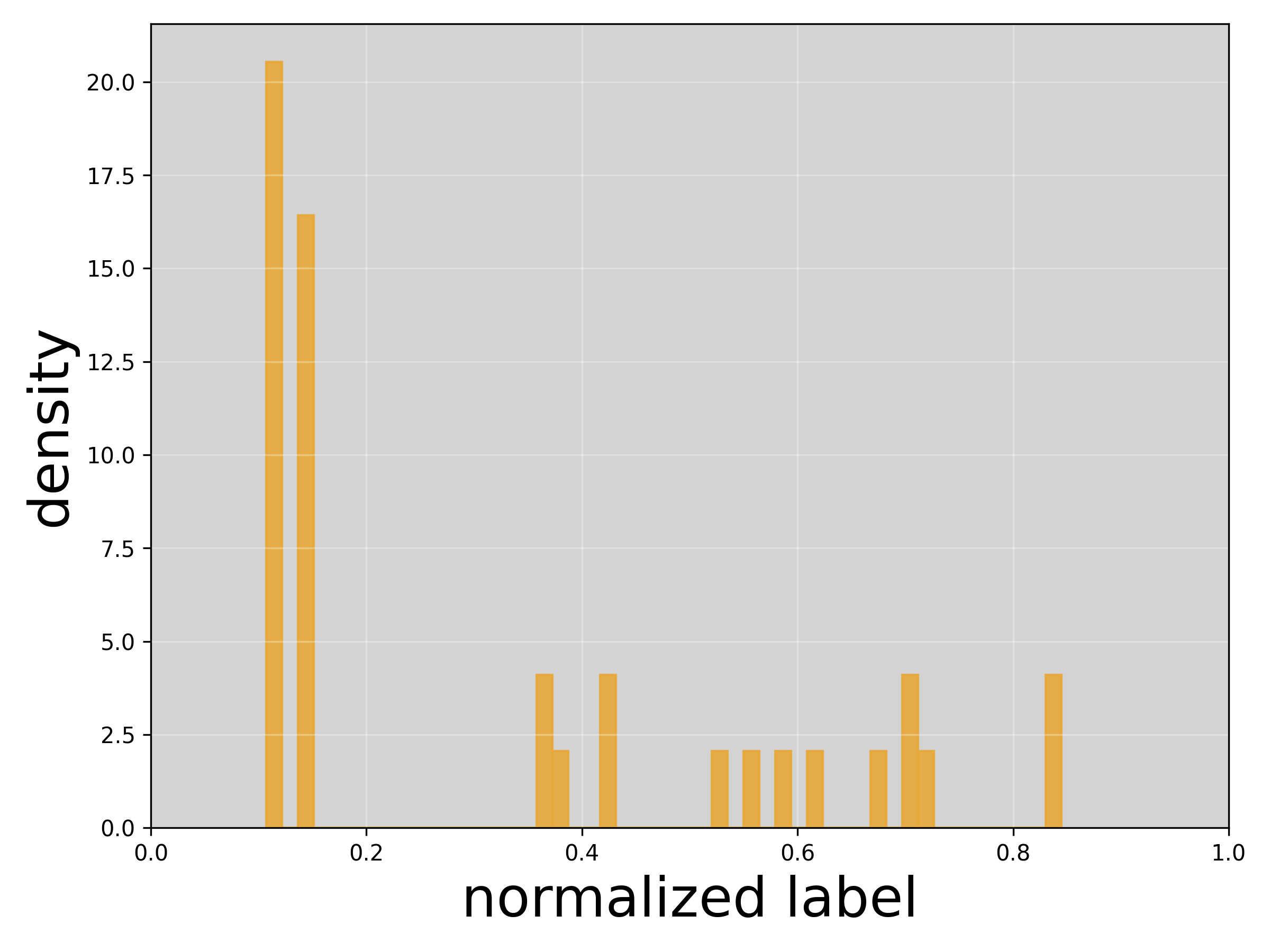}  
    \subcaption{Case 1}  
  \end{minipage}
  \hfill  
  \begin{minipage}[t]{0.19\textwidth}
    \centering
    \includegraphics[width=\linewidth]{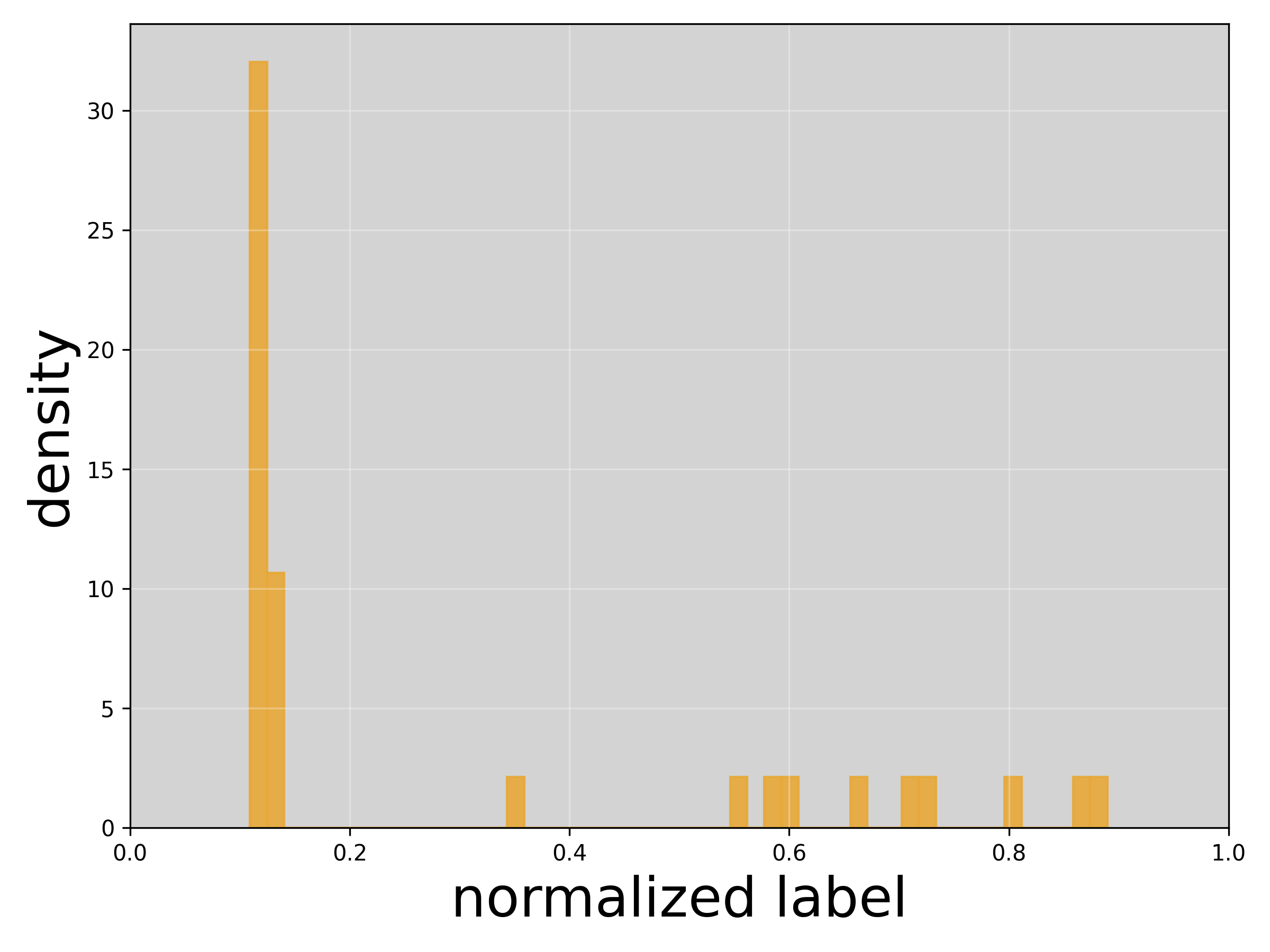}
    \subcaption{Case 2}
  \end{minipage}
  \hfill
  \begin{minipage}[t]{0.19\textwidth}
    \centering
    \includegraphics[width=\linewidth]{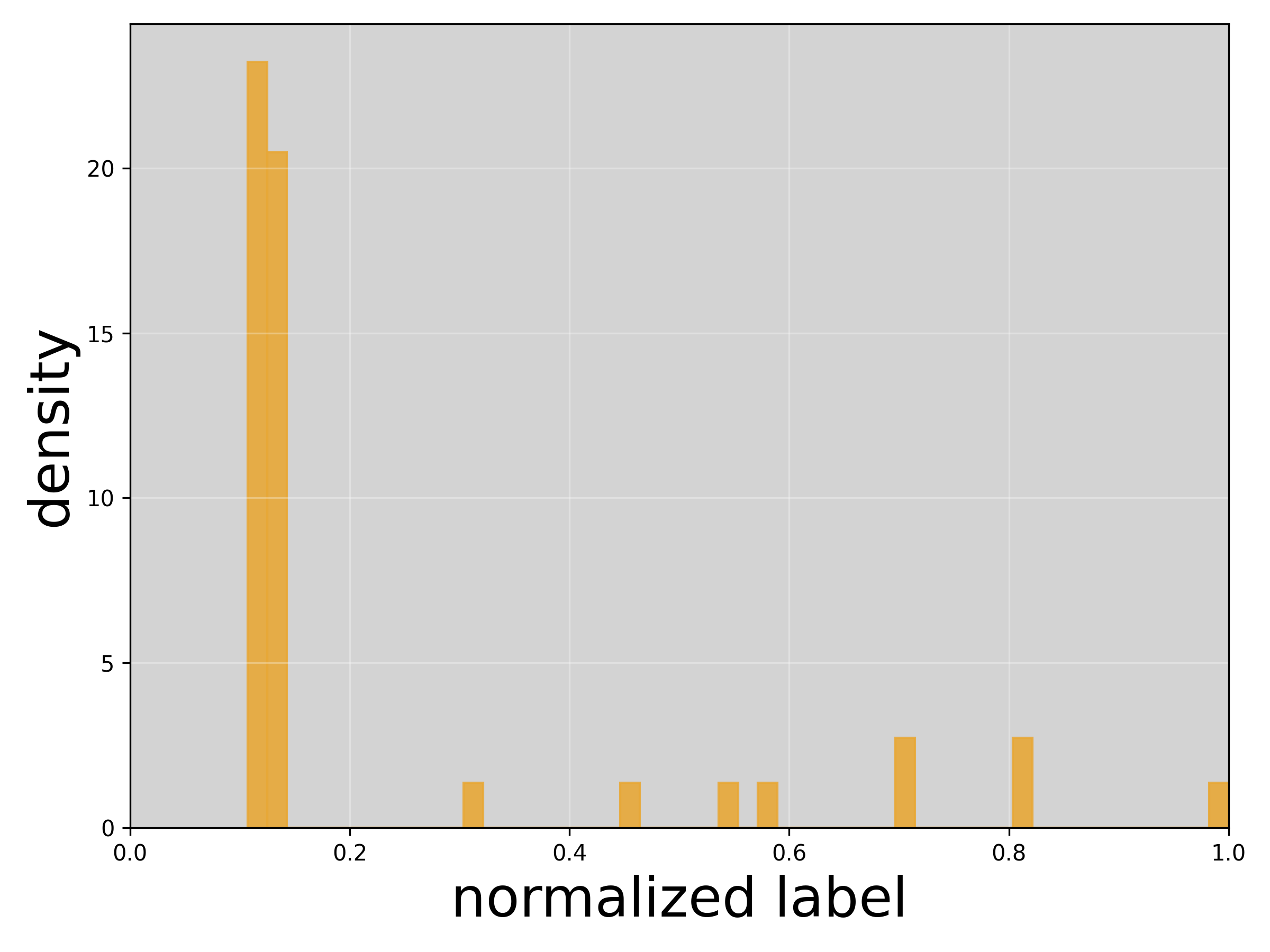}
    \subcaption{Case 3}
  \end{minipage}
  \hfill
  \begin{minipage}[t]{0.19\textwidth}
    \centering
    \includegraphics[width=\linewidth]{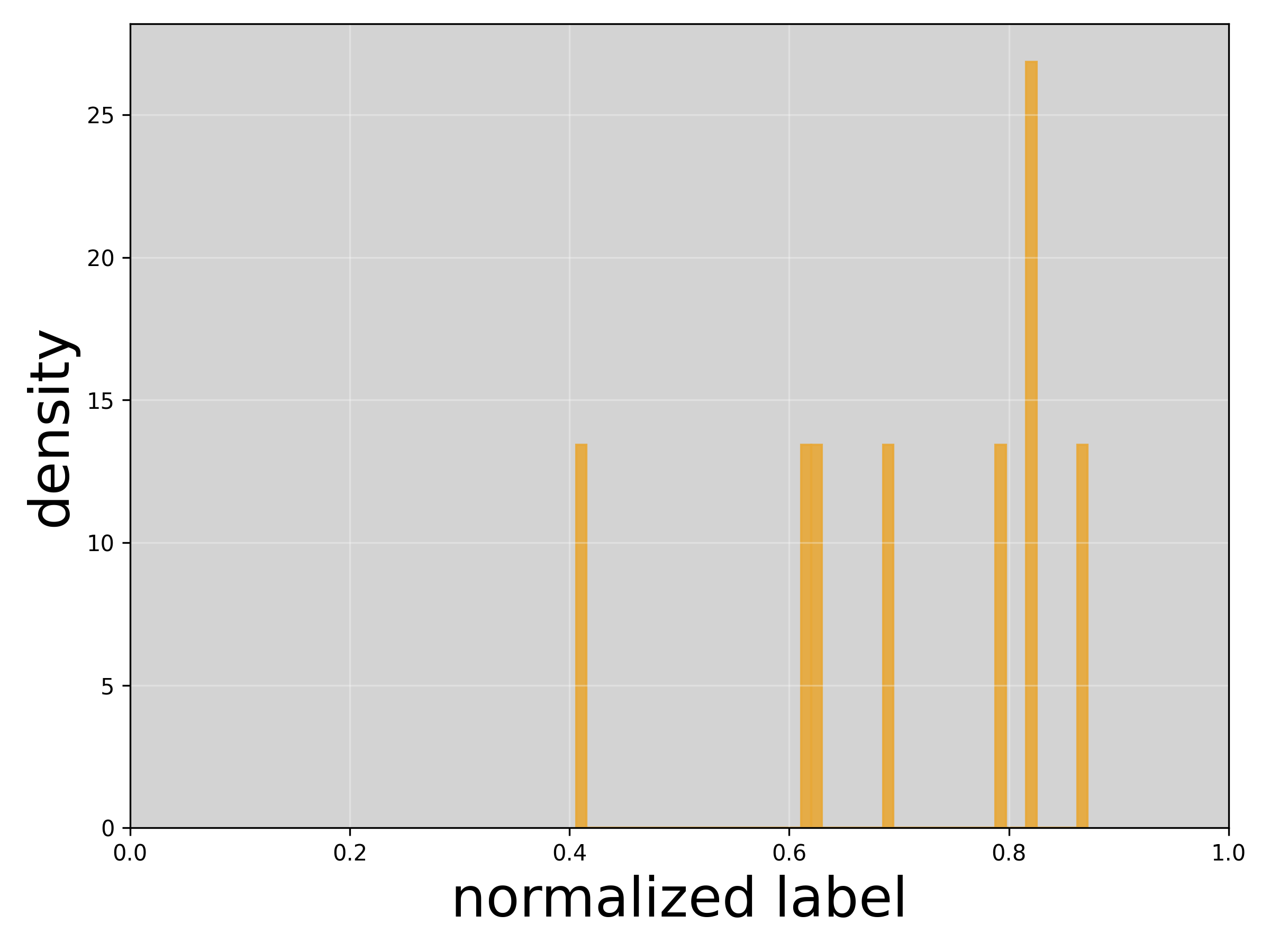}
    \subcaption{Case 4}
  \end{minipage}
  \hfill
  \begin{minipage}[t]{0.19\textwidth}
    \centering
    \includegraphics[width=\linewidth]{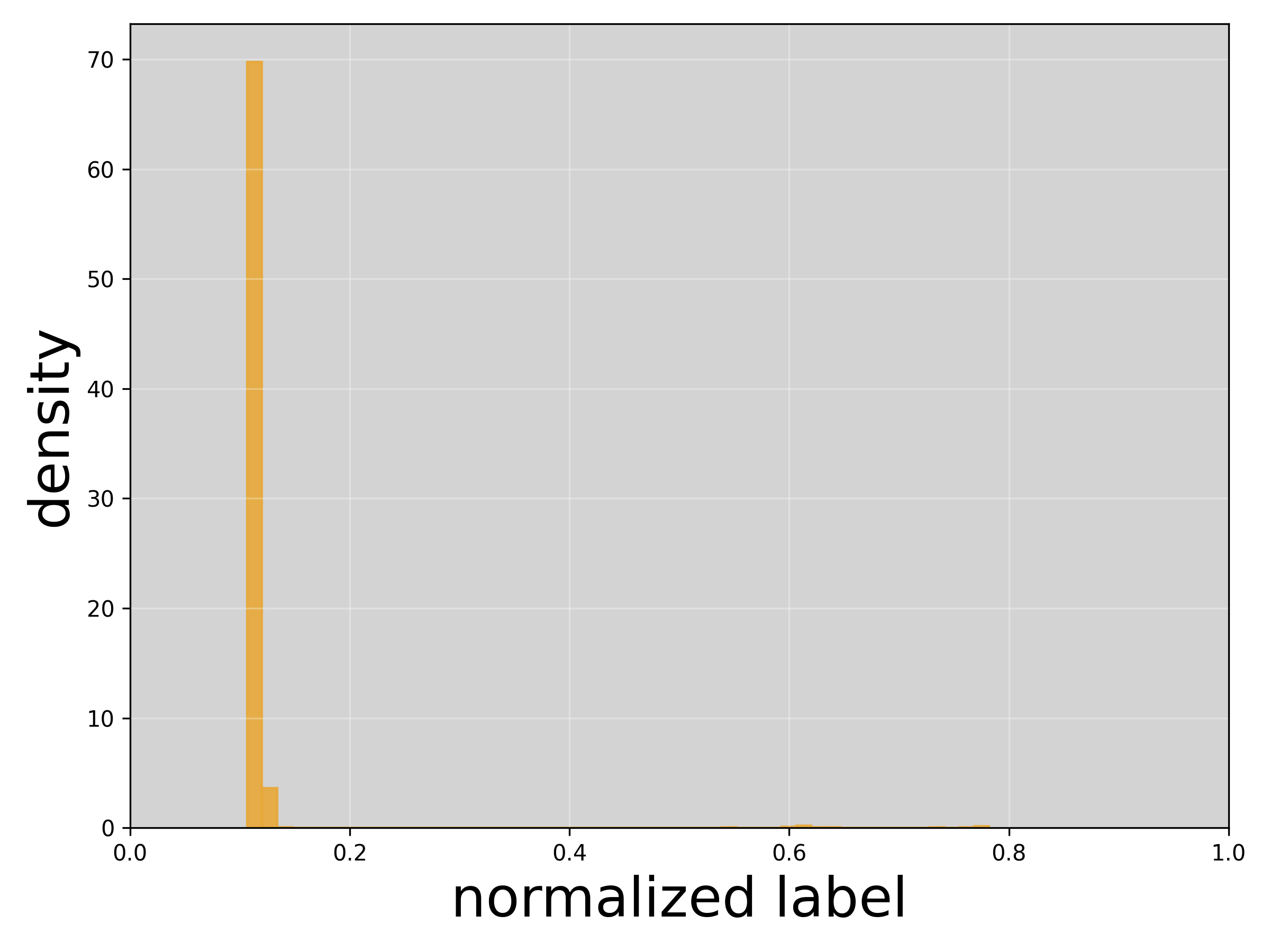}
    \subcaption{Case 5}
  \end{minipage}
  
  \vspace{5pt}  
  
  \begin{minipage}[t]{0.19\textwidth}
    \centering
    \includegraphics[width=\linewidth]{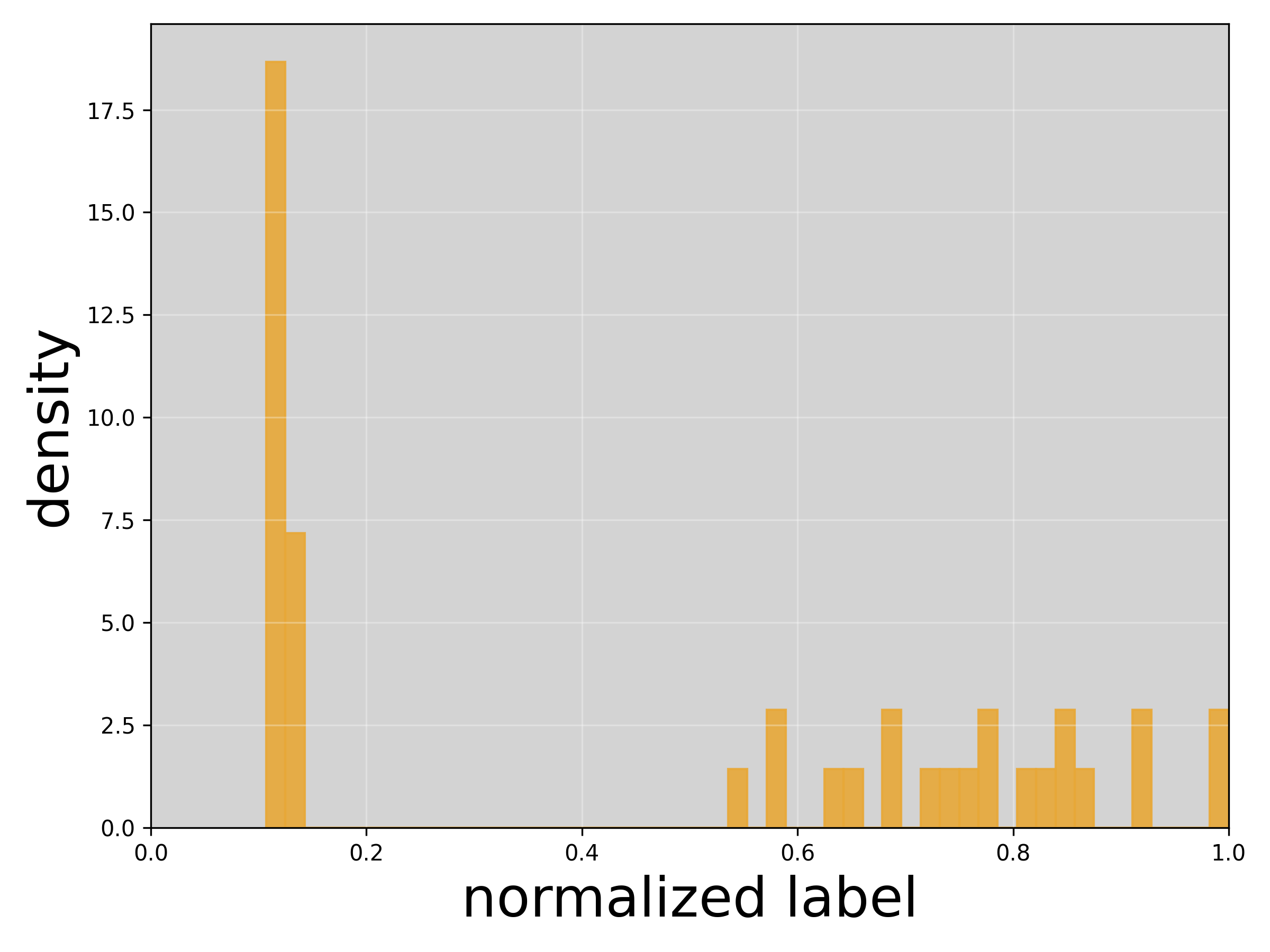}
    \subcaption{Case 6}
  \end{minipage}
  \hfill
  \begin{minipage}[t]{0.19\textwidth}
    \centering
    \includegraphics[width=\linewidth]{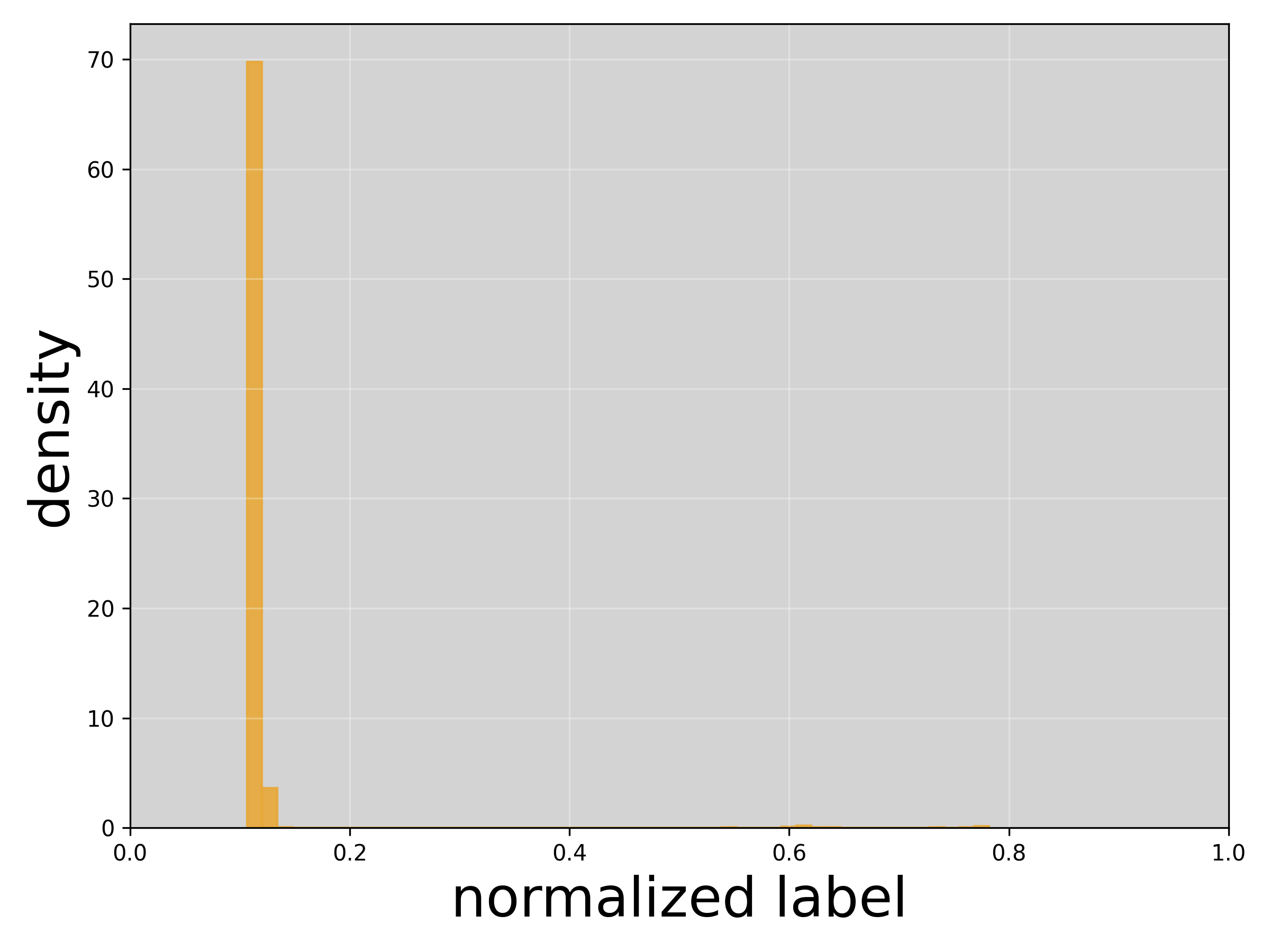}
    \subcaption{Case 7}
  \end{minipage}
  \hfill
  \begin{minipage}[t]{0.19\textwidth}
    \centering
    \includegraphics[width=\linewidth]{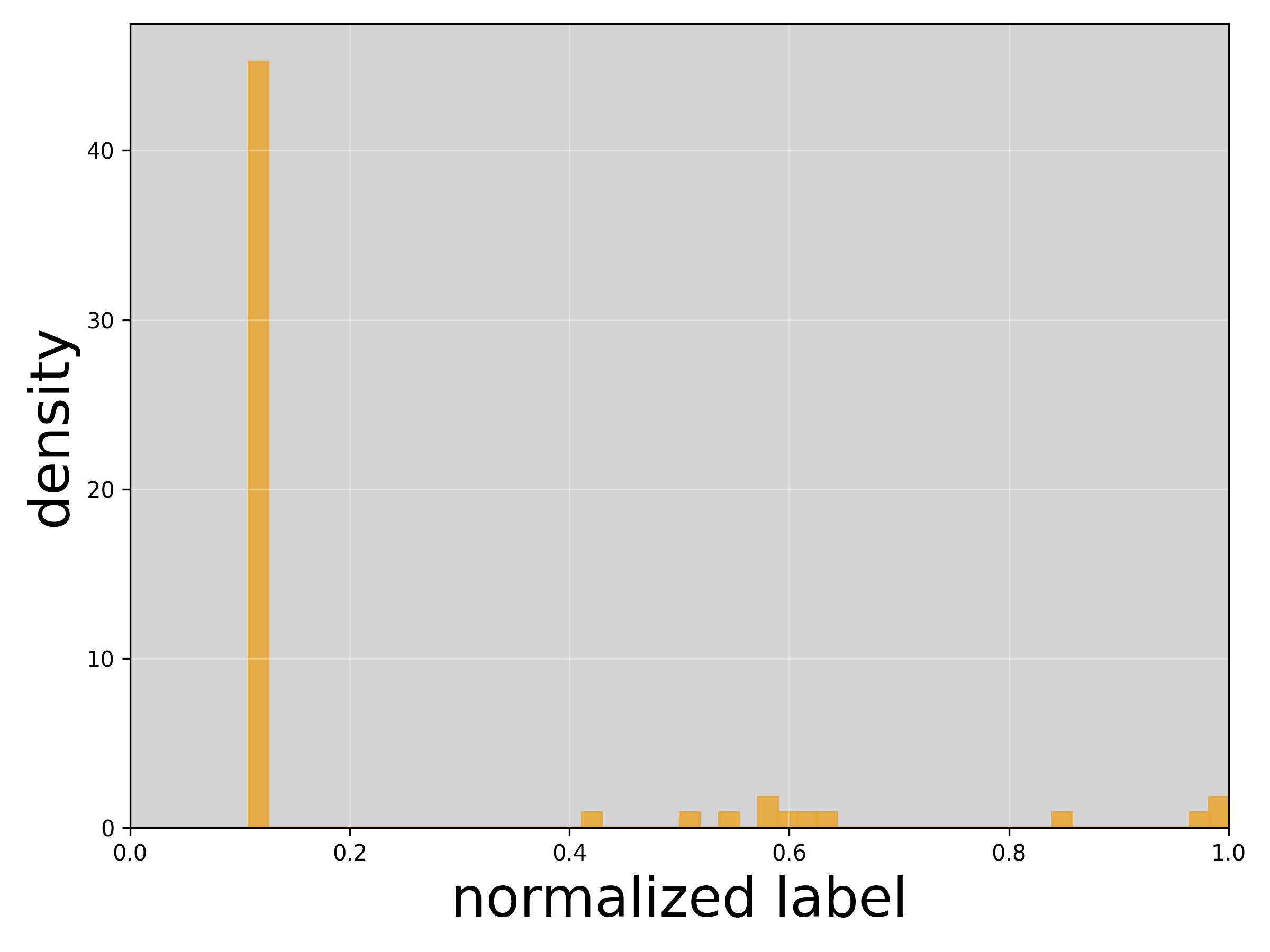}
    \subcaption{Case 8}
  \end{minipage}
  \hfill
  \begin{minipage}[t]{0.19\textwidth}
    \centering
    \includegraphics[width=\linewidth]{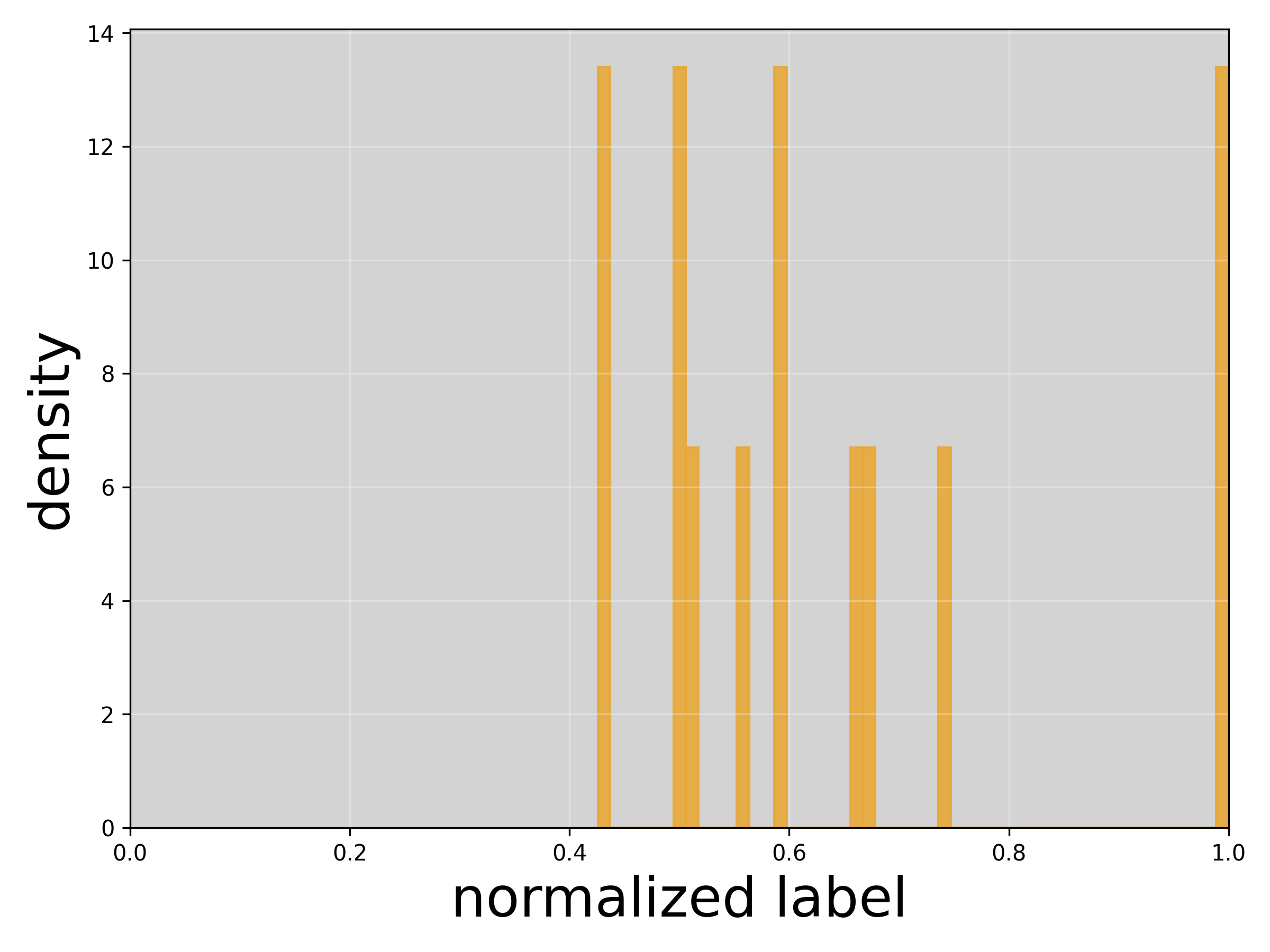}
    \subcaption{Case 9}
  \end{minipage}
  \hfill
  \begin{minipage}[t]{0.19\textwidth}
    \centering
    \includegraphics[width=\linewidth]{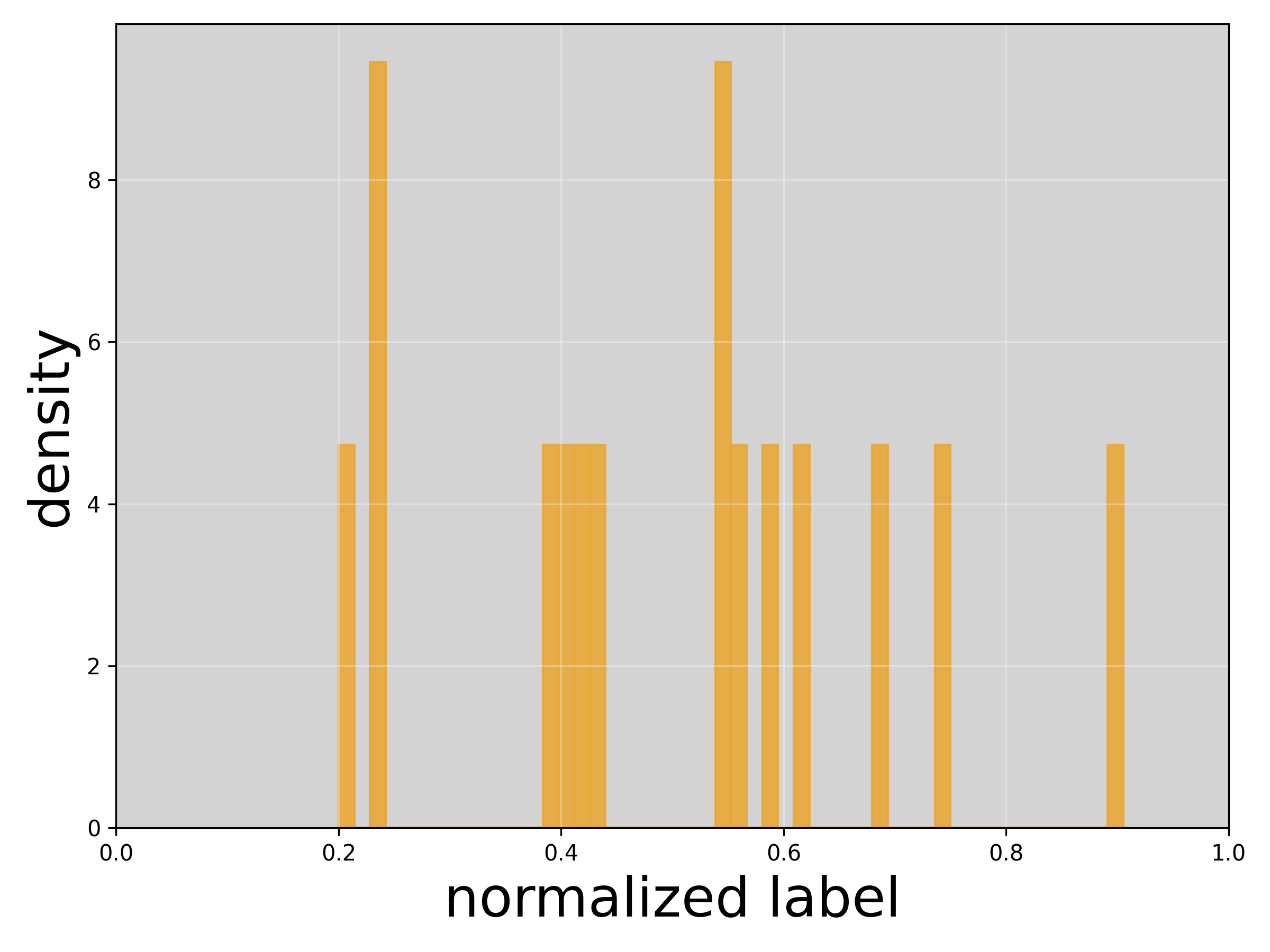}
    \subcaption{Case 10}
  \end{minipage}
  
  \vspace{5pt}
  
  \begin{minipage}[t]{0.19\textwidth}
    \centering
    \includegraphics[width=\linewidth]{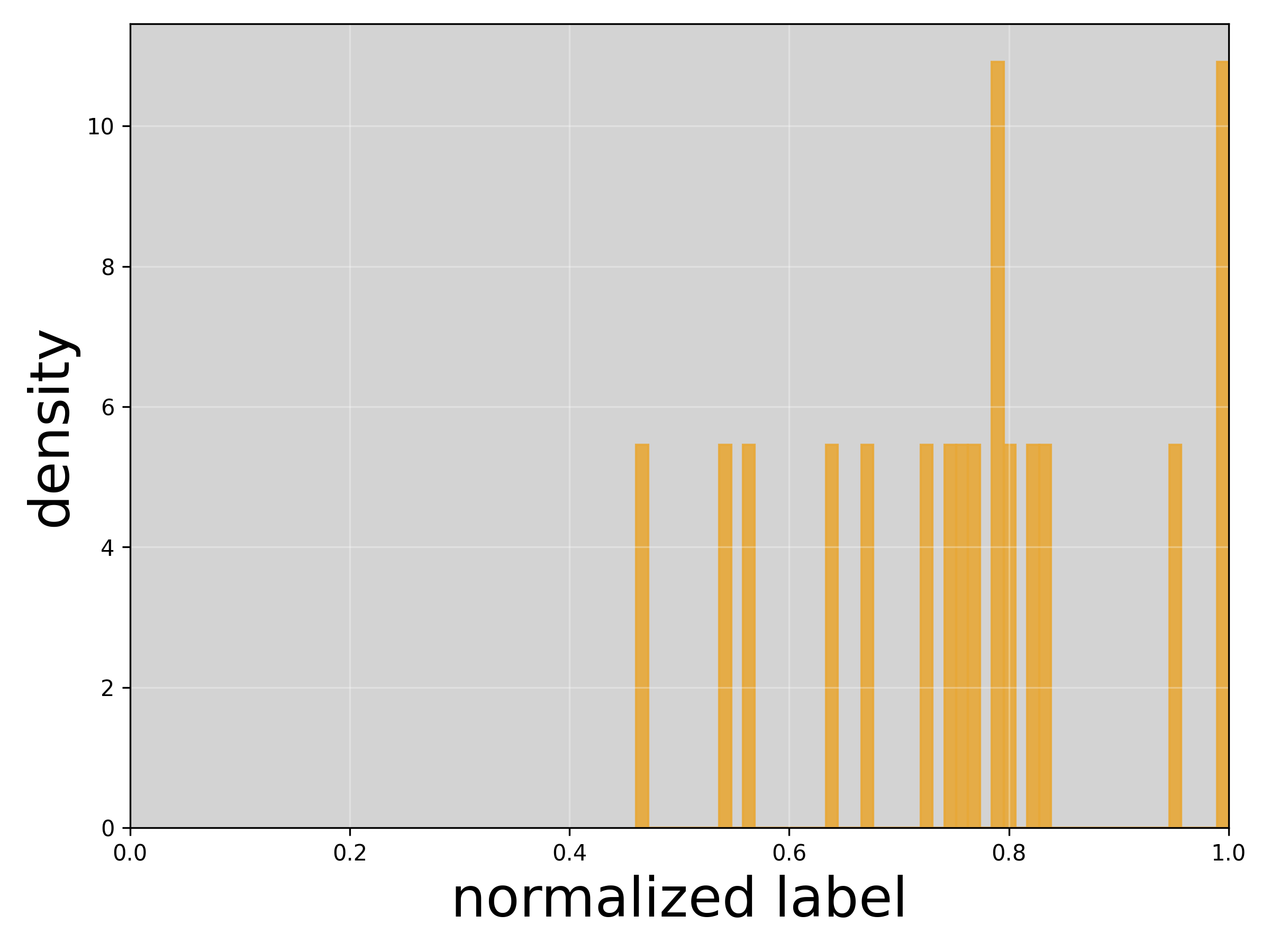}
    \subcaption{Case 11}
  \end{minipage}
  \hfill
  \begin{minipage}[t]{0.19\textwidth}
    \centering
    \includegraphics[width=\linewidth]{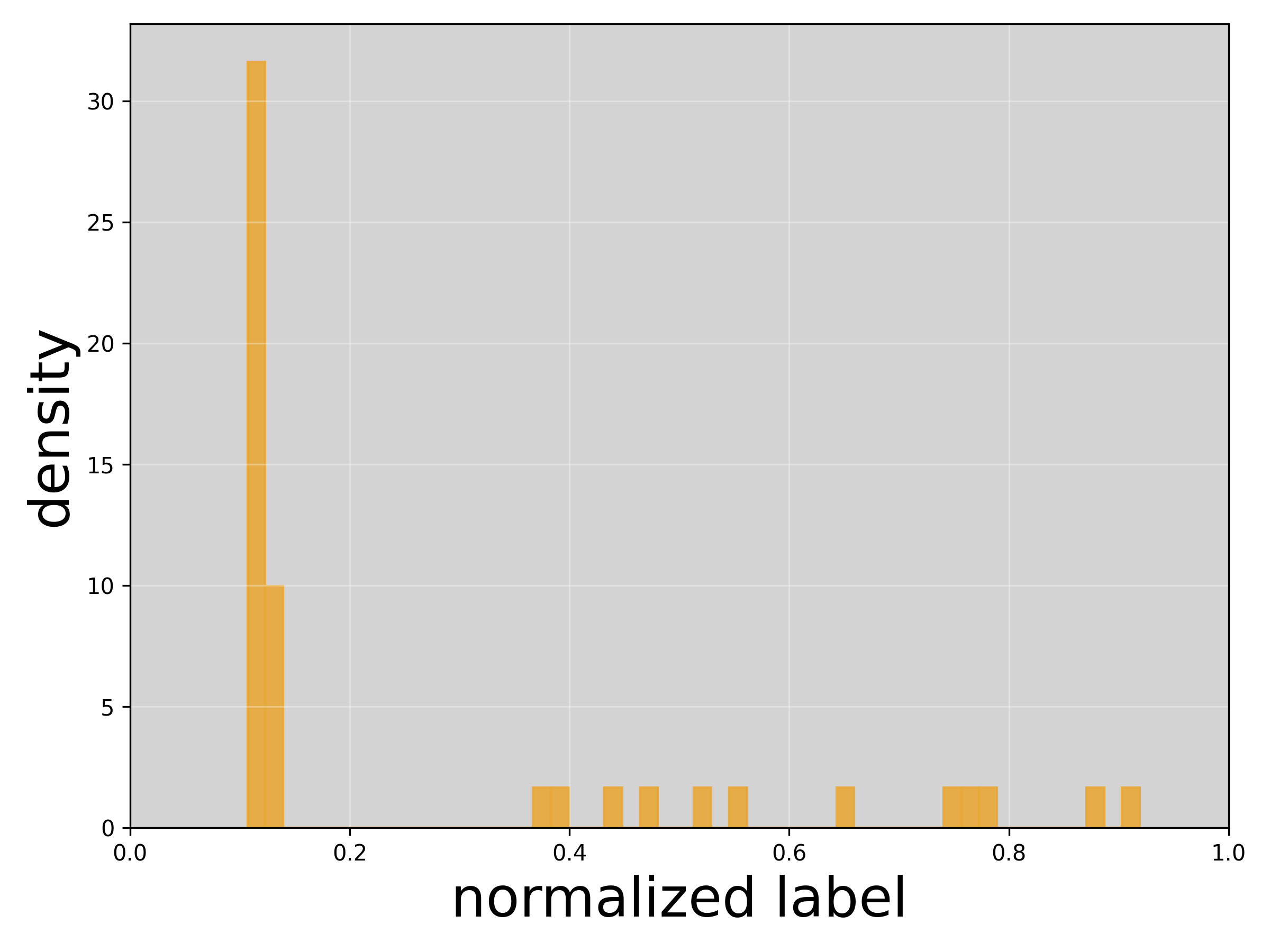}
    \subcaption{Case 12}
  \end{minipage}
  \hfill
  \begin{minipage}[t]{0.19\textwidth}
    \centering
    \includegraphics[width=\linewidth]{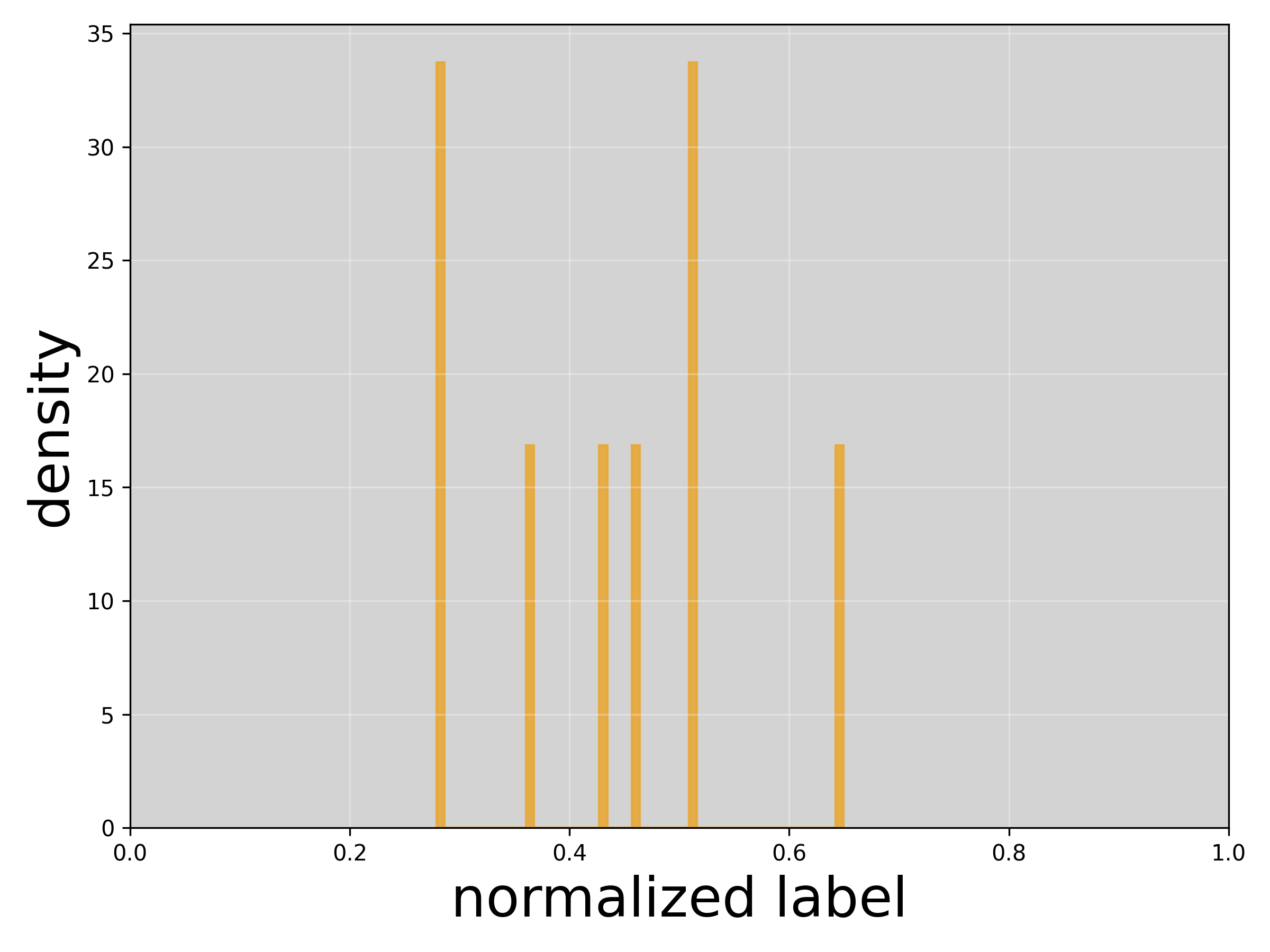}
    \subcaption{Case 13}
  \end{minipage}
  \hfill
  \begin{minipage}[t]{0.19\textwidth}
    \centering
    \includegraphics[width=\linewidth]{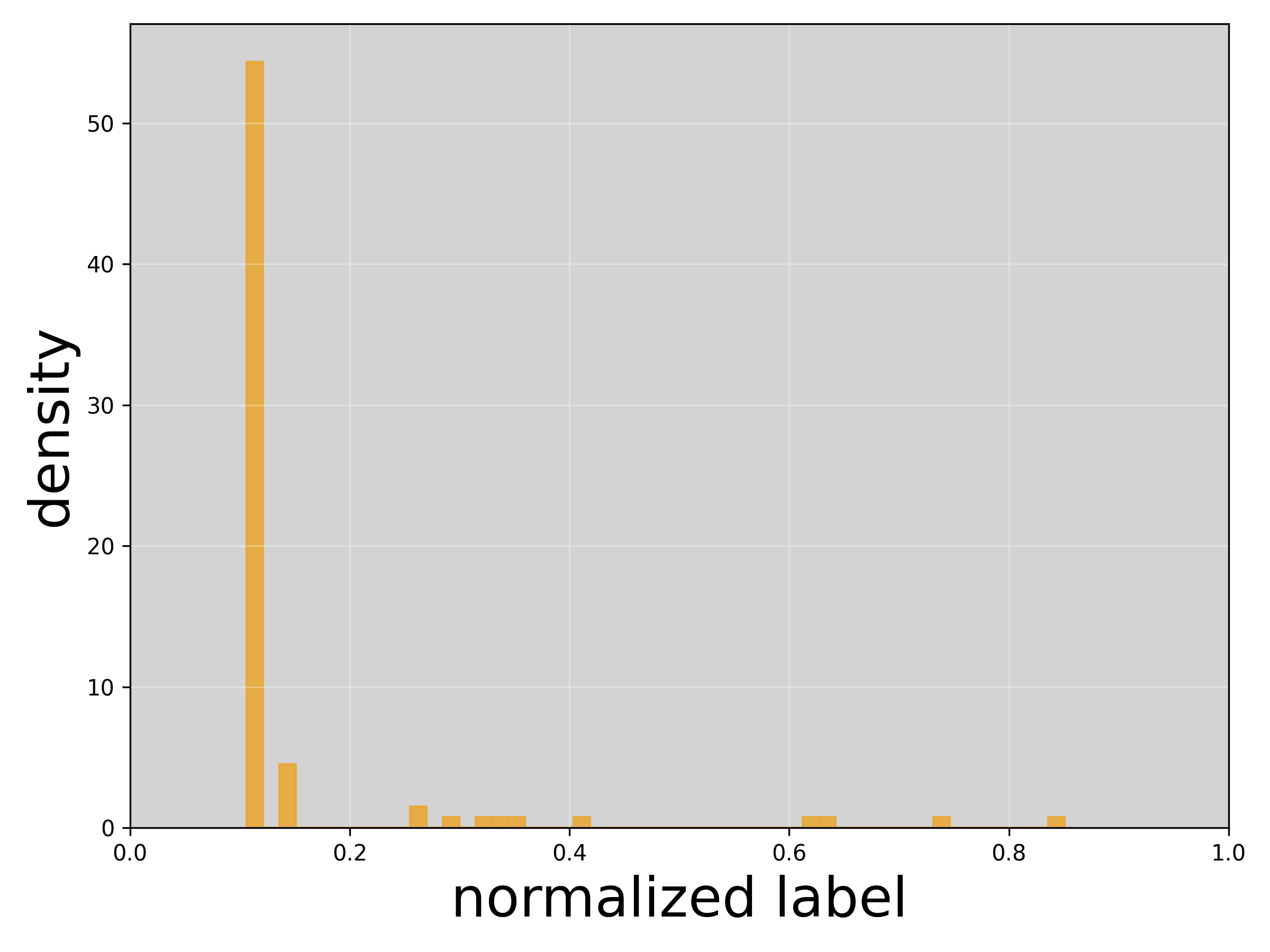}
    \subcaption{Case 14}
  \end{minipage}
  \hfill
  \begin{minipage}[t]{0.19\textwidth}
    \centering
    \includegraphics[width=\linewidth]{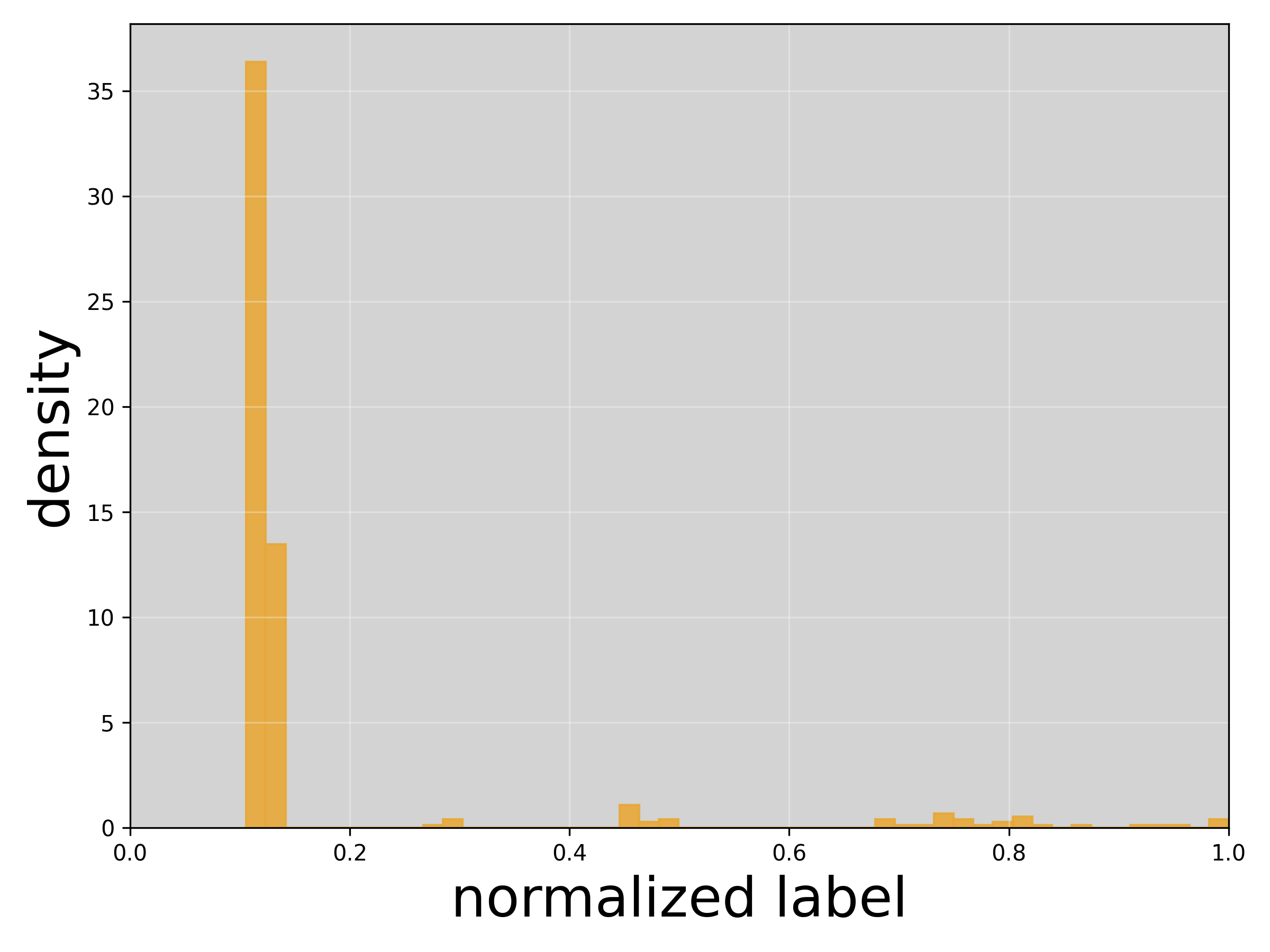}
    \subcaption{Case 15}
  \end{minipage}
  
  \vspace{5pt}
  
  \begin{minipage}[t]{0.19\textwidth}
    \centering
    \includegraphics[width=\linewidth]{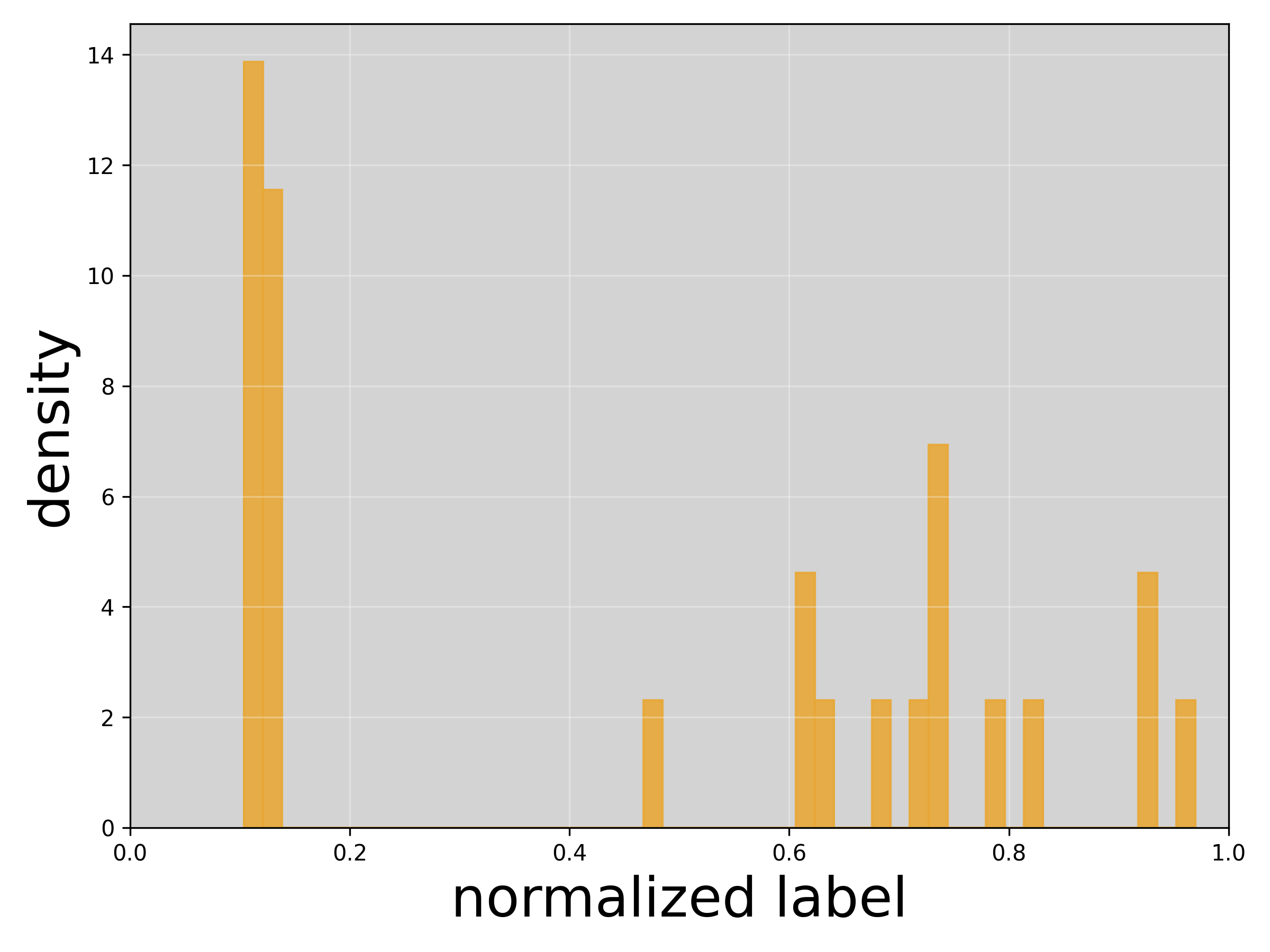}
    \subcaption{Case 16}
  \end{minipage}
  \hfill
  \begin{minipage}[t]{0.19\textwidth}
    \centering
    \includegraphics[width=\linewidth]{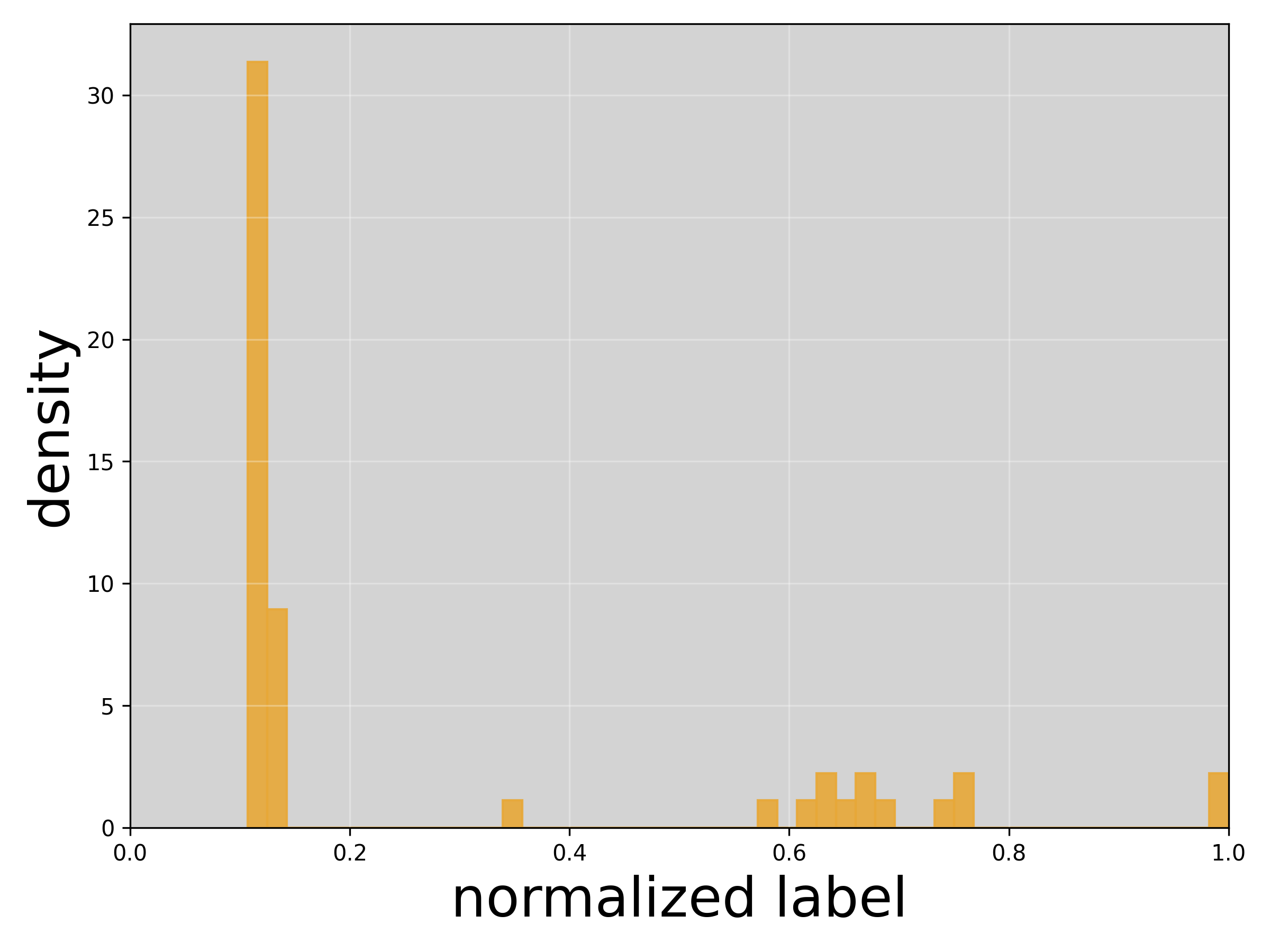}
    \subcaption{Case 17}
  \end{minipage}
  \hfill
  \begin{minipage}[t]{0.19\textwidth}
    \centering
    \includegraphics[width=\linewidth]{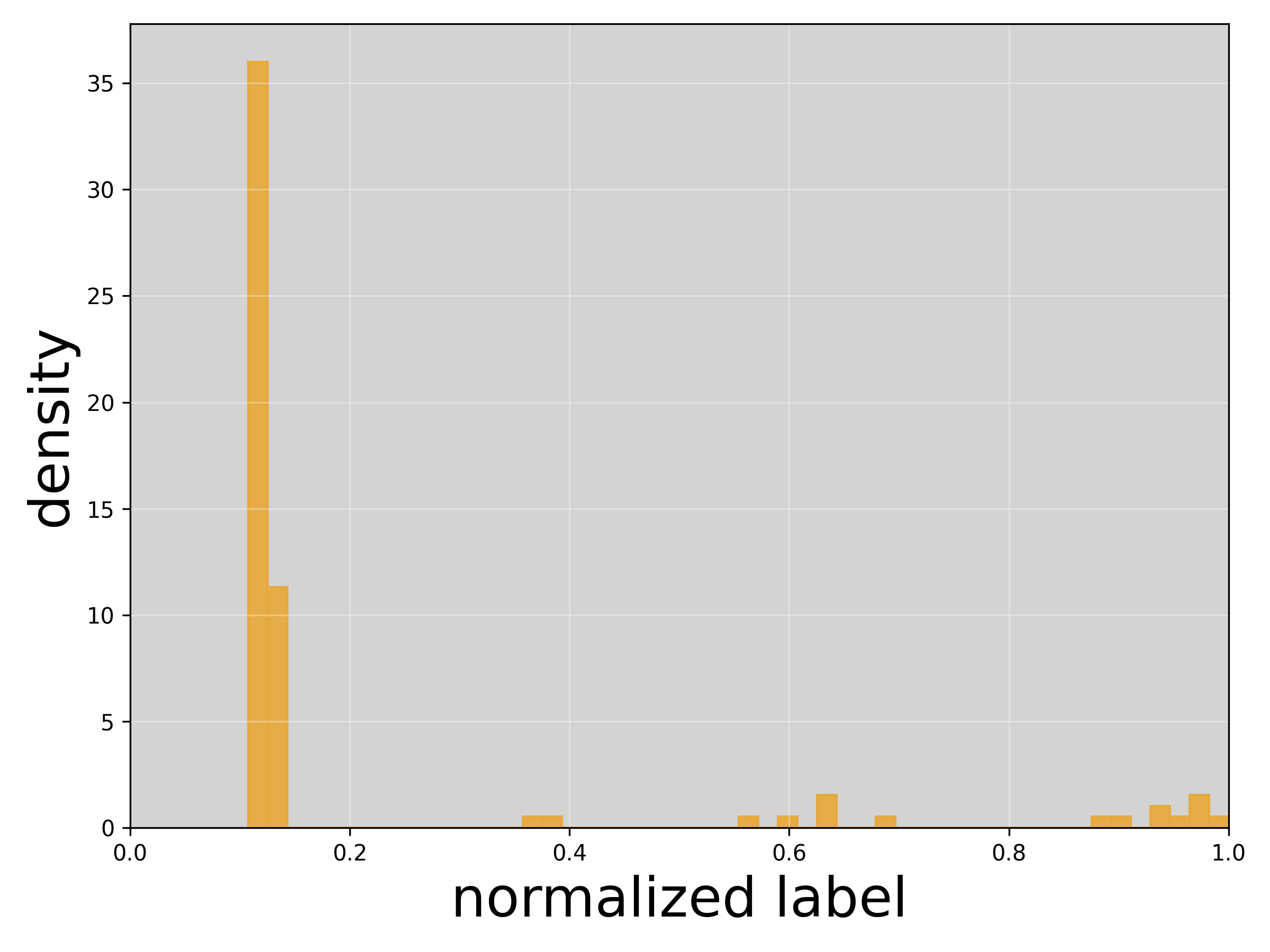}
    \subcaption{Case 18}
  \end{minipage}
  \hfill
  \begin{minipage}[t]{0.19\textwidth}
    \centering
    \includegraphics[width=\linewidth]{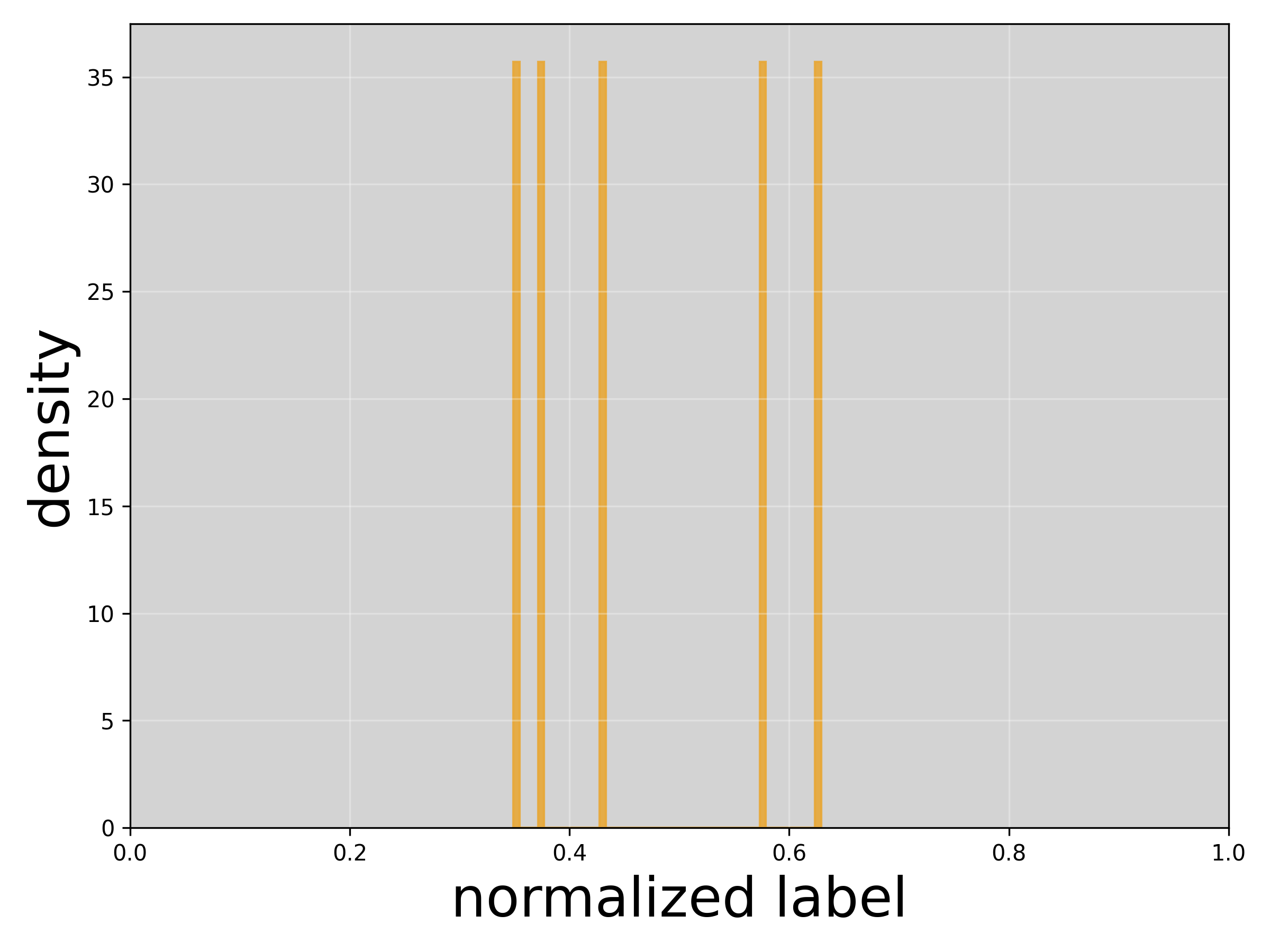}
    \subcaption{Case 19}
  \end{minipage}
  \hfill
  \begin{minipage}[t]{0.19\textwidth}
    \centering
    \includegraphics[width=\linewidth]{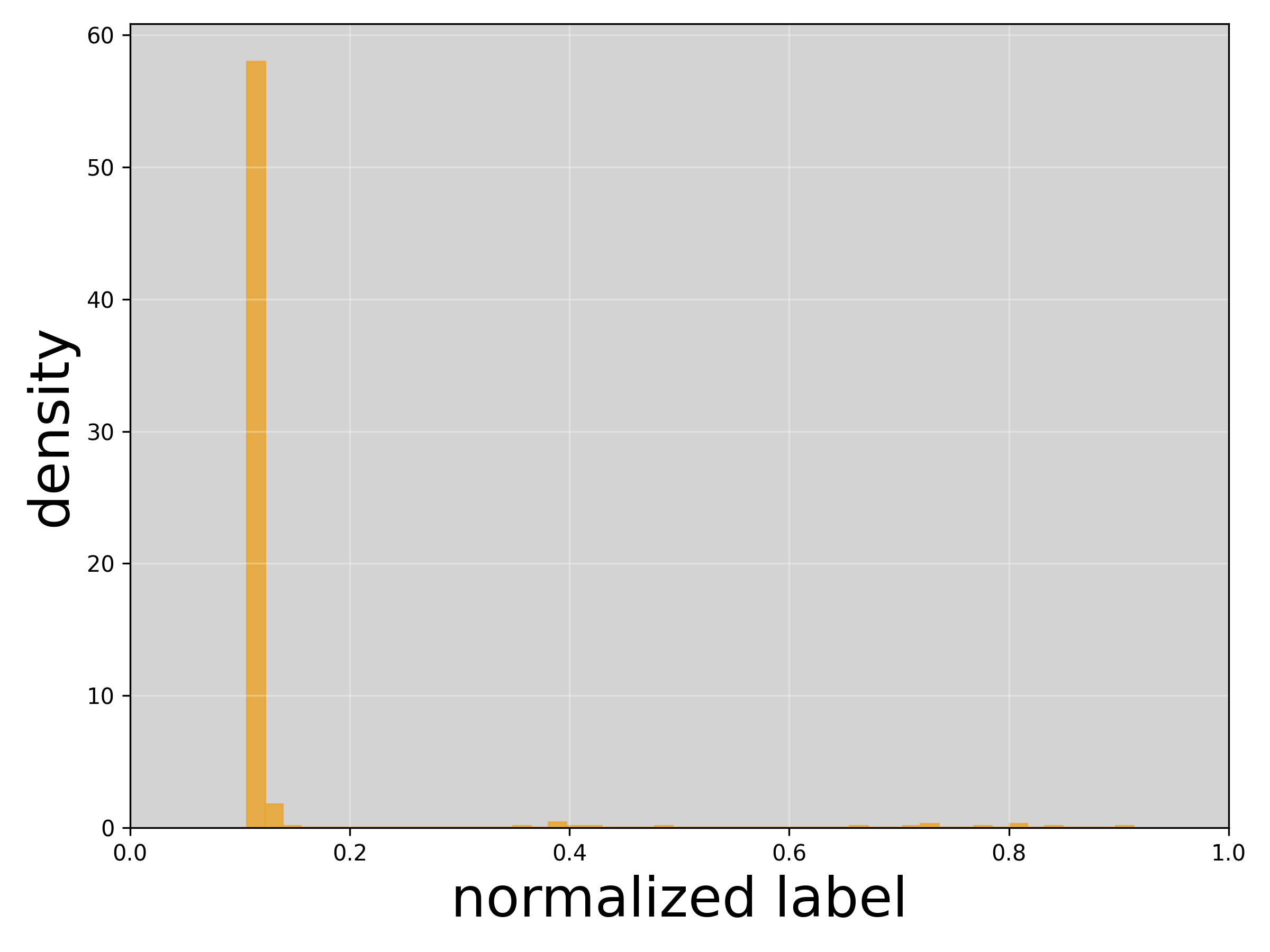}
    \subcaption{Case 20}
  \end{minipage}

  \caption{Overview of 20 Analog Circuit Ground Capacitance Label Distribution}
  \label{fig:analog_ground_each_distribution}
\end{figure}

%% file: tables/sram_ground_each_distribution.tex
\begin{figure*}[htbp]  
  \centering
  \begin{minipage}[t]{0.16\linewidth}  
    \centering
    \includegraphics[width=\linewidth]{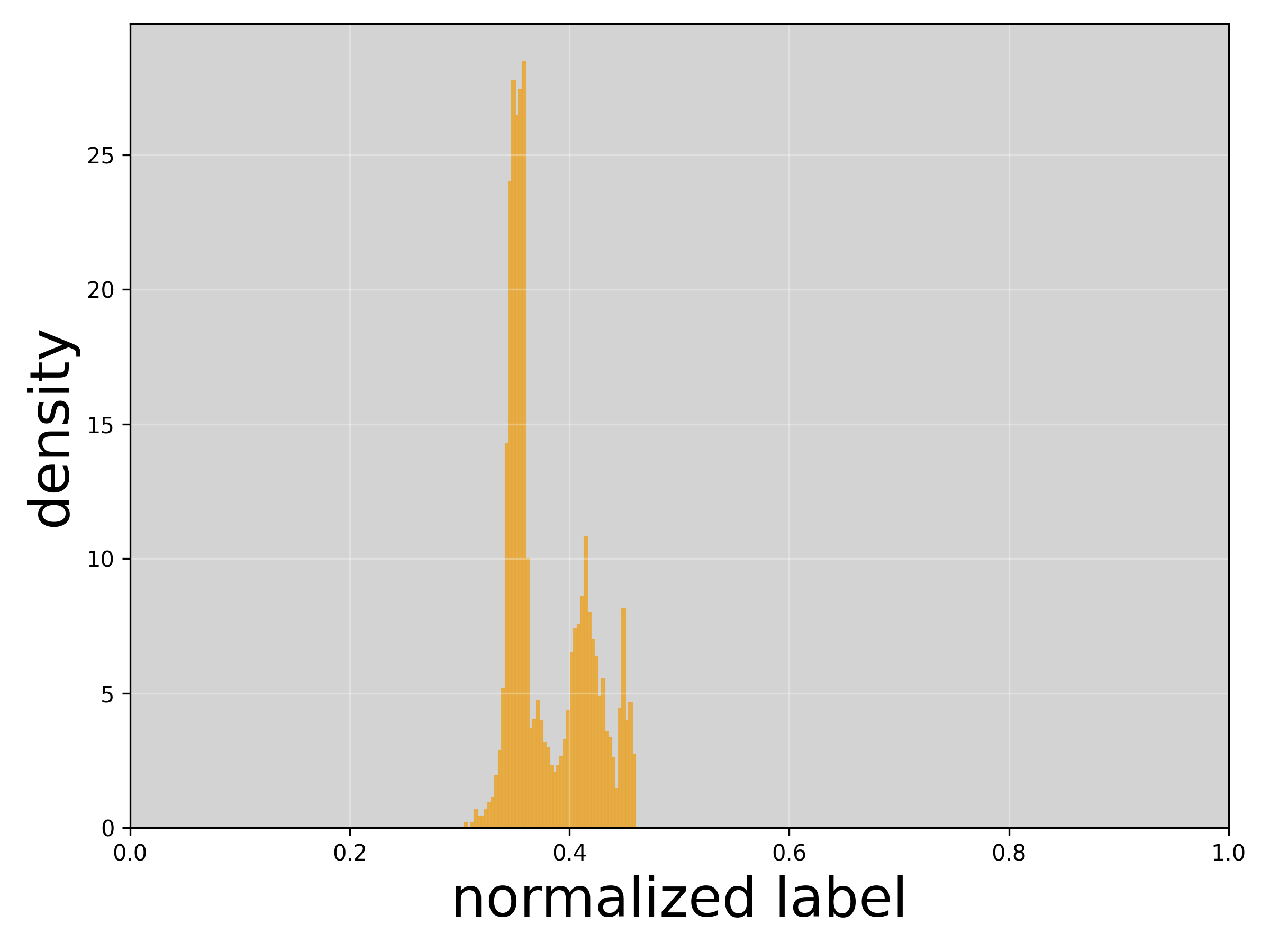}
 \subcaption{Array\_128\_32\_8t}
  \end{minipage}
  \begin{minipage}[t]{0.16\linewidth}  
    \centering
    \includegraphics[width=\linewidth]{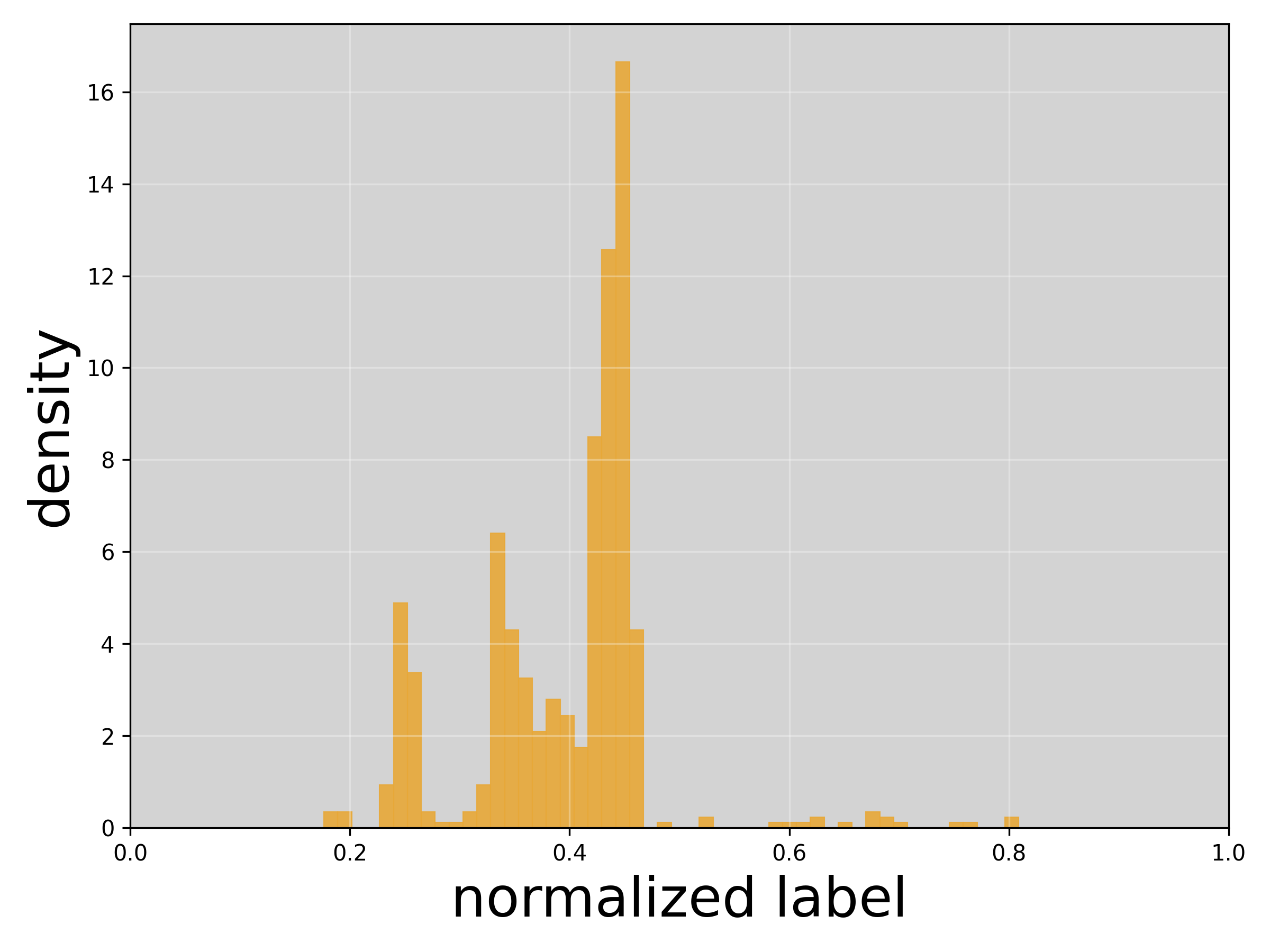}
\subcaption{Digtime}
  \end{minipage}
  \begin{minipage}[t]{0.16\linewidth}  
    \centering
    \includegraphics[width=\linewidth]{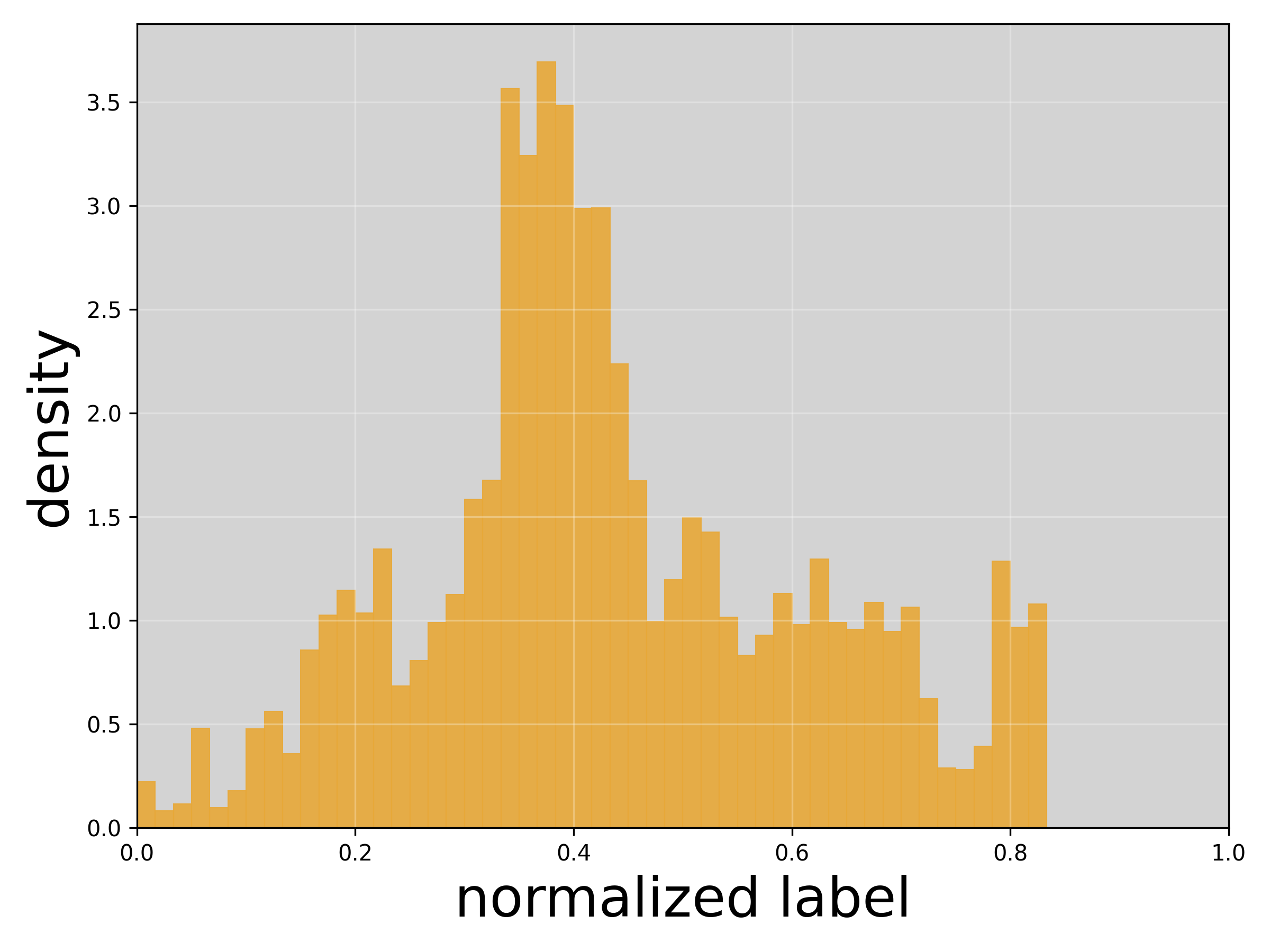}
\subcaption{Sandwich}
  \end{minipage}
  \begin{minipage}[t]{0.16\linewidth}  
    \centering
    \includegraphics[width=\linewidth]{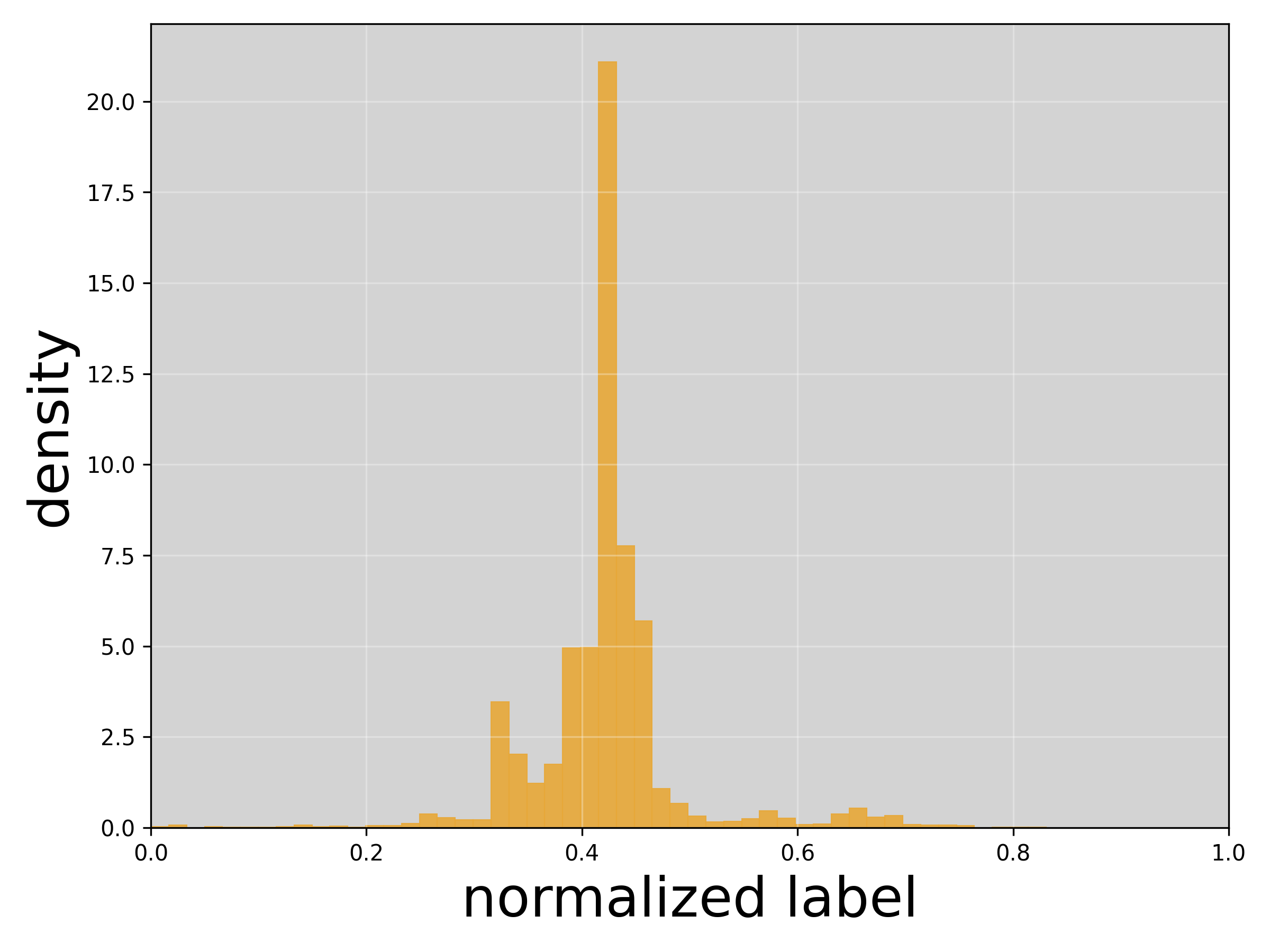}
    \subcaption{SSRAM}
  \end{minipage}
  \begin{minipage}[t]{0.16\linewidth}  
    \centering
    \includegraphics[width=\linewidth]{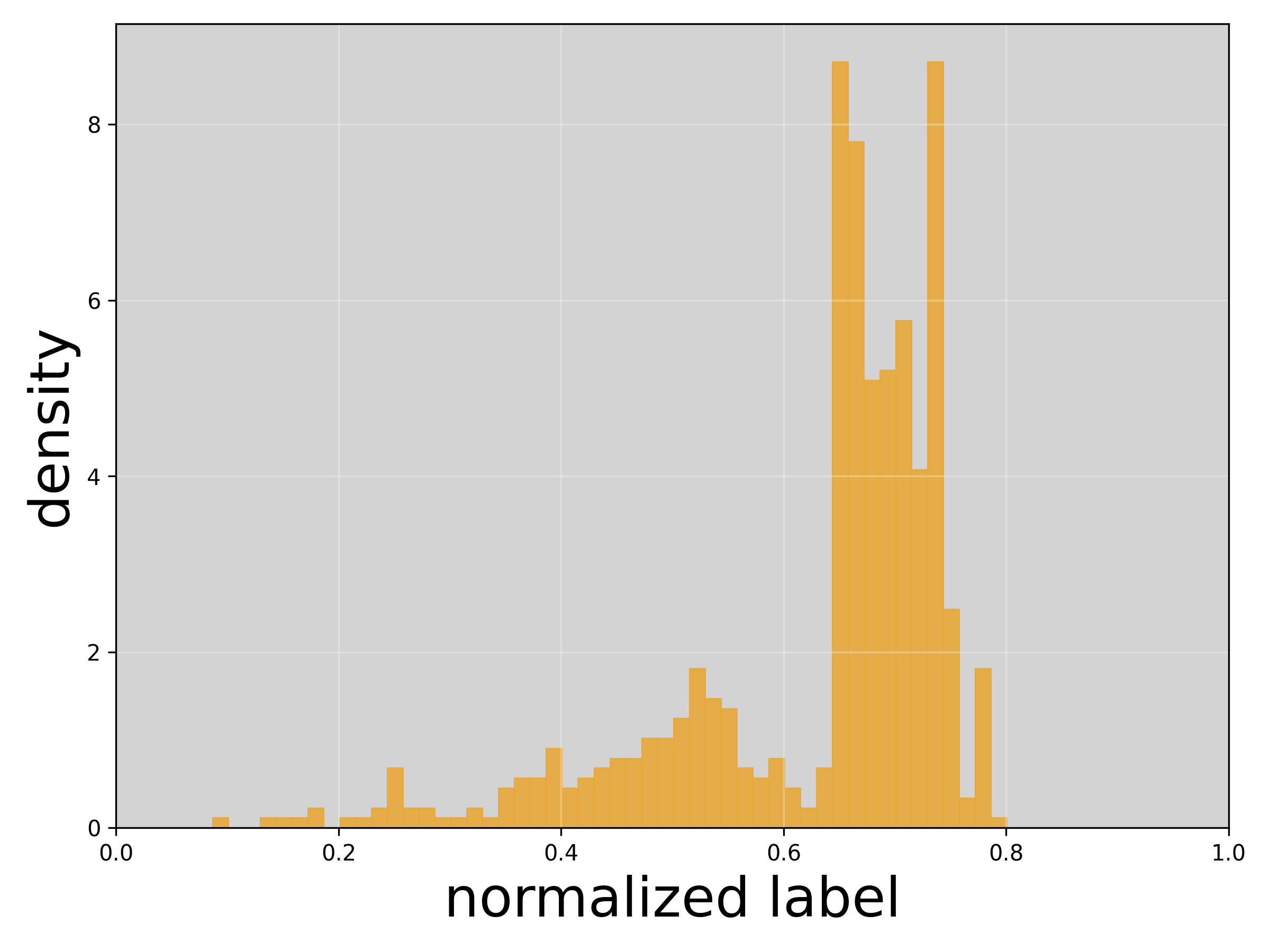}
    \subcaption{Timing\_Ctrl}
  \end{minipage}
  \begin{minipage}[t]{0.16\linewidth}  
    \centering
    \includegraphics[width=\linewidth]{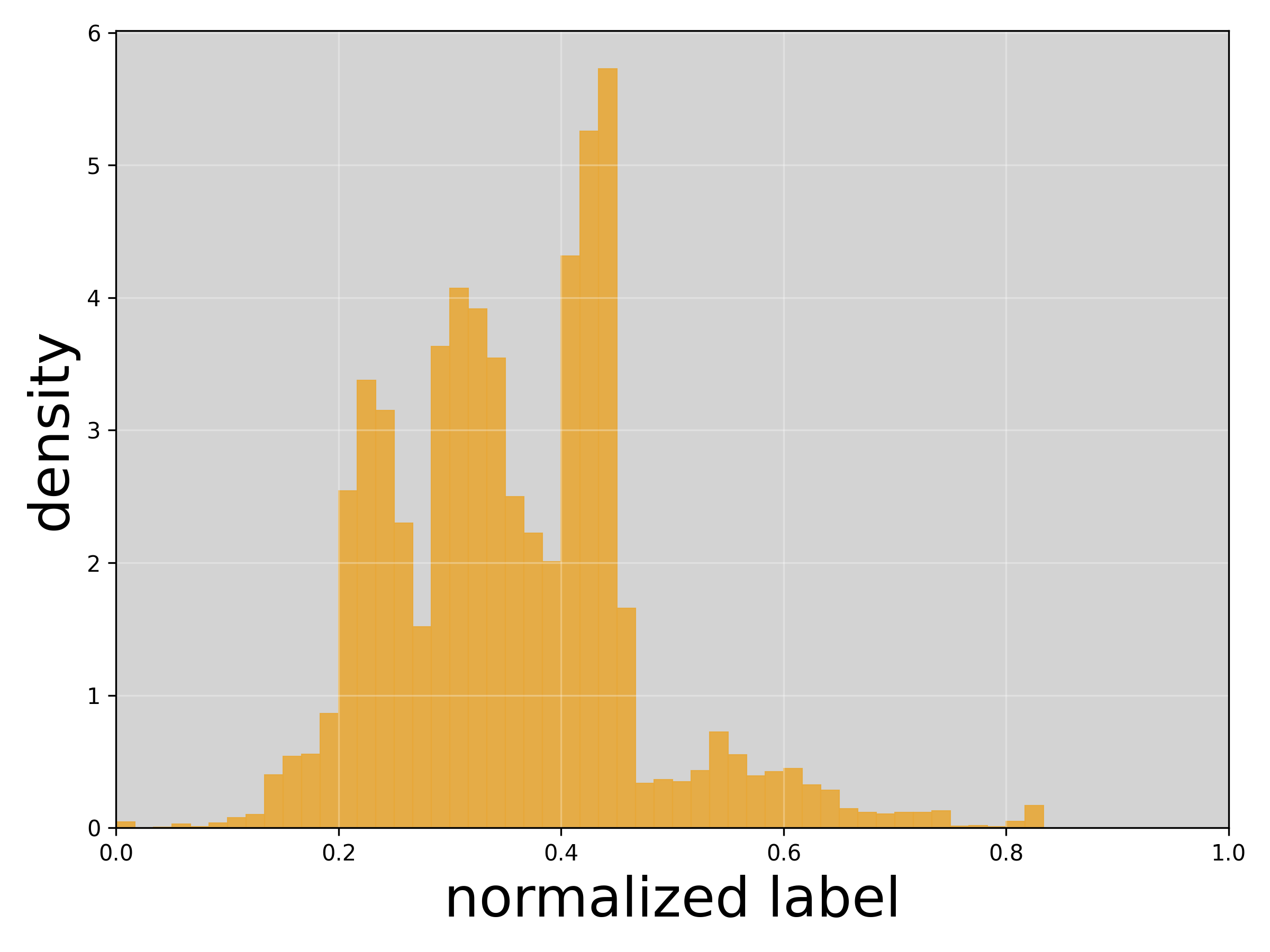}
    \subcaption{Ultra8t}
  \end{minipage}
    \caption{Overview of 6 SRAM Circuit Ground Capacitance Label Distribution}
  \label{fig:sram_ground_each_distribution}
  \hfill
\end{figure*}

%% file: tables/sram_ground_regression.tex
\begin{table*}[t]
  \centering
  \setlength{\tabcolsep}{5pt}
  \caption{Performance of Different Models on SRAM Circuits Ground Capacitance Node Regression Task}
  \label{tab:sram_ground_regression}
  \begin{tabular*}{\textwidth}{@{\extracolsep{\fill}}lcccccccc}
    \toprule
    \multirow{2}{*}{\textbf{Metric}}
      & \multicolumn{2}{c}{\textbf{sram+digtime+timing\_ctrl}}
      & \multicolumn{2}{c}{\textbf{sandwich}}
      & \multicolumn{2}{c}{\textbf{ultra8t}}
      & \multicolumn{2}{c}{\textbf{array\_128\_32\_8t}} \\
    \cmidrule(lr){2-3}\cmidrule(lr){4-5}\cmidrule(lr){6-7}\cmidrule(lr){8-9}
      & MAE $\downarrow$ & R$^2$ $\uparrow$
      & MAE $\downarrow$ & R$^2$ $\uparrow$
      & MAE $\downarrow$ & R$^2$ $\uparrow$
      & MAE $\downarrow$ & R$^2$ $\uparrow$ \\
    \midrule
    GCN & 0.0461 & 0.8401 & 0.1876 & 0.1636 & \textbf{0.0192} & 0.5446 & 0.1199 & -0.1967 \\
    GAT & 0.0454 & 0.8469 & 0.1838 & 0.2159 & 0.0955 & 0.5060 & 0.1211 & -0.2518 \\
    GraphSAGE & 0.0413 & 0.8800 & 0.1964 & 0.0585 & 0.1093 & 0.2620 & 0.2813 & -2.0157 \\
    \midrule
    PNA & 0.0415 & 0.8813 & 0.1845 & 0.3589 & 0.0894 & 0.4561 & 0.2024 & -0.9577 \\
    SGFormer & 0.0424 & 0.8729 & 0.2297 & -0.1733 & 0.1619 & -0.3985 & 0.2934 & -3.2101 \\
    PolyNormer & 0.0423 & 0.8667 & 0.1938 & 0.1186 & 0.1563 & -0.5618 & 0.1699 & -1.1344 \\
    \midrule
    CircuitGPS & \textbf{0.0063} & \textbf{0.9568} & \textbf{0.0298} & \textbf{0.6574} & 0.0222 & \textbf{0.7882} & \textbf{0.0194} & \textbf{0.8702} \\
    CircuitGCL & 0.0412 & 0.8991 & 0.2310 & 0.3726 & 0.1879 & 0.5003 & 0.1678 & 0.5234 \\
    \bottomrule
  \end{tabular*}
\end{table*}

%% file: tables/sram_coupling_regression.tex
\begin{table*}[t]
  \centering
  \setlength{\tabcolsep}{5pt}
  \caption{Performance of Different Models on SRAM Circuits Coupling Capacitance Edge Regression Task}
  \label{tab:sram_coupling_regression}
  \begin{tabular*}{\textwidth}{@{\extracolsep{\fill}}lcccccccc}
    \toprule
    \multirow{2}{*}{\textbf{Metric}}
      & \multicolumn{2}{c}{\textbf{sram+digtime+timing\_ctrl}}
      & \multicolumn{2}{c}{\textbf{sandwich}}
      & \multicolumn{2}{c}{\textbf{ultra8t}}
      & \multicolumn{2}{c}{\textbf{array\_128\_32\_8t}} \\
    \cmidrule(lr){2-3}\cmidrule(lr){4-5}\cmidrule(lr){6-7}\cmidrule(lr){8-9}
      & MAE $\downarrow$ & R$^2$ $\uparrow$
      & MAE $\downarrow$ & R$^2$ $\uparrow$
      & MAE $\downarrow$ & R$^2$ $\uparrow$
      & MAE $\downarrow$ & R$^2$ $\uparrow$ \\
    \midrule
    GCN & 0.0717 & 0.5008 & 0.1073 & 0.3511 & 0.0971 & 0.2866 & 0.0960 & 0.2543 \\
    GAT & 0.0717 & 0.4960 & 0.1075 & 0.3504 & 0.0962 & 0.2969 & 0.0956 & 0.2581 \\
    GraphSAGE & 0.0747 & 0.4701 & 0.1078 & 0.3427 & \textbf{0.0920} & 0.3334 & 0.0783 & 0.4182 \\
    \midrule
    PNA & 0.0715 & 0.5053 & 0.1067 & \textbf{0.3527} & 0.0935 & 0.3083 & 0.0863 & 0.3554 \\
    SGFormer & 0.0766 & 0.4592 & 0.1081 & 0.3212 & 0.0930 & 0.3093 & 0.0746 & \textbf{0.4378} \\
    PolyNormer & 0.0709 & 0.5053 & \textbf{0.1053} & 0.3449 & 0.0934 & 0.2951 & 0.0835 & 0.3797 \\
    \midrule
    CircuitGPS & 0.0714 & 0.4413 & 0.1093 & 0.3412 & 0.0933 & 0.3284 & \textbf{0.0723} & 0.3979 \\
    CircuitGCL & \textbf{0.0698} & \textbf{0.5340} & 0.1124 & 0.3455 & 0.0949 & \textbf{0.3375} & 0.0818 & 0.4067 \\
    \bottomrule
  \end{tabular*}
\end{table*}

%% file: tables/analog_ground_classification.tex
\begin{table*}[t]
  \centering
  \setlength{\tabcolsep}{5pt}
  \caption{Performance of Different Models on Analog Circuits Ground Capacitance Node Classification Task}
  \label{tab:analog_ground_classification}
  \begin{tabular*}{\textwidth}{@{\extracolsep{\fill}}lcccccccc}
    \toprule
    \multirow{2}{*}{\textbf{Metric}}
      & \multicolumn{2}{c}{\makecell[c]{\textbf{1--4, 6,}\\\textbf{8--12, 15--18}}}
      & \multicolumn{2}{c}{\textbf{5}}
      & \multicolumn{2}{c}{\textbf{14}}
      & \multicolumn{2}{c}{\textbf{20}} \\
    \cmidrule(lr){2-3}\cmidrule(lr){4-5}\cmidrule(lr){6-7}\cmidrule(lr){8-9}
      & Accuracy $\uparrow$ & F1-Score $\uparrow$
      & Accuracy $\uparrow$ & F1-Score $\uparrow$
      & Accuracy $\uparrow$ & F1-Score $\uparrow$
      & Accuracy $\uparrow$ & F1-Score $\uparrow$ \\
    \midrule
    GCN & 0.8718 & 0.5357 & 0.9935 & 0.4727 & 0.9740 & 0.3127 & 0.9101 & 0.3426 \\
    GAT & 0.7949 & 0.3000 & 0.9941 & 0.4991 & 0.9761 & 0.3239 & 0.9101 & 0.3426 \\
    GraphSAGE & 0.7949 & 0.3000 & \textbf{0.9946} & \textbf{0.8412} & 0.9761 & 0.3322 & 0.9213 & 0.5640 \\
    \midrule
    PNA & \textbf{0.8974} & 0.5804 & 0.9941 & 0.7028 & \textbf{0.9783} & \textbf{0.5189} & \textbf{0.9326} & \textbf{0.7926} \\
    SGFormer & 0.8462 & 0.4375 & 0.9941 & 0.4991 & \textbf{0.9783} & 0.4122 & 0.9101 & 0.3426 \\
    PolyNormer & \textbf{0.8974} & 0.6476 & 0.9941 & 0.7028 & \textbf{0.9783} & 0.4922 & 0.9213 & 0.5640 \\
    \midrule
    CircuitGPS & 0.8460 & 0.3806 & 0.9911 & 0.2492 & 0.9740 & 0.3215 & 0.8876 & 0.2714 \\
    CircuitGCL & 0.8966 & \textbf{0.9138} & 0.9855 & 0.3318 & 0.9681 & 0.3989 & 0.8981 & 0.3527 \\
    \bottomrule
  \end{tabular*}
\end{table*}

%% file: tables/analog_effective_classification.tex
\begin{table*}[t]
  \centering
  \setlength{\tabcolsep}{5pt}
  \caption{Performance of Different Models on Analog Circuits Effective Resistance Edge Classification Task}
  \label{tab:analog_resistance_classification}
  \begin{tabular*}{\textwidth}{@{\extracolsep{\fill}}lcccccccc}
    \toprule
    \multirow{2}{*}{\textbf{Metric}}
      & \multicolumn{2}{c}{\makecell[c]{\textbf{1--4, 6,}\\\textbf{8--12, 15--18}}}
      & \multicolumn{2}{c}{\textbf{5}}
      & \multicolumn{2}{c}{\textbf{14}}
      & \multicolumn{2}{c}{\textbf{20}} \\
    \cmidrule(lr){2-3}\cmidrule(lr){4-5}\cmidrule(lr){6-7}\cmidrule(lr){8-9}
      & Accuracy $\uparrow$ & F1-Score $\uparrow$
      & Accuracy $\uparrow$ & F1-Score $\uparrow$
      & Accuracy $\uparrow$ & F1-Score $\uparrow$
      & Accuracy $\uparrow$ & F1-Score $\uparrow$ \\
    \midrule
    GCN & 0.5314 & 0.4462 & 0.8474 & 0.4678 & 0.5007 & 0.2875 & 0.3316 & 0.3323 \\
    GAT & 0.5216 & 0.4240 & 0.8018 & 0.3424 & 0.2735 & 0.1651 & 0.1634 & 0.1255 \\
    GraphSAGE & 0.5164 & 0.4238 & 0.8415 & 0.4505 & 0.3934 & 0.2437 & 0.3381 & 0.3395 \\
    \midrule
    PNA & 0.4649 & 0.2971 & 0.8132 & 0.3918 & 0.3058 & 0.1995 & 0.1060 & 0.0809 \\
    SGFormer & 0.5199 & 0.5163 & \textbf{0.8684} & \textbf{0.5575} & 0.4024 & 0.3316 & \textbf{0.3638} & \textbf{0.3618} \\
    PolyNormer & 0.5017 & 0.4229 & 0.2636 & 0.1747 & 0.3565 & 0.2049 & 0.1859 & 0.2265 \\
    \midrule
    CircuitGPS & 0.6491 & 0.6415 & 0.6851 & 0.3573 & 0.2978 & 0.2778 & 0.3360 & 0.2386 \\
    CircuitGCL & \textbf{0.7018} & \textbf{0.7339} & 0.7837 & 0.4491 & 0.6189 & \textbf{0.5545} & 0.3265 & 0.2472 \\
    \bottomrule
  \end{tabular*}
\end{table*}